%% file: main.tex
\documentclass[11pt]{article}
\usepackage[T1]{fontenc}
\usepackage[utf8]{inputenc}
\usepackage{lmodern,graphicx}
\usepackage{amsmath,amssymb,amsthm,mathtools,bm}
\usepackage{aliascnt}
\usepackage{booktabs,tabularx,array}
\usepackage{enumitem}
\usepackage{float}
\usepackage{subcaption}
\usepackage[ruled,vlined,linesnumbered]{algorithm2e}
\usepackage{microtype}
\usepackage[hidelinks]{hyperref}
\usepackage{xcolor}
\usepackage{multirow}

\allowdisplaybreaks
\SetKwInput{KwIn}{Input}
\SetKwInput{KwOut}{Output}
\SetKwInput{Initialize}{Initialize}
\DontPrintSemicolon

\newcommand{\RR}{\mathbb{R}}
\newcommand{\PP}{\mathbb{P}}
\newcommand{\EE}{\mathbb{E}}
\newcommand{\one}{\mathbf{1}}

\newcommand{\cC}{\mathcal{C}}

\providecommand{\cP}{\mathcal{P}}
\newcommand{\xvec}{\bm{x}}

\DeclareMathOperator{\Reg}{Reg}

\usepackage{fullpage}
\usepackage{geometry}
\input{macros}

\usepackage{cleveref}
\crefname{theorem}{Theorem}{Theorems}
\crefname{lemma}{Lemma}{Lemmas}
\crefname{proposition}{Proposition}{Propositions}
\crefname{corollary}{Corollary}{Corollaries}
\crefname{definition}{Definition}{Definitions}
\crefname{assumption}{Assumption}{Assumptions}
\crefname{remark}{Remark}{Remarks}
\crefname{algorithm}{Algorithm}{Algorithms}

\usepackage[round]{natbib}
\definecolor{mydarkblue}{rgb}{0,0.08,0.45}
\definecolor{mydarkred}{rgb}{0.75, 0.27, 0.34}
\hypersetup{
    colorlinks=true,
    citecolor=mydarkblue,
    linkcolor=mydarkred,
}

\newif\ifAISTATSVersion
\AISTATSVersionfalse

\title{Near-Optimal Regret Bounds for Monotone Stochastic Optimization \\ under Semi-Bandit Feedback}

\author{Tao Hu\thanks{Department of Information, Risk, and Operations Management, University of Texas at Austin, Austin, USA.} \and Rohan Ghuge$^*$}

\begin{document}
\pagenumbering{roman}

\maketitle
\thispagestyle{empty}

\input{abstract}

\newpage

\setcounter{tocdepth}{2}
 {\small
    \tableofcontents
    \thispagestyle{empty}
 }
\newpage

\pagenumbering{arabic}

\input{intro}

\input{problem-defn}
\input{results-techniques}

\input{prelim}
\input{related-work}



\input{strongly-monotone}


\input{weakly-monotone}

\input{apps}


\input{computations-syn}


\input{conclusion}

\input{AIUseStatement}

\bibliographystyle{plainnat}
\bibliography{references}

\appendix

\counterwithin{equation}{section}
\counterwithin{table}{section}

\input{weak_bv_appendix}
\input{computations-appendix-syn}

\end{document}

%% file: macros.tex
\renewcommand{\Pr}{\mathbb{P}}

\newcommand{\OPT}{\mathrm{OPT}}

\newcommand{\bP}{\mathbf{P}}

\newcommand{\bU}{\mathbf{U}}

\newtheorem{theorem}{Theorem}[section]
\newtheorem{definition}{Definition}[section]

\newaliascnt{proposition}{theorem}
\newtheorem{proposition}[proposition]{Proposition}
\aliascntresetthe{proposition}

\newaliascnt{claim}{theorem}

\aliascntresetthe{claim}

\newaliascnt{lemma}{theorem}
\newtheorem{lemma}[lemma]{Lemma}
\aliascntresetthe{lemma}

\newaliascnt{observation}{theorem}

\aliascntresetthe{observation}

\newaliascnt{corollary}{theorem}
\newtheorem{corollary}[corollary]{Corollary}
\aliascntresetthe{corollary}

\newaliascnt{remark}{theorem}
\newtheorem{remark}[remark]{Remark}
\aliascntresetthe{remark}

\newcounter{note}[section]

\newcommand{\ignore}[1]{}

\def\sd{\preceq_{\rm{SD}}}
\renewcommand{\cite}[1]{\citep{#1}}

%% file: abstract.tex
\ifAISTATSVersion
\else

\begin{abstract}
Stochastic optimization is a central paradigm for modeling sequential decision-making under uncertainty. Uncertain input parameters are modeled by random variables, and algorithms typically assume that their distributions are known. Several recent works study learnability when these distributions are unknown, developing tailor-made learning algorithms for individual problems. 
\citet{agarwal2026semibandit} give an \emph{online-to-offline reduction} for monotone stochastic optimization under semi-bandit feedback, where only variables probed by the policy are observed. Given a known-distribution algorithm, it achieves expected regret \(\widetilde O(n\sqrt T)\) relative to its offline guarantee, where \(n\) is the number of variables and \(T\) the horizon.
For optimal stopping and Pandora's box, \citet{liu2025improved} obtain the improved bound \(\widetilde O(\sqrt{nT})\) through problem-specific learning algorithms. We ask: \emph{Can an online-to-offline reduction achieve this improved regret bound for a general class of stochastic optimization problems?}

We give such a reduction for ``strongly monotone'' stochastic optimization, answering an open question of \citet{agarwal2026semibandit}. Strong monotonicity is defined relative to the chosen offline algorithm. Our oracle-efficient algorithm achieves expected regret \(\widetilde O(\sqrt{nT})\) and preserves the offline approximation guarantee for approximation oracles satisfying the same condition. It invokes the offline oracle on optimistic distribution estimates constructed using Bernstein-type confidence bounds. The key technical ingredient is a Hellinger stability bound for adaptive policy transcripts, which controls changes in policy value by the square root of probing-weighted squared marginal errors. The rate is optimal up to logarithmic factors, even under full feedback.

Under monotonicity and a fixed uniform bound on coordinate-wise variation, the same algorithm achieves expected approximation regret \(\widetilde O(n^{3/4}\sqrt T)\), improving the linear dependence on \(n\) in \citet{agarwal2026semibandit} to \(n^{3/4}\). We show that this exponent is sharp for our algorithm's ordinary regret in the worst case over instances and exact oracle choices. Our framework applies to prophet inequality, Pandora's box, best-of-$K$ selection, stochastic knapsack, stochastic orienteering and stochastic submodular maximization.

\end{abstract}

\noindent\textbf{2020 Mathematics Subject Classification.}
Primary 68W27; Secondary 68Q32, 90C15, 68W25.

\medskip
\noindent\textbf{Keywords.}
Online Learning, Limited Feedback, Bandit Algorithms, Online-to-Offline Reductions, Strong Monotonicity, Stochastic Optimization
\fi

%% file: intro.tex
\section{Introduction}
\label{sec:introduction}

Stochastic optimization is a central paradigm for sequential decision-making under uncertainty. It models uncertain inputs as random variables and seeks policies that adapt their decisions to the information revealed along the way. This approach has led to exact and approximation algorithms for a wide range of problems, from optimal stopping and search to stochastic knapsack and submodular maximization \citep{samuelcahn1984comparison,weitzman1979optimal,dean2008stochastic,asadpour2016maximizing}. These algorithms typically assume that the input distributions are known, an assumption often justified by the availability of historical data. When such data are limited or unavailable, however, learning the distributions becomes part of the decision-making problem itself.

A natural approach is to learn through repeated interaction with the unknown distributions. Over \(T\) rounds, a learner chooses policies using past observations, with the goal of minimizing regret relative to a benchmark that knows the distributions. Recent works develop problem-specific learning algorithms for prophet inequality, Pandora's box, and sequential posted pricing, obtaining sublinear regret even under bandit feedback, where only the realized objective value is observed \citep{gatmiry2024bandit,singla2024bandit}. More generally, \citet{agarwal2026semibandit} give an \emph{online-to-offline reduction} for a broad class of monotone stochastic optimization problems under semi-bandit feedback, where the learner observes the random variables probed by its policy. Their reduction turns an algorithm for the known-distribution problem into a learning algorithm while preserving its approximation guarantee. The resulting regret has \(\widetilde O(\sqrt T)\) dependence on the horizon but linear dependence on the number \(n\) of uncertain inputs. For prophet inequality and Pandora's box, \citet{liu2025improved} improve this dimension dependence to \(\sqrt n\) through tailor-made semi-bandit learning algorithms. This leads to our central question: \emph{Can an online-to-offline reduction achieve square-root dependence on \(n\) for a general class of stochastic optimization problems?}

We answer this question affirmatively for strongly monotone stochastic optimization, resolving an open question of \citet{agarwal2026semibandit}. Here, strong monotonicity is defined relative to the chosen offline algorithm. Our oracle-efficient reduction achieves expected regret \(\widetilde O(G\sqrt{nT})\), where \(G\) is the objective range, and extends to approximation oracles satisfying the same condition. This rate is optimal up to logarithmic factors, even under full feedback. We also show that the earlier algorithm cannot achieve this improved rate: on a family of single-choice prophet instances, its continuous-distribution version incurs \(\Omega(T)\) regret at horizons \(T=\Theta(n^2)\), whereas our guarantee is \(\widetilde O(T/\sqrt n)\). This establishes a \(\sqrt n\) separation up to logarithmic factors; see \Cref{thm:agn-continuous-lower-bound}.

We also improve the dimension dependence under monotonicity without requiring the strong-monotonicity condition. 
For objectives normalized to \([0,1]\) and a fixed uniform bound on coordinate-wise variation, the same algorithm achieves expected approximation regret \(\widetilde O(n^{3/4}\sqrt T)\), improving the linear dependence on \(n\) in the earlier framework while preserving the offline approximation guarantee. The exponent \(3/4\) is sharp for our algorithm's ordinary regret in the worst case over admissible instances and exact oracle choices.
We complement these guarantees with computational experiments on best-of-\(K\) selection, prophet inequality, and Pandora's box (\Cref{sec:computations}). The experiments show competitive performance against prior work and other baselines.

\paragraph{Organization.} The remainder of this section introduces the model, summarizes our results and techniques, and reviews preliminaries and related work. \Cref{sec:strongly-monotone,sec:weak-monotone} analyze our algorithm under strong monotonicity and monotonicity with bounded variation, respectively. \Cref{sec:apps,sec:computations} present applications and computational experiments. We conclude with open questions. Detailed proofs of the bounded-variation results appear in Appendix~\ref{app:weak-bv-proof}.

%% file: problem-defn.tex
\subsection{Problem Setup}
\label{sec:model}

We begin by describing the stochastic optimization problem when the input
distributions are known.

\paragraph{Stochastic optimization.}
There are \(n\) items, where item \(i\in[n]\) is associated with an
independent real-valued random variable \(X_i\sim D_i\).  We use
\( \xvec=(X_1,\ldots,X_n) \sim \mathbf D:=\prod_{i=1}^n D_i\)
to denote the random outcome vector and its product distribution.  Throughout
the paper, the distributions are supported on known bounded intervals.  By
rescaling each coordinate, we may assume without loss of generality that
\(D_i\) is supported on \([0,1]\).

The realization of \(X_i\) is revealed only when item \(i\) is probed.  A policy \(\sigma\) probes items sequentially, where the choice of the next item may depend on the items probed so far and their observed realizations.  The policy may also stop at any point.  We assume that an item is probed at most once. 
We restrict the feasible policy class to deterministic policies. This is without loss for the optimal value, since a randomized policy is a distribution over deterministic policies and cannot improve on the optimal deterministic value. The offline oracle and the online learner may nevertheless randomize their choice of a feasible policy.

Equivalently, a policy can be represented by a decision tree.  Every internal
node is labeled by an item to probe, and its outgoing branches correspond to
the possible realizations of that item.  A policy execution follows a
root-to-leaf path in this tree.  For a realization
\(\xvec=(x_1,\ldots,x_n)\), let
\[
    \tau_\sigma(\xvec)
    =
    \bigl((i_1,x_{i_1}),\ldots,(i_k,x_{i_k})\bigr)
\]
denote the transcript of the execution of \(\sigma\) under realization \(\xvec\), namely, the sequence of items probed by \(\sigma\) together with their observed realizations. We also define \(S(\sigma,\xvec):=\{i_1,\ldots,i_k\}\) as the set of items probed during this execution.
Let \(\cC\) denote the set of feasible policies.  This set captures all
problem-specific constraints, including restrictions on which item may be
probed after any observed history.  Importantly, \(\cC\) does not depend on
the input distributions.

For a maximization problem, let \(r(\sigma,\xvec)\) denote the reward obtained
by policy \(\sigma\) under realization \(\xvec\).  We assume that the reward
is determined entirely by the observed transcript
\(\tau_\sigma(\xvec)\).  In particular, the reward cannot depend on the
realizations of unprobed items.  We also assume that
\( r(\sigma,\xvec)\in[r_{\min},r_{\max}]\)
for every feasible policy and realization, and write \( G:=r_{\max}-r_{\min}\)
for the range of the objective.
For any product distribution
\(\mathbf P = \prod_{i=1}^n P_i\), define the expected reward of
\(\sigma\) by
\(
    R(\sigma\mid\mathbf P)
    :=
    \EE_{\xvec\sim\mathbf P}
    \bigl[r(\sigma,\xvec)\bigr].
\)
The optimal value under \(\mathbf P\) is
\(
    \OPT(\mathbf P)
    :=
    \max_{\sigma\in\cC}
    R(\sigma\mid\mathbf P).
\)
Minimization problems are defined analogously.  In this case,
\(c(\sigma,\xvec)\) denotes the realized cost,
\(
    J(\sigma\mid\mathbf P)
    :=
    \EE_{\xvec\sim\mathbf P}
    \bigl[c(\sigma,\xvec)\bigr],
\)
and
\(
    \OPT(\mathbf P)
    :=
    \min_{\sigma\in\cC}
    J(\sigma\mid\mathbf P).
\)
We again assume that the realized objective lies in an interval of length at
most \(G\).

Many stochastic optimization problems are computationally hard even when the
input distributions are known.  We therefore allow the online algorithm to
use an approximation algorithm for the corresponding known-distribution
problem.  For a maximization problem, an \(\alpha\)-approximation algorithm,
where \(\alpha\in(0,1]\), takes a product distribution \(\mathbf P\) as input
and returns a policy \(\sigma\in\cC\) satisfying
\(
    R(\sigma\mid\mathbf P)
    \ge
    \alpha\cdot\OPT(\mathbf P).
\)
For a minimization problem, an \(\alpha\)-approximation algorithm, where
\(\alpha\ge1\), returns a policy satisfying
\(
    J(\sigma\mid\mathbf P)
    \le
    \alpha\cdot\OPT(\mathbf P).
\)
Whenever we consider multiplicative approximation guarantees, we additionally
assume that the objective is non-negative.

We also allow randomized approximation oracles. For each input \(\mathbf P\), such an oracle returns a deterministic policy \(\mathcal O(\mathbf P;\xi)\in\cC\), where \(\xi\) denotes its internal randomness. The approximation guarantee is then required in expectation:
\[
\EE_\xi\!\left[R(\mathcal O(\mathbf P;\xi)\mid\mathbf P)\right]\ge\alpha\cdot\OPT(\mathbf P)
\]
for maximization, and \(\EE_\xi[J(\mathcal O(\mathbf P;\xi)\mid\mathbf P)]\le\alpha\cdot\OPT(\mathbf P)\) for minimization. These guarantees hold for every input distribution, but need not hold for every realization of \(\xi\). In the online setting, each oracle call uses fresh internal randomness independent of the past and the current outcome vector.

\paragraph{The online semi-bandit setting.}
In the online problem, the underlying product distribution \(\mathbf D=\prod_{i=1}^nD_i\) is unknown and remains fixed over \(T\) rounds. 
In every round \(t\in[T]\), the learner uses its observations
from the previous rounds to select a policy \(\sigma^t\in\cC\).  A fresh
outcome vector \(\xvec^t\sim\mathbf D\)
is then drawn independently, and the learner executes \(\sigma^t\) on
\(\xvec^t\).
The learner receives \emph{semi-bandit feedback}: it observes \(X_i^t\) if
and only if policy \(\sigma^t\) probes item \(i\).  Thus, the feedback in round \(t\) is precisely the ordered transcript
\(
    \tau_{\sigma^t}(\xvec^t).
\)

For a maximization problem, the pseudo-regret after \(T\) rounds is
\begin{equation}
\label{eq:pseudo-regret}
    R(T)
    :=
    \sum_{t=1}^T
    \left(
        \OPT(\mathbf D)
        -
        R(\sigma^t\mid\mathbf D)
    \right).
\end{equation}
More generally, when the learner uses an \(\alpha\)-approximation algorithm,
we compare its performance with the corresponding approximation to the
optimal value, and define $\alpha$-regret as:
\begin{equation}
\label{eq:alpha-pseudo-regret-max}
    \alpha\text{-}R(T)
    :=
    \sum_{t=1}^T
    \left(
        \alpha\cdot\OPT(\mathbf D)
        -
        R(\sigma^t\mid\mathbf D)
    \right).
\end{equation}

For a minimization problem, the corresponding definitions are
\[
    R(T)
    :=
    \sum_{t=1}^T
    \left(
        J(\sigma^t\mid\mathbf D)-\OPT(\mathbf D)
    \right),
    \qquad
    \alpha\text{-}R(T)
    :=
    \sum_{t=1}^T
    \left(
        J(\sigma^t\mid\mathbf D)-\alpha\cdot\OPT(\mathbf D)
    \right).
\] 
The policies \(\sigma^t\) depend on observations from earlier rounds and any internal randomization, so the regret is itself random. We evaluate a learning algorithm by its expected (\(\alpha\)-)regret, where the expectation is over the outcome vectors and all internal randomness.

We assume throughout that \(T\ge n\); when \(T<n\), the trivial \(GT\) upper bound on (\(\alpha\)-)regret is no larger than \(G\sqrt{nT}\). 
We also assume throughout that policies, oracle outputs, and the corresponding
execution and reward maps are measurable.  This ensures that all conditional
expectations and transcript distributions used below are well defined.

%% file: results-techniques.tex
\subsection{Results and Techniques}

We first recall the notion of stochastic dominance.  For a distribution \(P\),
let \(F_P\) denote its cumulative distribution function.

\begin{definition}[Stochastic dominance]
\label{def:stochastic-dominance}
For two distributions \(P\) and \(Q\), we write \(P \sd Q\) if
\[
    F_P(z)\ge F_Q(z)
    \qquad\text{for every }z\in\RR.
\]
Thus, \(Q\) stochastically dominates \(P\).  For product distributions
\(\mathbf P=\prod_{i=1}^nP_i\) and
\(\mathbf Q=\prod_{i=1}^nQ_i\), we write
\(\mathbf P \sd \mathbf Q\) if \(P_i \sd Q_i\) for every \(i\in[n]\).
\end{definition}

We call one objective value \emph{at least as good as} another if it is larger for a maximization problem and smaller for a minimization problem.

\begin{definition}[Monotonicity]
\label{def:monotone}
A stochastic problem \(\cP\) is \emph{(weakly) up-monotone} if, for every pair of product distributions \(\bU\sd\bU'\), the value \(\OPT(\bU')\) is at least as good as \(\OPT(\bU)\).  Similarly, \(\cP\) is \emph{(weakly) down-monotone} if \(\OPT(\bU)\) is at least as good as \(\OPT(\bU')\).  We call \(\cP\) \emph{(weakly) monotone} if it is up- or down-monotone.
\end{definition}

Our first result relies on a transfer property of a fixed oracle selection, following \citet{guo2021generalizing}. It gives a tight regret guarantee, up to logarithmic factors, for \emph{strongly monotone problems}.

\begin{definition}[Oracle-compatible Strong Monotonicity]
\label{def:strong-monotone}
Let \(\mathcal O\) map every product distribution to a feasible policy for a stochastic problem \(\cP\). The pair \((\cP,\mathcal O)\) has \emph{oracle-compatible strong up-monotonicity} if \(\cP\) is up-monotone and, whenever \(\bU\sd\bU'\), the expected objective of \(\mathcal O(\bU)\) under \(\bU'\) is at least as good as its expected objective under \(\bU\). Similarly, \((\cP,\mathcal O)\) has \emph{oracle-compatible strong down-monotonicity} if \(\cP\) is down-monotone and the expected objective of \(\mathcal O(\bU')\) under \(\bU\) is at least as good as its expected objective under \(\bU'\). We call \(\cP\) \emph{strongly up-monotone} or \emph{strongly down-monotone} if it admits an exact oracle satisfying the corresponding condition, and \emph{strongly monotone} if either condition holds.
\end{definition}

For a randomized approximation oracle, the transfer comparisons in \Cref{def:strong-monotone} are required in expectation over its internal randomness. For example, oracle-compatible strong up-monotonicity for maximization requires \(\cP\) to be up-monotone and
\[
\EE_\xi\!\left[R(\mathcal O(\mathbf P;\xi)\mid\mathbf Q)\right]\ge\EE_\xi\!\left[R(\mathcal O(\mathbf P;\xi)\mid\mathbf P)\right]
\]
whenever \(\mathbf P\sd\mathbf Q\). The down-monotone and minimization cases are interpreted analogously. All subsequent transfer comparisons involving a randomized oracle are understood in this sense.

An exact oracle satisfying the relevant condition is said to \emph{witness} strong monotonicity. This formulation makes explicit the optimizer selection implicit in \citet{guo2021generalizing}. 
For conciseness, we sometimes drop the term ``oracle-compatibility'' and instead simply use \emph{strong monotonicity}.

\begin{theorem}[Informal]
\label{thm:main-informal}
Consider a stochastic optimization problem \(\cP\) with \(n\) items and objective range \(G\), and let \(\mathcal O\) be an exact oracle witnessing strong monotonicity. There is an oracle-efficient semi-bandit learning algorithm using \(\mathcal O\) whose expected regret after \(T\) rounds is \(O\left(G\sqrt{nT}\,\log^{3/2}(2nT)\right)\).
\end{theorem}

\Cref{thm:main-informal} answers the open question posed by \citet{agarwal2026semibandit} of whether strong monotonicity permits improved dependence on the number of items. Their general framework obtains regret with linear dependence on \(n\), whereas our bound has square-root dependence. 
This also extends the \(\sqrt n\) dependence obtained by \citet{liu2025improved} for Pandora's box and prophet inequality to the general class of strongly monotone stochastic optimization problems, with regret scaled by the objective range \(G\). 
For Pandora's box, the problem-specific analysis of \citet{liu2025improved} additionally removes the dependence on total inspection cost; our generic bound retains this dependence through \(G\). See \Cref{sec:apps} for details.

Our main technical contribution is a stability bound for adaptive policies. For every feasible policy \(\sigma\) and product distributions \(\mathbf P=\prod_iP_i\) and \(\mathbf Q=\prod_iQ_i\), \Cref{lem:stability} gives
\[
|R(\sigma\mid\mathbf P)-R(\sigma\mid\mathbf Q)|\le 2\sqrt{2}\,G\left(\sum_{i=1}^n\pi_i^{\mathbf P}(\sigma)H^2(P_i,Q_i)\right)^{1/2},
\]
where \(H\) denotes Hellinger distance and \(\pi_i^{\mathbf P}(\sigma)\) is the probability that \(\sigma\) probes item \(i\) under \(\mathbf P\). 
Our transcript-level Hellinger analysis is related to sequential information comparisons in interactive decision making \citep{foster2021statistical,foster2023foundations,foster2024online}. We apply the sequential Hellinger inequality \citep[Theorem~1]{wu2024chainrulehellinger} to adaptive probing, obtaining a comparison whose weights match the learner's observation probabilities under the true distribution.

The continuous-distribution analysis of \citet{agarwal2026semibandit} also weights marginal errors by probing probabilities, but its Kolmogorov--Smirnov stability bound sums these errors in an \(\ell_1\)-like fashion under threshold-policy and bounded-variation assumptions. Our bound combines observation-sensitive weighting with an \(\ell_2\)-like accumulation of marginal errors for arbitrary adaptive policies. The probing weights are essential in the online setting: an item that is rarely observed may remain poorly estimated, but its contribution is reduced by the probability that the policy uses it.
\citet{liu2025improved} obtain a similar probing-weighted square-root aggregation for prophet inequality and Pandora's box using properties of conditional reward functions and telescoping identities specific to these problems. Our transcript-level Hellinger bound provides a general route to this aggregation for arbitrary adaptive policies.

To apply the stability bound, we use the Bernstein shading construction of \citet{guo2021generalizing}, also used by \citet{liu2025improved}. Consider strongly up-monotone maximization, and let \(\sigma^t=\mathcal O(\widetilde{\mathbf E}^{\,t})\) be the policy selected under the shaded empirical distribution in round \(t\). For atomless marginals, the true and shaded empirical distributions are mutually singular, so their Hellinger distance does not vanish as observations accumulate. We therefore introduce an auxiliary distribution \(\mathbf U^t\), used only in the analysis, such that with high probability \(\mathbf D\sd\widetilde{\mathbf E}^{\,t}\sd\mathbf U^t\), while each marginal of \(\mathbf U^t\) is close to the corresponding true marginal in Hellinger distance. Oracle-compatible strong monotonicity then gives
\[
\OPT(\mathbf D)-R(\sigma^t\mid\mathbf D)\le R(\sigma^t\mid\mathbf U^t)-R(\sigma^t\mid\mathbf D).
\]
Thus, strong monotonicity supplies the comparison needed to apply the general stability bound.

Write \(N_i^t\) for the number of observations of item \(i\) before round \(t\). The shading estimate gives \(H^2(D_i,U_i^t)=\widetilde O(1/\max\{1,N_i^t\})\). Applying the stability bound with \(\mathbf P=\mathbf D\) makes its weights exactly the probabilities with which the executed policy acquires observations. The adaptive counting argument used in \citet{agarwal2026semibandit} then gives
\[
\sum_{t=1}^T\sum_{i=1}^n\EE\left[\frac{\pi_i^{\mathbf D}(\sigma^t)}{\max\{1,N_i^t\}}\right]=O(n\log(eT)).
\]
Cauchy--Schwarz across rounds yields the \(\widetilde O(G\sqrt{nT})\) regret guarantee after accounting separately for the concentration failure.

The choice of confidence construction also matters. The continuous-distribution algorithm of \citet{agarwal2026semibandit} applies a uniform shift to each empirical CDF. \Cref{thm:agn-continuous-lower-bound} shows that this algorithm incurs \(\Omega(n\sqrt T)\) regret on a strongly monotone instance at a horizon \(T=\Theta(n^2)\). Consequently, a sharper analysis alone cannot give that algorithm a uniform \(\widetilde O(\sqrt{nT})\) guarantee.

For \(T\ge n\), the dependence on \(n\) and \(T\) in \Cref{thm:main-informal} is optimal up to logarithmic factors.  This follows from the sample-complexity lower bound for strongly monotone single-parameter revenue maximization due to \citet{guo2019settling}; see also \citet[Theorem~18]{guo2021generalizing}.

\begin{theorem}[Regret lower bound]
\label{thm:strong-regret-lower-bound}
For every \(n,T\ge1\) and \(G>0\), there is a strongly monotone stochastic optimization problem with \(n\) items and objective range \(G\) such that every online learning algorithm incurs
\(
\EE[R(T)]=\Omega\left(G \cdot \min\left\{T,\sqrt{nT}\right\}\right)
\)
on some product distribution, even under full feedback.
\end{theorem}

\begin{proof}
It suffices to prove the result for \(G=1\), since the general case follows by scaling.  Given a full-feedback online algorithm, run it for \(T\) rounds on \(T\) independent samples and return a uniformly random policy among \(\sigma^1,\ldots,\sigma^T\).  The expected suboptimality of the returned policy is exactly \(\EE[R(T)]/T\).  The sample-complexity lower bound of \citet{guo2019settling} states that \(\Omega(n/\varepsilon^2)\) samples are necessary to obtain expected additive error \(\varepsilon\) for a strongly monotone problem.  Taking \(\varepsilon=\Theta(\min\{1,\sqrt{n/T}\})\) therefore gives \(\EE[R(T)]=\Omega(T\varepsilon)=\Omega(\min\{T,\sqrt{nT}\})\).
\end{proof}

Since the offline optimization problem may itself be NP-hard, we also extend the preceding upper bound to approximation oracles. 
The same oracle-compatible condition applies: the optimal value must be monotone, as in \Cref{def:monotone}, and the oracle must satisfy the corresponding transfer property. For exact oracles, monotonicity of the optimal value follows from optimality and the transfer property; for approximation oracles, it must be assumed separately.

\begin{theorem}[Informal]
\label{thm:approximate-oracle-informal}
Consider a stochastic optimization problem with \(n\) items and objective range \(G\), and let \(\mathcal O\) be an \(\alpha\)-approximation oracle, where \(\alpha\in(0,1]\) for maximization and \(\alpha\ge1\) for minimization.  If the pair \((\cP,\mathcal O)\) has oracle-compatible strong monotonicity, then there is an oracle-efficient semi-bandit
learning algorithm and an absolute constant \(C>0\) such that $\EE[\alpha\text{-}R(T)] \le C G\sqrt{nT}\,\log^{3/2}(2nT).$

\end{theorem}

For a monotone problem, \Cref{lem:nonadaptive-transfer} gives a black-box sufficient condition for oracle compatibility: every fixed policy that the oracle may return must have a coordinatewise monotone realized objective in the corresponding direction. The criterion also applies to randomized oracles whose approximation guarantees hold in expectation. We apply it to nonadaptive approximation algorithms returning fixed sets or orderings.

We obtain expected regret or approximation regret \(\widetilde O(G\sqrt{nT})\) for series-system testing, best-of-\(K\) selection, single- and multiple-choice prophet inequalities, Pandora's box, stochastic knapsack, stochastic orienteering, and stochastic monotone submodular maximization. \Cref{tab:application-summary} lists the approximation factors and distributional assumptions; \Cref{sec:apps} gives the application-specific arguments and comparisons with prior work.

We next ask what the same algorithm guarantees under monotonicity without an oracle transfer assumption. Monotonicity alone is insufficient: \Cref{thm:weak-monotonicity-linear-regret} constructs a one-item problem and an exact oracle for which \Cref{alg:strong-up} incurs linear regret. 
However, we obtain a sublinear guarantee by controlling the variation of the oracle's output policies. Specifically, we require their realized rewards or costs to have \emph{pointwise total variation} at most $V_0$ in each coordinate, uniformly over the remaining coordinates. We call this condition \emph{uniform separate bounded variation} (see \eqref{eq:weak-bv-definition} and \eqref{eq:weak-bv-assumption}). The guarantee applies to both maximization and minimization, in either monotonicity direction.

\begin{theorem}[Informal]\label{thm:monotone-bv-informal}
Consider an up- or down-monotone maximization or minimization problem with $n$ items and realized objectives in $[0,1]$. Let $\mathcal O$ be a possibly randomized $\alpha$-approximation oracle, where $\alpha\in(0,1]$ for maximization and $\alpha\ge1$ for minimization, whose output policies have uniform separate variation at most $V_0<\infty$. For $T\ge\max\{n,2\}$, \Cref{alg:strong-up}, using $\mathcal O$, $\delta=1/(T+1)$, and reverse shading in the down-monotone case, has expected $\alpha$-regret at most $\widetilde O((1+V_0)n^{3/4}\sqrt T)$.
\end{theorem}

\Cref{thm:monotone-bv-informal} also improves the dimension dependence of the earlier framework without requiring strong monotonicity. We extend the Kolmogorov--Smirnov analysis of \citet{agarwal2026semibandit} to the same separate-variation assumption, obtaining expected \(\alpha\)-regret \(O(V_0n\sqrt{T\log(2nT)}+1)\). Thus, for fixed \(V_0>0\), our bound improves the polynomial dependence on \(n\) from \(n\) to \(n^{3/4}\), up to logarithmic factors.
In fact, their regularity assumptions imply ours. For rewards in \([0,1]\), a \(k\)-threshold policy whose leaf reward functions have separate variation at most \(V_{\mathrm{leaf}}\) satisfies \(V_0\le kV_{\mathrm{leaf}}+(k-1)\). Consequently, under their assumptions, our expected \(\alpha\)-regret bound is \(\widetilde O(k(1+V_{\mathrm{leaf}})n^{3/4}\sqrt T)\), preserving the offline approximation factor. Our condition also permits bounded-variation rewards with infinitely many jumps and requires no finite threshold representation. The detailed comparison appears in Appendix~\ref{sec:app-agarwal-comparison}.

\Cref{thm:monotone-bv-informal} improves the expected \(\alpha\)-regret for stochastic matching from \(\widetilde O(Gn\sqrt T)\) to \(\widetilde O(Gn^{3/4}\sqrt T)\), preserving the offline approximation factors of \citet{agarwal2026semibandit}. It also improves the dependence on \(n\) for stochastic knapsack with fixed support size while retaining the adaptive \((1/2-\varepsilon)\)-approximation. 

For every fixed \(V_0>0\), \Cref{thm:weak-bv-exact-exponent} constructs instances and exact oracles for which our algorithm incurs expected ordinary regret \(\Omega_{V_0}(n^{3/4}\sqrt T)\) at horizons \(T=\Theta(n^{3/2})\). Thus no bound of the form \(\widetilde O(n^\beta\sqrt T)\) with \(\beta<3/4\) can hold uniformly for this algorithm over admissible instances and exact oracle selections.

Monotonicity alone does not justify the oracle transfer step involving the auxiliary distribution \(\mathbf U^t\), so we instead compare the selected policy's value under the shaded empirical and true distributions directly.
The improvement over the linear dependence on \(n\) in prior work comes from a variance-sensitive stability bound. The Kolmogorov--Smirnov analysis of \citet{agarwal2026semibandit} bounds the marginal estimation errors separately and sums their contributions over items. 
The bounded-variation assumption lets us obtain a tighter error bound when the continuation reward has low variance.
We then control the accumulated conditional variance along a policy execution by telescoping the continuation values and their squares under the true distribution. This allows us to bound the marginal contributions jointly, rather than paying separately for the worst-case error of each item. Combining this stability bound with adaptive counting yields the \(n^{3/4}\) dependence.

%% file: prelim.tex
\subsection{Preliminaries}\label{sec:prelim}

We begin by recording the statistical distances and concentration inequalities used in our analysis. For probability distributions \(P\) and \(Q\) on a common measurable space, let \(\mu:=P+Q\) and define their \emph{squared Hellinger} distance by
\[
H^2(P,Q):=\frac12\int\left(\sqrt{\frac{\mathrm dP}{\mathrm d\mu}}-\sqrt{\frac{\mathrm dQ}{\mathrm d\mu}}\right)^2\mathrm d\mu.
\]
We write \(H(P,Q):=\sqrt{H^2(P,Q)}\) and define their \emph{total variation} distance by \(d_{\mathrm{TV}}(P,Q):=\sup_A|P(A)-Q(A)|\), where the supremum is over measurable events. If \(f\) takes values in an interval of length \(G\), then
\begin{equation}\label{eq:tv-hellinger}
\left|\int f\,\mathrm dP-\int f\,\mathrm dQ\right|\le G\,d_{\mathrm{TV}}(P,Q)\le\sqrt{2}\,G\,H(P,Q).
\end{equation}

We also use the following sequential Hellinger inequality. We state the constant-$4$ version recorded by \citet[Theorem~1]{wu2024chainrulehellinger}; see also the constant-$7$ version in \citet[Lemma~D.2]{foster2024online}.

\begin{lemma}[Sequential Hellinger chain rule]\label{lem:hellinger-chain-rule}
Let \(P\) and \(Q\) be distributions of \(Y_{1:m}=(Y_1,\ldots,Y_m)\), with regular conditional distributions \(P_t(\cdot\mid Y_{<t})\) and \(Q_t(\cdot\mid Y_{<t})\), where \(Y_{<t}:=(Y_1,\ldots,Y_{t-1})\). Then
\[
H^2(P,Q)\le4\sum_{t=1}^m\EE_{Y_{<t}\sim P}\left[H^2\left(P_t(\cdot\mid Y_{<t}),Q_t(\cdot\mid Y_{<t})\right)\right].
\]
\end{lemma}

The next result is the one-dimensional specialization of the empirical CDF concentration bound of \citet[Lemma~23]{guo2021generalizing}.

\begin{lemma}[Empirical Bernstein bound for CDFs]\label{lem:empirical-bernstein-cdf}
Let \(P\) be a probability distribution, let \(X_1,\ldots,X_m\) be independent samples from \(P\), and let \(\widehat F_m\) denote their empirical CDF. For every \(\delta\in(0,1)\), with probability at least \(1-\delta\), simultaneously for every \(x\in\mathbb R\),
\[
\left|\widehat F_m(x)-F_P(x)\right|\le\sqrt{\frac{2F_P(x)(1-F_P(x))\log(2m/\delta)}{m}}+\frac{\log(2m/\delta)}{m}.
\]
\end{lemma}

Finally, we record the one-dimensional Hellinger estimate underlying the Bernstein shading construction.

\begin{lemma}[Hellinger distance under Bernstein shading]\label{lem:hellinger-shading}
Let \(P\) be a probability distribution supported on \([0,1]\), let \(\gamma>0\), and let \(P^\gamma\) be the distribution with CDF
\[
F_{P^\gamma}(x):=\begin{cases}
0, & x<0,\\
\max\left\{0,F_P(x)-\sqrt{2\gamma F_P(x)(1-F_P(x))}-\gamma\right\}, & 0\le x<1,\\
1, & x\ge1.
\end{cases}
\]
There is an absolute constant \(C>0\) such that
\[
H^2(P,P^\gamma)\le C\min\left\{1,\gamma\log\left(e+\frac1\gamma\right)\right\}.
\]
\end{lemma}

\begin{proof}
For sufficiently small \(\gamma\), this is the coordinate-wise estimate established in the proofs of Lemmas~25 and~26 of \citet{guo2021generalizing}. For the remaining values of \(\gamma\), the result follows from the universal bound \(H^2(P,P^\gamma)\le1\), after increasing the absolute constant.
\end{proof}

%% file: related-work.tex
\subsection{Further Related Work}

Stochastic optimization under known distributions has been studied extensively, beginning with classical optimal-stopping problems such as prophet inequality and Pandora's box \citep{samuelcahn1984comparison,weitzman1979optimal}. Approximation algorithms are also known for stochastic combinatorial problems including probing, knapsack, orienteering, and submodular maximization \citep{gupta2013probing,dean2008stochastic,gupta2015orienteering,ma2018improvements,asadpour2016maximizing}. Our focus is instead on learning the unknown input distributions through repeated interaction.

The closest prior work is the approximation-preserving semi-bandit framework of \citet{agarwal2026semibandit} for monotone stochastic optimization. Under oracle-compatible strong monotonicity, we improve its linear dimension dependence to the optimal \(\widetilde O(G\sqrt{nT})\) rate; \Cref{thm:agn-continuous-lower-bound} shows that this improvement requires a different confidence construction. Under monotonicity, our algorithm obtains expected $\alpha$-regret $\widetilde O((1+V_0)n^{3/4}\sqrt T)$ for problems with objectives in $[0,1]$, provided that the $\alpha$-approximation oracle's output policies have uniform separate variation at most $V_0$. This condition requires no finite threshold representation.

For specific sequential-decision problems, \citet{gatmiry2024bandit} give \(\widetilde O(\operatorname{poly}(n)\sqrt T)\) regret algorithms for prophet inequality and Pandora's box under bandit feedback. 
\citet{singla2024bandit} study single-item sequential posted pricing under the same feedback model, obtaining \(\widetilde O(\operatorname{poly}(n)\sqrt T)\) regret for regular buyer distributions and \(\widetilde O(\operatorname{poly}(n)T^{2/3})\) regret for general distributions. Under semi-bandit feedback, \citet{liu2025improved} obtain the sharper \(\widetilde O(\sqrt{nT})\) rate for prophet inequality and Pandora's box, as well as contextual linear extensions. More recently, \citet{zhuo2026fullybandit} obtain nearly tight \(\widetilde O(\sqrt T)\) guarantees for ordered MDPs under fully bandit feedback, where the learner observes only the aggregate reward and not the realized trajectory. Their applications include \(k\)-item prophet inequality and sequential posted pricing, with an exponential dependence on \(k\) that they show is unavoidable in general.

A related offline literature studies the number of independent samples required to learn near-optimal policies. \citet{guo2021generalizing} give nearly tight \(\widetilde\Theta(n/\epsilon^2)\) sample complexity under strong monotonicity; the lower bound originates in single-parameter revenue maximization \citep{guo2019settling}. Our strongly monotone upper bound can be viewed as an adaptive semi-bandit analogue: the observation count of each marginal is selected by the policies executed by the learner rather than fixed in advance.

Combinatorial multi-armed bandits provide another general framework for learning from semi-bandit feedback \citep{kveton2015tight,chen2016generalreward}. In the non-triggering setting, the learner selects a set of arms and observes their individual outcomes. For objectives that depend on the full marginal distributions, \citet{chen2016generalreward} use stochastically dominant confidence bounds to obtain $\widetilde O(n\sqrt T)$ approximation regret. For objectives determined by arm means, \citet{pmlr-v99-merlis19a} introduce Gini-weighted smoothness and obtain bounds without polynomial dependence on the number of selected arms. Their results already give $\widetilde O(\sqrt{nT})$ regret for Bernoulli Best-of-$K$ selection. Our Best-of-$K$ comparison with \citet{chen2016generalreward} concerns arbitrary independent marginals on $[0,1]$.

The setting of combinatorial bandits with probabilistically triggered arms allows observations to depend on realized outcomes. \citet{wang2017triggered} introduce triggering probability modulated (TPM) smoothness, which weights estimation errors by observation probabilities. \citet{liu2022batch} combine variance-normalized quadratic and linear error terms through triggering probability and variance modulated (TPVM) smoothness to reduce batch-size dependence. These guarantees assume that expected rewards are determined by arm means and satisfy the corresponding smoothness conditions. Their triggering-weighted comparisons can be considered an analogue of \Cref{lem:stability}, which compares full marginal distributions in Hellinger distance for arbitrary adaptive policies without additional reward-smoothness assumptions.

%% file: strongly-monotone.tex
\section{Strongly Monotone Stochastic Optimization}
\label{sec:strongly-monotone}

In this section, we prove our main regret guarantee for strongly monotone problems. We first consider a maximization problem equipped with an exact oracle witnessing strong up-monotonicity. We discuss approximation guarantees and the other monotonicity and objective directions in subsequent subsections.

\begin{theorem}\label{thm:main}
Let \(\cP\) be a maximization problem with \(n\) items and objective range \(G\), and let \(\mathcal O\) be an exact optimization oracle. If \((\cP,\mathcal O)\) has oracle-compatible strong up-monotonicity, then there is an oracle-efficient semi-bandit learning algorithm using \(\mathcal O\) whose expected regret satisfies
\[
\EE[R(T)]=O\!\left(G\sqrt{nT}\,\log^{3/2}(2nT)\right).
\]
\end{theorem}

Before delving into the details, we provide a brief overview of the main argument.
In each round \(t \in [T]\), the algorithm constructs an optimistic product distribution \(\widetilde{\mathbf E}^{\,t}\) by shifting probability mass in each empirical marginal toward larger values, with the shift calibrated to a Bernstein-type confidence radius based on the number of observations of that item.
It then sets \(\sigma^t:=\mathcal O(\widetilde{\mathbf E}^{\,t})\) and executes this policy.
For the analysis, we show that on a suitable high-probability event, there is an auxiliary product distribution \(\mathbf U^t\) such that
\[
    \mathbf D
    \sd
    \widetilde{\mathbf E}^{\,t}
    \sd
    \mathbf U^t,
\]
where \(\mathbf U^t\) is simultaneously close to the true distribution \(\mathbf D\) and stochastically dominates $\widetilde{\mathbf E}^{\,t}$.
On the sandwich event, strong monotonicity gives
\[
\begin{aligned}
    \OPT(\mathbf D)-R(\sigma^t\mid\mathbf D)
    &\le
    \OPT(\widetilde{\mathbf E}^{\,t})
        -R(\sigma^t\mid\mathbf D)
    \le
    R(\sigma^t\mid\mathbf U^t)
        -R(\sigma^t\mid\mathbf D).
\end{aligned}
\]
Thus, the one-round regret is bounded by the change in value of the same executed policy under two nearby product distributions. 
Our \emph{stability lemma} compares the transcript laws induced by this policy and, through a sequential Hellinger chain rule, charges the squared marginal distance of item \(i\) only with the probability that \(i\) is probed, resulting in an \(\ell_2\)-type accumulation of the marginal errors.
Summing these bounds and charging them to the corresponding observation counts yields the theorem. 

\subsection{Constructing Optimistic Distributions}
\label{subsec:optimistic-distributions}
We first isolate the sampling result underlying our learning algorithm.
We begin by setting up some notation.
Fix a confidence parameter \(\delta\in(0,1)\), and define $\lambda := \log\left(\frac{2nT^2}{\delta}\right)$.
For every \(i\in[n]\), let \(X_{i,1},\ldots,X_{i,T}\) be independent samples from \(D_i\).  
For \(m\in[T]\), let \(\widehat D_{i,m}\) be the empirical distribution of \(X_{i,1},\ldots,X_{i,m}\), and let \(\widehat F_{i,m}\) denote its CDF.
We couple \(X_{i,m}\) with the value revealed on the \(m\)th occasion that item \(i\) is probed; this is valid because an item is selected before its current realization is observed and realizations are independent across items and rounds.

We construct an optimistic version of each empirical marginal using the
following Bernstein adjustment.  For \(\gamma>0\) and \(u\in[0,1]\), define
\begin{equation}
\label{eq:shade-scalar}
    s_\gamma(u)
    :=
    \max\left\{
        0,\,
        u-\sqrt{2\gamma u(1-u)}-\gamma
    \right\}.
\end{equation}
The map \(s_\gamma\) lowers the CDF and therefore shifts probability mass
toward larger values.  We refer to this operation as \emph{shading}, and use
\(\mathsf{Sh}\) as shorthand for the resulting shaded distribution.
Concretely, for a distribution \(P\) supported on \([0,1]\), let
\(\mathsf{Sh}_\gamma(P)\) be the distribution with CDF
\begin{equation}
\label{eq:shade-cdf}
    F_{\mathsf{Sh}_\gamma(P)}(x)
    :=
    \begin{cases}
        0, & x<0,\\
        s_\gamma(F_P(x)), & 0\le x<1,\\
        1, & x\ge1.
    \end{cases}
\end{equation}

The following lemma verifies that shading produces a valid optimistic
distribution.

\begin{lemma}
\label{lem:shading-validity}
For every \(\gamma>0\), \(s_\gamma\) is nondecreasing on \([0,1]\), and
\eqref{eq:shade-cdf} defines a valid probability distribution.  Moreover, $P\sd\mathsf{Sh}_\gamma(P).$
\end{lemma}

\begin{proof}
Let $g_\gamma(u) := u-\sqrt{2\gamma u(1-u)}-\gamma$.
Whenever \(g_\gamma(u)>0\), we have \(u>\gamma\).  If \(u\ge1/2\), then \(g_\gamma'(u)\ge1\).  If \(u<1/2\), the inequalities \(u>\gamma\) and $2u(1-u) > 2\gamma(1-u) > \gamma(1-2u)^2$ imply \(g_\gamma'(u)>0\).  
Hence \(s_\gamma=\max\{0,g_\gamma\}\) is nondecreasing.  
Since \(s_\gamma\) is continuous and nondecreasing and \(F_P\) is right-continuous, the endpoint conventions in \eqref{eq:shade-cdf} make \(F_{\mathsf{Sh}_\gamma(P)}\) a valid CDF.  
Moreover, \(s_\gamma(u)\le u\) for every \(u\in[0,1]\), and therefore \(F_{\mathsf{Sh}_\gamma(P)}(x)\le F_P(x)\) for every $x$, which implies \(P\sd\mathsf{Sh}_\gamma(P)\).
\end{proof}

We construct the optimistic distribution as follows. For \(m\in[T]\), define $ \eta_m:=\frac{\lambda}{m},$ and $\widetilde E_{i,m} := \mathsf{Sh}_{4\eta_m}(\widehat D_{i,m})$.
The distribution \(\widetilde E_{i,m}\) is computable from the first \(m\) samples of item \(i\).  
For the analysis, we also introduce the auxiliary distribution $U_{i,m} := \mathsf{Sh}_{9\eta_m}(D_i).$
Unlike \(\widetilde E_{i,m}\), the distribution \(U_{i,m}\) is not known to
the learner.

The main sampling result is that the true, optimistic, and auxiliary
distributions are ordered stochastically, while the true and auxiliary
distributions are close in Hellinger distance.

\begin{theorem}\label{thm:optimistic-distributions}
There is an absolute constant \(C_H>0\) such that, with probability at least
\(1-\delta\), simultaneously for every \(i\in[n]\) and \(m\in[T]\),
\[
    D_i
    \sd
    \widetilde E_{i,m}
    \sd
    U_{i,m}.
\]
Moreover, for every \(i\in[n]\) and \(m\in[T]\),
\[
    H^2(D_i,U_{i,m})
    \le
    C_H
    \min\left\{
        1,\,
        \eta_m\log\left(e+\frac{1}{\eta_m}\right)
    \right\},
\]
where \(H\) denotes the Hellinger distance.
\end{theorem}

We will use the following composition property of the shading map in our proof.

\begin{lemma}
\label{lem:shading-composition}
For every \(a,b>0\) and \(u\in[0,1]\),
\begin{equation}
\label{eq:shading-composition}
    s_{(\sqrt a+\sqrt b)^2}(u)
    \le
    s_b\bigl(s_a(u)\bigr).
\end{equation}
\end{lemma}

\begin{proof}
Let \(v(u):=u(1-u)\) and
\(\kappa:=(\sqrt a+\sqrt b)^2\).  
Suppose first that \(s_a(u)=0\).  
Since \(\kappa=(\sqrt a+\sqrt b)^2\ge a\) and \(\gamma\mapsto s_\gamma(u)\) is nonincreasing, we have $0\le s_\kappa(u)\le s_a(u)=0.$
Therefore \(s_\kappa(u)=0=s_b(s_a(u))\).
Otherwise, writing $q:=s_a(u)=u-\sqrt{2av(u)}-a,$ and using that \(v\) is \(1\)-Lipschitz, we obtain
\[
v(q) \le v(u)+\sqrt{2av(u)}+a \le \left(\sqrt{v(u)}+\sqrt{2a}\right)^2.
\]
Multiplying by $2b$ and taking square-roots gives $\sqrt{2bv(q)} \le \sqrt{2bv(u)}+2\sqrt{ab}$.
It follows that
\[
q - \sqrt{2bv(q)}-b \ge  u - \sqrt{2\kappa v(u)}-\kappa.
\]
Taking positive parts proves \eqref{eq:shading-composition}.
\end{proof}

We are now ready to complete the proof of \Cref{thm:optimistic-distributions}.

\begin{proof}[Proof of \Cref{thm:optimistic-distributions}.]
We begin by establishing a uniform concentration event.
Apply \Cref{lem:empirical-bernstein-cdf} separately to each pair
\((i,m)\), with failure probability \(\delta/(nT)\), and take a union
bound over \(i\in[n]\) and \(m\in[T]\).  Since
\(    \log\left(\frac{2mnT}{\delta}\right)
    \le
    \log\left(\frac{2nT^2}{\delta}\right)
    =
    \lambda,
\)
with probability at least \(1-\delta\),
\begin{equation}
\label{eq:uniform-cdf-event}
    \left|
        \widehat F_{i,m}(x)-F_{D_i}(x)
    \right|
    \le
    \sqrt{
        2\eta_m F_{D_i}(x)\bigl(1-F_{D_i}(x)\bigr)
    }
    +\eta_m
\end{equation}
simultaneously for every \(i\in[n]\), \(m\in[T]\), and \(x\in[0,1]\).

We condition on this event and fix \(i,m\).
For \(x<0\), all three CDFs equal \(0\), while for \(x\ge1\), they all equal
\(1\).  It therefore suffices to fix \(x\in[0,1)\).
For ease of notation, we further define $p:=F_{D_i}(x),$ $\widehat p:=\widehat F_{i,m}(x),$ and $\eta:=\eta_m.$
Recall that $s_\gamma(u) = \max\left\{ 0,\, u - \sqrt{2\gamma u(1-u)}-\gamma \right\}$.
We first show that $s_{4\eta}(\widehat p)\le p$.
This is immediate if \(\widehat p\le p\).  
Otherwise, let \(d:=\widehat p-p>0\).  Since \(u\mapsto u(1-u)\) is \(1\)-Lipschitz,
\eqref{eq:uniform-cdf-event} gives
\begin{align*}
d \,\, &\le \,\, \sqrt{2\eta p(1-p)}+\eta 
\,\, \le \,\, \sqrt{2\eta\widehat p(1-\widehat p)} + \sqrt{2\eta d}+\eta \,\, \le \,\,
\sqrt{2\eta\widehat p(1-\widehat p)} + \frac d2+2\eta.
\end{align*}
Therefore, $d \le \sqrt{8\eta\widehat p(1-\widehat p)}+4\eta$, which implies \begin{equation}
\label{eq:shading-preserves-optimism}
    s_{4\eta}(\widehat p)\le p.
\end{equation}
Therefore, $F_{\widetilde E_{i,m}}(x) = s_{4\eta}(\widehat p) \le p = F_{D_i}(x).$
The other side of \eqref{eq:uniform-cdf-event} similarly gives $\widehat p\ge s_\eta(p)$.
By the monotonicity of \(s_{4\eta}\) and \Cref{lem:shading-composition}, we also have $s_{4\eta}(\widehat p) \ge s_{4\eta}(s_\eta(p)) \ge s_{9\eta}(p).$ 
Therefore, $F_{\widetilde E_{i,m}}(x) = s_{4\eta}(\widehat p) \ge s_{9\eta}(p) = F_{U_{i,m}}(x).$
Thus, $F_{D_i}(x) \ge F_{\widetilde E_{i,m}}(x) \ge F_{U_{i,m}}(x),$
which is equivalent to
$D_i\sd\widetilde E_{i,m}\sd U_{i,m}.$

Finally, since
\(U_{i,m}=\mathsf{Sh}_{9\eta_m}(D_i)\),
\Cref{lem:hellinger-shading} gives
\[
    H^2(D_i,U_{i,m})
    \le
    C_H
    \min\left\{
        1,\,
        \eta_m\log\left(e+\frac1{\eta_m}\right)
    \right\},
\]
after adjusting the absolute constant \(C_H\).
\end{proof}

\subsection{The Online Algorithm and Its Analysis}
\label{subsec:online-algorithm}

We now use the preceding construction to define the online algorithm.
For \(x\in[0,1]\), let \(\delta_x\) denote the point mass at \(x\).
For each item \(i \in [n]\), let $N_i^t := \sum_{s=1}^{t-1} \mathbf 1\{i\in S(\sigma^s,\xvec^s)\}$
denote the number of times item \(i\) has been observed before round \(t\).
At the beginning of the round, the learner constructs the optimistic marginal
\[
    \widetilde E_i^t
    :=
    \begin{cases}
        \delta_1, & N_i^t=0,\\[0.2em]
        \displaystyle
        \mathsf{Sh}_{4\lambda/N_i^t}
        \bigl(\widehat D_{i,N_i^t}\bigr),
        & N_i^t\ge1.
    \end{cases}
\]
and lets
\(
    \widetilde{\mathbf E}^t
    :=
    \prod_{i=1}^n\widetilde E_i^t.
\)
It then computes \(\sigma^t:=\mathcal O(\widetilde{\mathbf E}^{\,t})\), executes this policy, and adds the values of the probed items to their respective empirical distributions. 
The complete procedure is given in \Cref{alg:strong-up}.

\begin{algorithm}[!h]
\caption{The Online Algorithm}
\label{alg:strong-up}
\KwIn{Horizon \(T\), confidence parameter \(\delta\), and an offline oracle \(\mathcal O\)}

\Initialize{\(N_i^1\gets0\) for every \(i\in[n]\)}
\For{\(t=1,\ldots,T\)}{
    \ForEach{\(i\in[n]\)}{
        \textbf{if } \(N_i^t=0\) \textbf{ then } \(\widetilde E_i^t\gets\delta_1\) \\
        \textbf{else } \(\widetilde E_i^t
              \gets\widetilde E_{i,N_i^t}\) 
    }
    Set \(\widetilde{\mathbf E}^t   \gets\prod_{i=1}^n\widetilde E_i^t\)\;
    Set \(\sigma^t\gets\mathcal O(\widetilde{\mathbf E}^t)\)\;
    Execute \(\sigma^t\) on a fresh realization
    \(\xvec^t\sim\mathbf D\), and observe
    \(X_i^t\) for every \(i\in S(\sigma^t,\xvec^t)\)\;
    Add each observed \(X_i^t\) to the samples of item \(i\), and set
    \[
        N_i^{t+1}
        \gets
        N_i^t+\mathbf 1\{i\in S(\sigma^t,\xvec^t)\}
        \qquad\text{for every }i\in[n].
    \]
}
\end{algorithm}

To analyze the algorithm, we associate each optimistic marginal with a
second, slightly more optimistic distribution.  Define
\[
    U_i^t
    :=
    \begin{cases}
        \delta_1, & N_i^t=0,\\[0.2em]
        \displaystyle
        \mathsf{Sh}_{9\lambda/N_i^t}(D_i),
        & N_i^t\ge1,
    \end{cases}
    \qquad\text{and}\qquad
    \mathbf U^t:=\prod_{i=1}^n U_i^t.
\]
These distributions are used only in the analysis: unlike
\(\widetilde{\mathbf E}^t\), they depend on the unknown true marginals.
On the event in \Cref{thm:optimistic-distributions}, the definitions and
the fact that every \(D_i\) is supported on \([0,1]\) give
\begin{equation}
\label{eq:online-sandwich}
    \mathbf D
    \sd
    \widetilde{\mathbf E}^t
    \sd
    \mathbf U^t
    \qquad\text{for every }t\in[T].
\end{equation}
Moreover, if
\[
    a_0:=1,
    \qquad
    a_m
    :=
    C_H
    \min\left\{
        1,\,
        \frac{\lambda}{m}
        \log\left(e+\frac{m}{\lambda}\right)
    \right\}
    \quad\text{for }m\ge1,
\]
then
\begin{equation}
\label{eq:online-hellinger}
    H^2(D_i,U_i^t)
    \le
    a_{N_i^t}
    \qquad
    \text{for every }i\in[n]\text{ and }t\in[T].
\end{equation}
Thus, \(\mathbf U^t\) provides an outer optimistic model whose distance
from the truth is controlled solely by the number of observations of each
item.

We next give the stability property that converts these marginal bounds
into a bound on the value of a policy.  For a product distribution
\(\mathbf P=\prod_{i=1}^n P_i\) and a policy \(\sigma\), let
\[
    \pi_i^{\mathbf P}(\sigma)
    :=
    \PP_{\xvec\sim\mathbf P}
    \bigl(i\in S(\sigma,\xvec)\bigr)
\]
denote the probability that \(\sigma\) probes item \(i\) under distribution $\bP$.  
Although two product distributions may differ in every coordinate, a policy can detect a difference in coordinate \(i\) only when it probes that item.  
The following \emph{stability lemma} formalizes this observation, including for adaptive policies.

\begin{lemma}[Stability lemma]
\label{lem:stability}
For every policy \(\sigma\) and product distributions
\(\mathbf P=\prod_{i=1}^nP_i\) and
\(\mathbf Q=\prod_{i=1}^nQ_i\),
\begin{equation}
\label{eq:stability}
    \left| R(\sigma\mid\mathbf P) - R(\sigma\mid\mathbf Q) \right| \,\, \le \,\, 2\sqrt{2}\,G \left( \sum_{i=1}^n \pi_i^{\mathbf P}(\sigma) \cdot  H^2(P_i,Q_i) \right)^{1/2}.
\end{equation}
\end{lemma}

The probing probabilities in \Cref{lem:stability} are measured under
\(\mathbf P\).  This asymmetry will be useful below: we apply the lemma
with \(\mathbf P=\mathbf D\), so that
\(\pi_i^{\mathbf D}(\sigma^t)\) is precisely the conditional probability
that item \(i\) is observed in round \(t\).  
We first use the stability lemma to complete the proof of the main theorem and prove the lemma afterward.

\begin{proof}[Proof of \Cref{thm:main}]
Let \(\mathcal G\) denote the event in
\Cref{thm:optimistic-distributions}, which has probability at least
\(1-\delta\).  On this event, \eqref{eq:online-sandwich} holds in every
round.  
Fix a round \(t\) and the history before that round. By strong up-monotonicity,
\[
\begin{aligned}
\OPT(\mathbf D)-R(\sigma^t\mid\mathbf D)
\le \OPT(\widetilde{\mathbf E}^t)-R(\sigma^t\mid\mathbf D) = R(\sigma^t\mid\widetilde{\mathbf E}^t)-R(\sigma^t\mid\mathbf D) \le R(\sigma^t\mid\mathbf U^t)-R(\sigma^t\mid\mathbf D).
\end{aligned}
\]
The last inequality follows from strong monotonicity because
\(\widetilde{\mathbf E}^t\sd\mathbf U^t\) and \(\sigma^t\) is optimal
under \(\widetilde{\mathbf E}^t\).

For ease of notation, write $\pi_i^t := \pi_i^{\mathbf D}(\sigma^t).$
Applying \Cref{lem:stability} with \(\mathbf P=\mathbf D\) and \(\mathbf Q=\mathbf U^t\), and then using \eqref{eq:online-hellinger}, gives
\[
    \OPT(\mathbf D)-R(\sigma^t\mid\mathbf D) \le R(\sigma^t\mid\mathbf U^t)
        - R(\sigma^t\mid\mathbf D) 
    \le
    2\sqrt{2}\,G
    \left(
        \sum_{i=1}^n
        \pi_i^t a_{N_i^t}
    \right)^{1/2}.
\]
Observe that $\OPT(\mathbf D)-R(\sigma^t\mid\mathbf D)$ is the regret suffered by the learning algorithm at time $t$. 
Consequently, on \(\mathcal G\), the Cauchy--Schwarz inequality yields
\begin{equation}\label{eq:regret-on-good-event}
    R(T) \le 2\sqrt{2}\,G \sqrt{ T \sum_{t=1}^T\sum_{i=1}^n \pi_i^t a_{N_i^t}}.
\end{equation}
It remains to bound the sum inside the square root.  Let $I_i^t := \mathbf 1\{i\in S(\sigma^t,\xvec^t)\}.$
Conditional on the complete history before round \(t\), the policy
\(\sigma^t\), the count \(N_i^t\), and hence \(a_{N_i^t}\), are fixed.
Since \(\xvec^t\) is a fresh draw from \(\mathbf D\), the conditional
probability that item \(i\) is probed in round \(t\) is \(\pi_i^t\).
Therefore, $\EE\left[I_i^t a_{N_i^t}\right] = \EE\left[\pi_i^t a_{N_i^t}\right].$
It follows that
\begin{equation}\label{eq:adaptive-counting}
\EE\left[ \sum_{t=1}^T\sum_{i=1}^n \pi_i^t a_{N_i^t} \right] = \EE\left[ \sum_{t=1}^T\sum_{i=1}^n I_i^t a_{N_i^t} \right] = \EE\left[\sum_{i=1}^n \sum_{m=0}^{N_i^{T+1}-1}a_m \right] \le n\sum_{m=0}^{T-1}a_m.
\end{equation}
The second equality holds pathwise: whenever item \(i\) is observed, its
count immediately before that observation takes the next successive value
\(0,1,\ldots,N_i^{T+1}-1\).

We next bound the deterministic sum in \eqref{eq:adaptive-counting}. Since \(a_m=O(1)\), the terms with \(m\le\lceil\lambda\rceil\) contribute \(O(\lambda)\). If \(T>\lambda\), comparison with an integral shows that the remaining terms contribute at most \(C_H\lambda\left(1+\int_\lambda^T\frac{\log(e+x/\lambda)}{x}\,\mathrm dx\right)=O\left(\lambda[1+\log^2(e+T/\lambda)]\right)\). Hence, in all cases,
\begin{equation}\label{eq:sum-am}
\sum_{m=0}^{T-1}a_m \le C\lambda\left[1+\log^2(e+T/\lambda)\right]
\end{equation}
for an absolute constant \(C>0\).

We now account separately for the good and bad events.  
Importantly, we do not condition the observation process on \(\mathcal G\), since doing so could change the conditional probing probabilities.  Instead, multiply \eqref{eq:regret-on-good-event} by \(\mathbf 1_{\mathcal G}\), drop this indicator from the nonnegative quantity inside the expectation, and use Jensen's inequality together with \eqref{eq:adaptive-counting} and \eqref{eq:sum-am}.   This gives
\[
\begin{aligned}
    \EE\bigl[R(T)\mathbf 1_{\mathcal G}\bigr]
    &\le 2\sqrt{2}\,G\sqrt{T}\,
    \EE\left[ \mathbf 1_{\mathcal G} \sqrt{\sum_{t=1}^T\sum_{i=1}^n \pi_i^t a_{N_i^t}}
    \right] \le C' G \sqrt{ nT\lambda \left[ 1 + \log^2\left(e+\frac{T}{\lambda}\right) \right] }
\end{aligned}
\]
for another absolute constant \(C'>0\).

Since the objective has range \(G\), the regret in any round is at most
\(G\), and hence \(0\le R(T)\le GT\).  Since
\(\PP(\mathcal G^c)\le\delta\), we have $\EE\bigl[R(T)\mathbf 1_{\mathcal G^c}\bigr] \le \delta GT.$ Combining the two events, we obtain
\[ \EE[R(T)] \le C' G \sqrt{ nT\lambda \left[1 +
\log^2\left(e+\frac{T}{\lambda}\right) \right]} +  \delta GT.\]
Finally, take \(\delta=1/(T+1)\).  Then
\( \lambda = \log\!\bigl(2nT^2(T+1)\bigr) = O\!\left(\log(2nT)\right),\)
and
\( \log\!\left(e+\frac{T}{\lambda}\right) = O\!\left(\log(2nT)\right). \)
Since \(\delta GT\le G\), we get $\EE[R(T)] = O( G\sqrt{nT}\, \log^{3/2}(2nT))$.
\end{proof}

We conclude this subsection by proving the stability lemma. The main observation is that an adaptive policy can distinguish two product distributions only through the coordinates it actually probes. We formalize this by decomposing the Hellinger distance between the transcript distributions over successive probe positions.

\begin{proof}[Proof of \Cref{lem:stability}]
Fix a policy \(\sigma\), and let \(\mathsf T_{\mathbf P}^{\sigma}\) and \(\mathsf T_{\mathbf Q}^{\sigma}\) denote the distributions of its transcript under \(\mathbf P\) and \(\mathbf Q\), respectively. We view a transcript as a sequence \(Z=(Z_1,\ldots,Z_n)\), where \(Z_k\) records the item and value observed in the \(k\)th probe, and all entries after the policy stops equal a dummy symbol \(\perp\).

For a transcript prefix
\(z_{<k}=(z_1,\ldots,z_{k-1})\), let
\(
    A_k^\sigma(z_{<k})
    \in
    [n]\cup\{\perp\}
\)
denote the item selected for the \(k\)th probe, with
\(A_k^\sigma(z_{<k})=\perp\) if the policy makes no \(k\)th probe after
this prefix.

We take versions of the next-entry conditional laws induced by the policy
execution as follows.  If \(A_k^\sigma(z_{<k})=i\), then under
\(\mathbf P\) the conditional law of \(Z_k\) is the law of
\((i,X_i)\) with \(X_i\sim P_i\), while under \(\mathbf Q\) it is the law
of \((i,X_i)\) with \(X_i\sim Q_i\).  This follows from product
independence and the fact that item \(i\) has not previously been probed.
If \(A_k^\sigma(z_{<k})=\perp\), both conditional laws are the point mass
at \(\perp\).  Hence
\begin{equation}
\label{eq:conditional-transcript-hellinger}
\begin{aligned}
&H^2\left(
    \operatorname{Law}_{\mathbf P}(Z_k\mid Z_{<k}=z_{<k}),
    \operatorname{Law}_{\mathbf Q}(Z_k\mid Z_{<k}=z_{<k})
\right)=
\sum_{i=1}^n
\mathbf 1\{A_k^\sigma(z_{<k})=i\}
H^2(P_i,Q_i).
\end{aligned}
\end{equation}
Applying \Cref{lem:hellinger-chain-rule} and then interchanging the sums over probe positions and items gives
\begin{align}
H^2\left(\mathsf T_{\mathbf P}^{\sigma},\mathsf T_{\mathbf Q}^{\sigma}\right)
&\le 4\sum_{k=1}^n\sum_{i=1}^n\PP_{\xvec\sim\mathbf P}\left(A_k^\sigma(Z_{<k})=i\right)H^2(P_i,Q_i)\notag\\
&=4\sum_{i=1}^nH^2(P_i,Q_i)\sum_{k=1}^n\PP_{\xvec\sim\mathbf P}\left(A_k^\sigma(Z_{<k})=i\right).\label{eq:transcript-probe-sum}
\end{align}
Because each item is probed at most once,
$\sum_{k=1}^n\mathbf 1\{A_k^\sigma(Z_{<k})=i\}=\mathbf 1\{i\in S(\sigma,\xvec)\}$
for every realization \(\xvec\). Taking expectations in this identity and substituting into \eqref{eq:transcript-probe-sum} yields
\begin{equation}\label{eq:transcript-hellinger}
H^2\left(\mathsf T_{\mathbf P}^{\sigma},\mathsf T_{\mathbf Q}^{\sigma}\right)\le 4\sum_{i=1}^n\pi_i^{\mathbf P}(\sigma)H^2(P_i,Q_i).
\end{equation}
Finally, because the realized reward depends only on the observed transcript, there is a measurable function \(\rho_\sigma\), taking values in an interval of length \(G\), such that \(r(\sigma,\xvec)=\rho_\sigma(\tau_\sigma(\xvec))\). Hence,
\[
R(\sigma\mid\mathbf P)=\int \rho_\sigma\,\mathrm d\mathsf T_{\mathbf P}^{\sigma}
\qquad\text{and}\qquad
R(\sigma\mid\mathbf Q)=\int \rho_\sigma\,\mathrm d\mathsf T_{\mathbf Q}^{\sigma}.
\]
Therefore, by \Cref{eq:tv-hellinger},
\[
\left|R(\sigma\mid\mathbf P)-R(\sigma\mid\mathbf Q)\right|
\le
G\,d_{\mathrm{TV}}\left(
    \mathsf T_{\mathbf P}^{\sigma},
    \mathsf T_{\mathbf Q}^{\sigma}
\right)
\le
\sqrt{2}\,G\,H\left(
    \mathsf T_{\mathbf P}^{\sigma},
    \mathsf T_{\mathbf Q}^{\sigma}
\right).
\]
Combining this inequality with \eqref{eq:transcript-hellinger} proves \eqref{eq:stability}.
\end{proof}

\subsection{Extending the Framework to Approximate Offline Oracles}
\label{subsec:approximate-oracles}

The proof of \Cref{thm:main} uses exact optimality only in the one-round comparison. The same comparison extends to an \(\alpha\)-approximation oracle when the problem is monotone and the policy selected by the oracle satisfies the corresponding transfer property. This is precisely the oracle-compatible strong monotonicity condition in \Cref{def:strong-monotone}; an approximation guarantee under the oracle's input distribution alone does not imply it.

The following criterion verifies oracle compatibility directly from realized objectives.

\begin{lemma}[Black-box transfer from pointwise monotonicity.]
\label{lem:nonadaptive-transfer}
Let \(\mathcal N\subseteq\cC\) be a family of feasible policies, and let \(\mathcal O\) be a possibly randomized \(\alpha\)-approximation oracle taking values in \(\mathcal N\). 
Suppose that \(\cP\) is up-monotone and, for every \(\sigma\in\mathcal N\) and every pair of realizations \(\xvec\le\mathbf y\), the realized objective of \(\sigma\) under \(\mathbf y\) is at least as good as under \(\xvec\). Then \((\cP,\mathcal O)\) has oracle-compatible strong up-monotonicity. Analogously, if \(\cP\) is down-monotone and the realized objective under \(\xvec\) is at least as good as under \(\mathbf y\), then \((\cP,\mathcal O)\) has oracle-compatible strong down-monotonicity.
\end{lemma}

\begin{proof}
Consider first up-monotone maximization. For product distributions \(\mathbf P\sd\mathbf Q\), independent coordinatewise monotone couplings give vectors \(\mathbf X\sim\mathbf P\) and \(\mathbf Y\sim\mathbf Q\) with \(\mathbf X\le\mathbf Y\) almost surely. Thus, for every fixed \(\sigma\in\mathcal N\),
\[
R(\sigma\mid\mathbf P)=\EE[r(\sigma,\mathbf X)]\le\EE[r(\sigma,\mathbf Y)]=R(\sigma\mid\mathbf Q).
\]
For each realization of the oracle's internal randomness, apply this inequality to \(\sigma=\mathcal O(\mathbf P;\xi)\), and then average over \(\xi\). This gives the required transfer comparison. In the down-monotone case, the assumed comparison of realized rewards gives the reversed inequality, which we apply to \(\sigma=\mathcal O(\mathbf Q;\xi)\). For minimization, replace \(r,R\) by \(c,J\) and reverse the numerical inequalities. Together with the assumed monotonicity of the optimal value, these comparisons establish oracle compatibility.
\end{proof}

The criterion applies to both adaptive and nonadaptive policies. In our applications, we verify pointwise monotonicity for nonadaptive policies specified by fixed sets or orderings. Nonadaptivity alone guarantees neither this property nor monotonicity of the optimal value.

\begin{theorem}
\label{thm:approximate-oracle}
Let \(\cP\) be a maximization problem with \(n\) items and a nonnegative objective of range \(G\), and let \(\mathcal O\) be a possibly randomized \(\alpha\)-approximation oracle for \(\cP\), where \(\alpha\in(0,1]\). If \((\cP,\mathcal O)\) has oracle-compatible strong up-monotonicity, then the online algorithm in \Cref{alg:strong-up}, using \(\mathcal O\) and setting \(\delta=1/(T+1)\), satisfies
\[
\EE[\alpha\text{-}R(T)]\le C G\sqrt{nT}\,\log^{3/2}(2nT),
\]
where \(C>0\) is an absolute constant.
\end{theorem}

\begin{proof}
Let \(\mathcal H_{t-1}\) denote the information available before the oracle call in round \(t\), and write \(\EE_t[\cdot]:=\EE[\cdot\mid\mathcal H_{t-1}]\). The algorithm selects \(\sigma^t:=\mathcal O(\widetilde{\mathbf E}^{\,t};\xi^t)\), where \(\xi^t\) is fresh internal randomness. Let \(\mathcal G_t\) be the event \(\mathbf D\sd\widetilde{\mathbf E}^{\,t}\sd\mathbf U^t\). This event is \(\mathcal H_{t-1}\)-measurable, and \Cref{thm:optimistic-distributions} together with the initialization at unobserved items gives \(\PP(\mathcal G_t^c)\le\delta\).

Write \(\Delta_t:=\alpha\OPT(\mathbf D)-\EE_t[R(\sigma^t\mid\mathbf D)]\) and \(\bar\pi_i^t:=\EE_t[\pi_i^{\mathbf D}(\sigma^t)]\). On \(\mathcal G_t\), monotonicity of the optimal value, the expected approximation guarantee, and oracle compatibility give
\[
\begin{aligned}
\Delta_t
&\le \alpha\OPT(\widetilde{\mathbf E}^{\,t})-\EE_t[R(\sigma^t\mid\mathbf D)] \le \EE_t\!\left[R(\sigma^t\mid\widetilde{\mathbf E}^{\,t})-R(\sigma^t\mid\mathbf D)\right]\\
&\le \EE_t\!\left[R(\sigma^t\mid\mathbf U^t)-R(\sigma^t\mid\mathbf D)\right]\le 2\sqrt{2}\,G\left(\sum_{i=1}^n\bar\pi_i^t a_{N_i^t}\right)^{1/2}.
\end{aligned}
\]
The last inequality follows by applying \Cref{lem:stability} to each sampled deterministic policy, using \eqref{eq:online-hellinger}, and then applying Jensen's inequality.

Let \(I_i^t:=\mathbf 1\{i\in S(\sigma^t,\xvec^t)\}\). Since the current outcome vector is independent of the history and the oracle's fresh randomness, \(\EE_t[I_i^t]=\bar\pi_i^t\). Consequently, the same counting argument as in \eqref{eq:adaptive-counting} gives
\[
\EE\!\left[\sum_{t=1}^T\sum_{i=1}^n\bar\pi_i^t a_{N_i^t}\right]
=\EE\!\left[\sum_{i=1}^n\sum_{m=0}^{N_i^{T+1}-1}a_m\right]
\le n\sum_{m=0}^{T-1}a_m.
\]
Moreover, nonnegativity of the objective and \(\alpha\le1\) imply \(\Delta_t\le G\) for every history. Bounding separately on \(\mathcal G_t\) and its complement, and applying Cauchy--Schwarz and Jensen's inequality, therefore yields
\[
\EE[\alpha\text{-}R(T)]
=\sum_{t=1}^T\EE[\Delta_t]
\le 2\sqrt{2}\,G\sqrt{nT\sum_{m=0}^{T-1}a_m}+\delta GT.
\]
The result follows from \eqref{eq:sum-am} and \(\delta=1/(T+1)\).
\end{proof}

\subsection{Strongly Down-Monotone Maximization}
The preceding analysis assumed an exact oracle witnessing strong up-monotonicity. For an oracle witnessing strong down-monotonicity, the optimistic distribution must instead be stochastically smaller than the true distribution. We obtain such a distribution by applying the same shading construction with the order on \([0,1]\) reversed; equivalently, apply the construction to the samples \(1-X_{i,s}\) and map the resulting distribution back to \([0,1]\).

The symmetric version of \Cref{thm:optimistic-distributions} gives a distribution \(\widetilde E_{i,m}\) and an auxiliary distribution \(L_{i,m}\) such that, with probability at least \(1-\delta\),
$L_{i,m}\sd\widetilde E_{i,m}\sd D_i,$
where \(H^2(D_i,L_{i,m})\) satisfies the same bound as \(H^2(D_i,U_{i,m})\).  When item \(i\) has not yet been observed, we use the point mass at \(0\). The main result (formalizing \Cref{thm:main-informal} for strongly down-monotone maximization problems) is as follows. 

\begin{theorem}
\label{thm:strong-down-max}
Let \(\cP\) be a maximization problem with \(n\) items and objective range \(G\), and let \(\mathcal O\) be an exact optimization oracle. If \((\cP,\mathcal O)\) has oracle-compatible strong down-monotonicity, then there is an oracle-efficient semi-bandit learning algorithm using \(\mathcal O\) such that
\[
\EE[R(T)]=O\left(G\sqrt{nT}\,\log^{3/2}(2nT)\right).
\]
\end{theorem}

\begin{proof}
In each round, the algorithm executes \(\sigma^t:=\mathcal O(\widetilde{\mathbf E}^{\,t})\). Let \(\mathbf L^t\) be the corresponding auxiliary distribution. On the good event, \(\mathbf L^t\sd\widetilde{\mathbf E}^{\,t}\sd\mathbf D\). 
By strong down-monotonicity, we get
\[
\OPT(\mathbf D)-R(\sigma^t\mid\mathbf D)
\le \OPT(\widetilde{\mathbf E}^{\,t})-R(\sigma^t\mid\mathbf D)
\le R(\sigma^t\mid\mathbf L^t)-R(\sigma^t\mid\mathbf D).
\]
The stability lemma, adaptive counting argument, and treatment of the good and bad events are now identical to the proof of \Cref{thm:main}.
\end{proof}

The same argument shows that the expected \(\alpha\)-regret is at most \(C G\sqrt{nT}\log^{3/2}(2nT)\) for an absolute constant \(C>0\), provided that \(\cP\) is down-monotone and \(R(\mathcal O(\mathbf Q)\mid\mathbf P)\ge R(\mathcal O(\mathbf Q)\mid\mathbf Q)\) whenever \(\mathbf P\sd\mathbf Q\).

\subsection{Minimization Problems}
\label{subsec:strong-minimization}

We finally describe the changes for minimization problems.  The algorithm and the construction of the optimistic distributions are unchanged: we use the original shading construction for strongly up-monotone problems and the reverse-shading construction for strongly down-monotone problems.

\begin{theorem}
\label{thm:strong-min}
Let \(\cP\) be a minimization problem with \(n\) items and objective range \(G\), and let \(\mathcal O\) be an exact optimization oracle. If \((\cP,\mathcal O)\) has oracle-compatible strong up- or down-monotonicity, then there is an oracle-efficient semi-bandit learning algorithm using \(\mathcal O\) such that
\[
\EE[R(T)]=O\left(G\sqrt{nT}\,\log^{3/2}(2nT)\right).
\]
\end{theorem}

\begin{proof}
Suppose first that \((\cP,\mathcal O)\) has oracle-compatible strong up-monotonicity, and set \(\sigma^t:=\mathcal O(\widetilde{\mathbf E}^{\,t})\).  On the good event, \(\mathbf D\sd\widetilde{\mathbf E}^{\,t}\sd\mathbf U^t\).  
By strong up-monotonicity,
\[
J(\sigma^t\mid\mathbf D)-\OPT(\mathbf D)
\le J(\sigma^t\mid\mathbf D)-\OPT(\widetilde{\mathbf E}^{\,t})
\le J(\sigma^t\mid\mathbf D)-J(\sigma^t\mid\mathbf U^t).
\]
The stability lemma applies with \(J\) in place of \(R\), since its proof uses only the range of the realized objective.  The remainder of the analysis is therefore unchanged.

If \((\cP,\mathcal O)\) has oracle-compatible strong down-monotonicity, the algorithm instead uses the reverse-shaded distributions from the preceding subsection.  On the good event, \(\mathbf L^t\sd\widetilde{\mathbf E}^{\,t}\sd\mathbf D\), and the corresponding comparison is
\[
J(\sigma^t\mid\mathbf D)-\OPT(\mathbf D)
\le J(\sigma^t\mid\mathbf D)-\OPT(\widetilde{\mathbf E}^{\,t})
\le J(\sigma^t\mid\mathbf D)-J(\sigma^t\mid\mathbf L^t).
\]
Applying the same stability and counting arguments proves the theorem.
\end{proof}

The approximation-oracle guarantee extends analogously for \(\alpha\ge1\): it suffices that \(\cP\) be monotone in the relevant direction and, whenever \(\mathbf P\sd\mathbf Q\), that \(J(\mathcal O(\mathbf P)\mid\mathbf Q)\le J(\mathcal O(\mathbf P)\mid\mathbf P)\) in the up-monotone case or \(J(\mathcal O(\mathbf Q)\mid\mathbf P)\le J(\mathcal O(\mathbf Q)\mid\mathbf Q)\) in the down-monotone case.

\subsection{A Limitation of Uniform CDF Shifts} \label{subsec:agn-lower-bound}

The continuous-distribution algorithm of \citet{agarwal2026semibandit} constructs an optimistic distribution by applying a uniform downward shift to the empirical CDF. Given \(m\) observations with empirical CDF \(\widehat F_m\), it forms a distribution on \([0,1]\) with CDF
\begin{equation}
\label{eq:agn-uniform-shift}
\widetilde F_m(x)=
\begin{cases}
0,&x<0,\\
\max\{0,\widehat F_m(x)-\epsilon_m\},&0\le x<1,\\
1,&x\ge1,
\end{cases}
\qquad
\epsilon_m=\sqrt{\frac{\log(2/\delta_0)}{2m}}.
\end{equation}
We show that this uniform shift cannot yield the \(\widetilde O(\sqrt{nT})\) rate even for a strongly monotone problem with a canonical exact oracle.

Consider fixed-order single-choice stopping (that is, the classic optimal stopping problem): rewards arrive in the order \(1,\ldots,n\), and a policy may accept at most one reward, terminating immediately upon acceptance. For a product distribution \(\mathbf P\), let the canonical exact oracle be given by
\[
V_{n+1}^{\mathbf P}=0,\qquad V_i^{\mathbf P}=\EE_{Y_i\sim P_i}\!\left[\max\{Y_i,V_{i+1}^{\mathbf P}\}\right],
\]
with the policy accepting \(Y_i\) if and only if \(Y_i\ge V_{i+1}^{\mathbf P}\). This is a two-threshold policy and witnesses strong up-monotonicity \citep[Lemma~28]{guo2021generalizing}.

\begin{theorem}
\label{thm:agn-continuous-lower-bound}
For every \(n\ge2\), consider fixed-order single-choice stopping with independent rewards
\[
X_1\sim\operatorname{Unif}[0,1],\qquad X_i\sim\operatorname{Beta}(1,16n)\quad(2\le i\le n).
\]
Run the uniform-shift algorithm of \citet{agarwal2026semibandit}, using the canonical exact oracle, for \(L=n^2\) online rounds after \(n\) initialization rounds, with \(\delta_0=2/(nL)^3\). For the total horizon \(T:=L+n=n^2+n\),
\[
\EE[R(T)]\ge\frac{n^2}{16}.
\]
\end{theorem}

\begin{proof}
In the initialization phase, we observe a sample for every item through the feasible policy that rejects the first \(n-1\) rewards and accepts the last; all these observations are retained. 
Since \(L=n^2\), we have \(\delta_0 = 2/n^9\).
Since every true observation is strictly below \(1\) almost surely, \eqref{eq:agn-uniform-shift} gives optimistic marginal \(i\) an atom at \(1\) of mass \(q_i^t=\min\{1,\epsilon_{N_i^t}\}\), where \(N_i^t\) is its number of observations before round \(t\). Because \(N_i^t\le T\le2n^2\), the confidence radius satisfies
\[
q_i^t\ge\min\left\{1,\sqrt{\frac{9\log n}{4n^2}}\right\}\ge\frac1n.
\]
Under the optimistic product distribution, the feasible policy that waits for a reward equal to \(1\) among items \(2,\ldots,n\) has value at least
\[
1-\prod_{i=2}^n(1-q_i^t)\ge1-\left(1-\frac1n\right)^{n-1}\ge\frac12.
\]
Thus the canonical oracle's acceptance threshold for item \(1\) is at least \(1/2\) in every online round.

Conditional on the preceding history, the expected reward of the resulting policy under the true distribution is at most
\[
\EE\!\left[X_1\mathbf1\{X_1\ge1/2\}\right]+\EE\!\left[\max_{i\ge2}X_i\right]
\le\int_{1/2}^1x\,dx+\sum_{i=2}^n\EE[X_i]
=\frac38+\frac{n-1}{16n+1}<\frac7{16}.
\]
The true optimal value is at least \(\EE[X_1]=1/2\), since accepting the first reward is feasible. Hence every one of the \(n^2\) online rounds contributes at least \(1/16\) in conditional expected regret. The initialization rounds contribute nonnegative regret, proving the result.
\end{proof}

The instance satisfies the regularity assumptions imposed by the continuous-distribution analysis of \citet{agarwal2026semibandit}: every probe has two branches, every leaf reward is either zero or the accepted coordinate and has coordinatewise variation at most one, and all marginals have positive density on \((0,1)\).
Since \(n^2+n=\Theta(n^2)\), \Cref{thm:agn-continuous-lower-bound} gives \(\Omega(n\sqrt T)\) regret and therefore rules out a uniform \(\widetilde O(\sqrt{nT})\) guarantee for this algorithm, even on strongly monotone problems.

The obstruction is the endpoint mass introduced by the uniform shift: after \(m\) observations it is \(\min\{1,\epsilon_m\}\). When the empirical marginal has no atom at \(1\), our Bernstein shading instead adds \(\min\{1,4\lambda/m\}\), where \(\lambda\) is defined in \Cref{subsec:optimistic-distributions}. At the same horizon, \Cref{thm:main} therefore gives expected regret \(O(n^{3/2}\log^{3/2}n)=o(n^2)\).

%% file: weakly-monotone.tex
\section{Weakly Monotone Stochastic Optimization}\label{sec:weak-monotone}

We now study what our optimistic algorithm guarantees for weakly monotone stochastic optimization when the oracle need not satisfy the strong-monotonicity transfer property. Weak monotonicity alone is insufficient: even on a one-item instance, an exact oracle selection can lead to linear regret (\Cref{thm:weak-monotonicity-linear-regret}).

We obtain a sublinear guarantee by imposing a uniform separate-variation bound $V_0$ on the oracle's output policies, as defined in \eqref{eq:weak-bv-assumption}. For objectives in $[0,1]$, \Cref{thm:weak-bv-exact-exponent} gives expected $\alpha$-regret $\widetilde O((1+V_0)n^{3/4}\sqrt T)$ for any possibly randomized $\alpha$-approximation oracle satisfying this condition. For fixed $V_0>0$, the dimension exponent $3/4$ is sharp for ordinary-regret guarantees that hold uniformly over admissible problems and exact oracle selections of this algorithm.

\begin{remark}
We give the proofs for up-monotone maximization. The other cases follow by reversing the shading direction for down-monotone problems and applying the same absolute stability bound to costs, with the reversed oracle inequality, for minimization. For minimization, the variation definitions below use $c_\sigma(x):=c(\sigma,x)$ in place of $r_\sigma$. We omit the details of these straightforward extensions.
\end{remark}

Finally, the threshold-policy and leaf-variation assumptions of \citet{agarwal2026semibandit} imply our separate-variation condition. We extend their Kolmogorov--Smirnov stability analysis to this condition, allowing both regret bounds to be expressed in terms of the same parameter $V_0$. For fixed $V_0>0$, our analysis improves the polynomial dependence on $n$ from $n$ to $n^{3/4}$ (up to log factors).

\subsection{A Limitation of Monotonicity}
\label{subsec:weak-monotonicity-limitation}

For maximization, up-monotonicity requires only that $\mathbf P\sd\mathbf Q$ implies $\OPT(\mathbf P)\le\OPT(\mathbf Q)$. It imposes no transfer condition on the policy returned by an exact oracle. The following construction exploits this distinction: the oracle selects a policy that earns optimal reward on the finite support of the optimistic distribution but earns zero reward under the true distribution.

\begin{theorem} \label{thm:weak-monotonicity-linear-regret}
There exists a one-item maximization problem with a distribution-independent feasible policy class, objective range $G=1$, and a measurable exact oracle $\mathcal O$ defined on all probability distributions on $[0,1]$, such that $\OPT(P)=1$ for every distribution $P$. Nevertheless, under $D=\operatorname{Unif}[0,1]$, \Cref{alg:strong-up} using $\mathcal O$ satisfies $R(T)=T$ almost surely for every horizon $T\ge1$ and confidence parameter $\delta\in(0,1)$.
\end{theorem}

\begin{proof}
Let $\cC$ contain a policy $\sigma^\circ$ that stops immediately and receives reward $1$. For every nonempty finite set $S\subset[0,1]$, it also contains a policy $\sigma^S$ that probes the item and receives reward $r(\sigma^S,x)=\one\{x\in S\}.$
Each policy includes its specified terminal reward rule. These policies are deterministic and nonanticipating, and their rewards depend only on their observed transcripts. The class is fixed independently of the distribution and horizon. Rewards attain both $0$ and $1$, so $G=1$, and the guaranteed-reward policy gives
\begin{equation}\label{eq:weak-opt-constant}
\OPT(P)=1 \qquad\text{for every probability distribution }P\text{ on }[0,1].
\end{equation}
Thus the problem is up-monotone.

For a finitely supported distribution $P$, write $\operatorname{At}(P):=\{x:P(\{x\})>0\}$. Define the oracle by
\[
\mathcal O(P)=
\begin{cases}
\sigma^{\operatorname{At}(P)}, & P\text{ is finitely supported},\\
\sigma^\circ, & \text{otherwise}.
\end{cases}
\]
If $P$ is finitely supported, its atom set is finite and nonempty, and the selected policy earns expected reward $P(\operatorname{At}(P))=1$. Otherwise, the oracle selects the guaranteed-reward policy. Hence $\mathcal O$ is exact on every input distribution.

Now run \Cref{alg:strong-up} under $D=\operatorname{Unif}[0,1]$. Initially, the optimistic distribution is $\widetilde E^1=\delta_1$, so the oracle selects $\sigma^{\{1\}}$. Every subsequent optimistic distribution is also finitely supported: shading changes the masses of the empirical distribution and can add only the endpoint $1$. Therefore, in every round, we select $\sigma^t=\sigma^{S_t}$ where $S_t:=\operatorname{At}(\widetilde E^t)$.
The algorithm probes the item in every round, and  $\emptyset \ne S_t\subseteq\{1\}\cup\{X^s:1\le s<t\}$.

Let $\mathcal H_{t-1}$ denote the observations before round $t$. The finite set $S_t$ is a measurable function of this history. Since the fresh observation $X^t$ is independent of the history and uniformly distributed on $[0,1]$, we have
$\PP(X^t \in S_t \mid \mathcal H_{t-1}) = D(S_t) = 0$ almost surely.
Thus $R(\sigma^t\mid D)=0$ in every round. Together with \eqref{eq:weak-opt-constant}, this gives
\[
R(T)=\sum_{t=1}^T\bigl(\OPT(D)-R(\sigma^t\mid D)\bigr)=T
\qquad\text{almost surely}.
\]
Consequently, $\EE[R(T)]=T$. The conditional probability calculation also shows that the realized cumulative reward is zero almost surely.
\end{proof}

\begin{remark}
The same example gives $\alpha$-regret $\alpha T$ almost surely for every fixed $\alpha\in(0,1]$, since an exact oracle is also an $\alpha$-approximation oracle. The failure depends on the oracle selection: the constant oracle $\mathbf P\mapsto\sigma^\circ$ witnesses strong monotonicity and achieves zero ordinary regret. Thus the construction shows a limitation of \Cref{alg:strong-up} with arbitrary oracle selections, rather than an impossibility for all learning algorithms.
\end{remark}

\subsection{Regret under Uniform Separate Bounded Variation}
\label{subsec:weak-bv-exact-exponent}

The construction in \Cref{thm:weak-monotonicity-linear-regret} uses policies that interpolate the finite support of the optimistic distribution. The reward $\one\{x\in S\}$ has pointwise total variation of order $|S|$, so these policies have no uniform variation bound. We now impose such a bound on the policies returned by the oracle.

For $g:[0,1]\to\mathbb R$, its \emph{pointwise total variation} is
\[
\operatorname{TV}_{[0,1]}(g):=\sup_{0=u_0<\cdots<u_m=1}\sum_{j=1}^m|g(u_j)-g(u_{j-1})|,
\]
where the supremum ranges over all finite partitions of $[0,1]$. 
For a policy $\sigma$, write $r_\sigma(x):=r(\sigma,x)$. We define the \emph{separate total variation} of $r_\sigma$ as
\begin{equation}\label{eq:weak-bv-definition}
\operatorname{TV}_{\rm sep}(r_\sigma):=\max_{i\in[n]}\sup_{x_{-i}\in[0,1]^{n-1}}\operatorname{TV}_{[0,1]}\bigl(u\mapsto r_\sigma(u,x_{-i})\bigr),
\end{equation}
where $r_\sigma(u,x_{-i})$ denotes the reward with coordinate $i$ equal to $u$ and all remaining coordinates fixed at $x_{-i}$.

A family $\mathcal N\subseteq\cC$ has \emph{uniform separate variation} at most $V_0$ if
\begin{equation}\label{eq:weak-bv-assumption}
\sup_{\sigma\in\mathcal N}\operatorname{TV}_{\rm sep}(r_\sigma)\le V_0.
\end{equation}
We require this condition only for a family containing the oracle's output policies. The optimal value and approximation guarantee remain defined relative to the full feasible class $\cC$.

The main result of this section is as follows.

\begin{theorem} \label{thm:weak-bv-exact-exponent}
Consider an up- or down-monotone maximization or minimization problem with $n$ items and realized objectives in $[0,1]$. Let $\mathcal O$ be a possibly randomized $\alpha$-approximation oracle, where $\alpha\in(0,1]$ for maximization and $\alpha\ge1$ for minimization, taking values in a family $\mathcal N\subseteq\cC$ satisfying \eqref{eq:weak-bv-assumption} for some finite $V_0\ge0$. For every $T\ge\max\{n,2\}$, \Cref{alg:strong-up}, using $\mathcal O$, $\delta=1/(T+1)$, and reverse shading in the down-monotone case, satisfies
\begin{equation}\label{eq:weak-bv-upper}
\EE[\alpha\text{-}R(T)]\le C(1+V_0)n^{3/4}\sqrt T\,\log^2(2nT)+1,
\end{equation}
where $C$ is universal. For randomized oracles, the approximation guarantee holds in expectation over the oracle's internal randomness, as in \Cref{sec:model}.

Furthermore, for every fixed $V_0>0$, there exists $c_{V_0}>0$ such that, for all sufficiently large $n$, there is an up-monotone maximization problem with rewards in $[0,1]$, a distribution-independent feasible class $\cC_n$ satisfying
\[
\sup_{\sigma\in\cC_n}\operatorname{TV}_{\rm sep}(r_\sigma)\le V_0,
\]
and an exact oracle $\mathcal O_n$ defined on all product distributions on $[0,1]^n$, for which the following holds. 
Under $\mathbf D = \prod_{i=1}^n\operatorname{Unif}[0,1]$, at a horizon $T_n=\Theta(n^{3/2})$, \Cref{alg:strong-up}, using $\mathcal O_n$ and $\delta=1/(T_n+1)$, satisfies
\begin{equation}\label{eq:weak-bv-lower}
\EE[R(T_n)]\ge c_{V_0}n^{3/2}.
\end{equation}
Consequently, for fixed $V_0>0$, no ordinary-regret bound of the form $\widetilde O(n^\beta\sqrt T)$ with $\beta<3/4$ holds uniformly for this algorithm over admissible problems and exact oracle selections.
\end{theorem}

The upper bound is proved in Appendix~\ref{sec:app-wbv-upper-bound}. The lower bound follows from \Cref{thm:app-wbv-lower-bound}; its horizon $T_n=\Theta(n^{3/2})$ gives the claimed restriction on the dimension exponent.

\paragraph{Proof sketch of the upper bound.}
To bound the regret, we need to compare the reward of the selected policy under the optimistic and true distributions. We show that the bounded-variation assumption allows us to control this difference in terms of the marginal estimation errors and the probabilities with which the policy probes the items.

We first describe the concentration bound used in the analysis. Let $\ell_T:=\log(2nT^2(T+1))$, $\rho_0:=1$, and $\rho_m:=\min\{1,C_0\ell_T/m\}$ for $m\ge1$, where $C_0$ is a sufficiently large absolute constant. Recall that $N_i^t$ is the number of observations of item $i$ before round $t$, and $\widetilde{\mathbf E}^{\,t}=\prod_i\widetilde E_i^t$ is the optimistic distribution used in that round. In \Cref{lem:app-wbv-shifted-interval}, we show that, with probability at least $1-1/T$, simultaneously for all rounds and items, $D_i\sd\widetilde E_i^t$ and
\begin{equation}\label{eq:weak-bv-relative-interval}
|\widetilde E_i^t(I)-D_i(I)|\le C\left(\sqrt{\rho_{N_i^t}D_i(I)}+\rho_{N_i^t}\right)
\end{equation}
for every interval $I\subseteq[0,1]$, including singletons. Thus, the error in estimating the probability of an interval depends on its probability under the true distribution.

The key step is to translate this estimate into a bound on the reward of a policy. Fix $\sigma\in\mathcal N$ and a product distribution $\mathbf E=\prod_iE_i$, and suppose that each pair $(D_i,E_i)$ satisfies the analogue of \eqref{eq:weak-bv-relative-interval} with parameter $\rho_i$. Recall that $\pi_i^{\mathbf D}(\sigma)$ is the probability that $\sigma$ probes item $i$ under $\mathbf D$. Define $A:=\sum_i\pi_i^{\mathbf D}(\sigma)\rho_i$ and $U:=\sum_i\pi_i^{\mathbf D}(\sigma)\rho_i^{2/3}$. We prove in \Cref{prop:app-wbv-one-round} that
\[
|R(\sigma\mid\mathbf E)-R(\sigma\mid\mathbf D)|\le C(1+V_0)\left(U^{3/4}+U+A\right).
\]
The dependence on the probing probabilities is crucial. An item may have a large estimation error because it has rarely been observed, but its contribution to the bound is weighted by the probability that the policy probes it.

To prove this stability bound, we first establish a one-dimensional inequality. Suppose distributions $P,Q$ satisfy the relative interval bound with parameter $\rho$, and $g:[0,1]\to[0,1]$ has pointwise total variation at most $V$. Then \Cref{lem:app-wbv-bv-fourth-root} gives
\begin{equation}\label{eq:weak-bv-fourth-root}
\left|\int g\,d(Q-P)\right|\le C\left(\sqrt{V\rho}\,\operatorname{Var}_P(g)^{1/4}+V\rho\right).
\end{equation}
We obtain this inequality by integrating the interval bounds over the level sets of the positive and negative parts of $g-\EE_P[g]$. The variation bound controls the number of interval components integrated over these levels.

We next apply \eqref{eq:weak-bv-fourth-root} to the continuation rewards of $\sigma$ under $\mathbf E$. After any prefix of the execution transcript, the expected reward from continuing the policy is a function of the next observed value. This function has variation at most $V_0$, since it is an average of coordinate sections of $r_\sigma$. We analyze these continuation rewards along an execution under $\mathbf D$. By telescoping their values and their squares, we relate the difference in expected rewards to the accumulated conditional variances. Applying H\"older's and Young's inequalities then gives the stability bound above.

We complete the proof using the same arguments for randomized oracles and adaptive counting as in \Cref{subsec:approximate-oracles}. Whenever the current marginals satisfy the concentration inequalities, we apply up-monotonicity, the oracle guarantee, and Jensen's inequality to bound the conditional $\alpha$-regret by $C(1+V_0)(U_t^{3/4}+U_t+A_t)$. Here $A_t,U_t$ are defined as above with $\rho_i=\rho_{N_i^t}$ and the conditional probing probabilities $\bar\pi_i^t$ in place of $\pi_i^{\mathbf D}(\sigma)$.

By the counting argument in \eqref{eq:adaptive-counting}, we have $\sum_t\EE[A_t]\le Cn\ell_T\log(eT)$ and $\sum_t\EE[U_t]\le Cn\ell_T^{2/3}T^{1/3}$. Applying Jensen's and H\"older's inequalities, we obtain
\[
\sum_{t=1}^T\EE[U_t^{3/4}]
\le T^{1/4}\left(\sum_{t=1}^T\EE[U_t]\right)^{3/4}
\le Cn^{3/4}\ell_T^{1/2}\sqrt T.
\]
This gives the $n^{3/4}$ dependence. Accounting for the concentration failure and using the trivial bound $\EE[\alpha\text{-}R(T)]\le T$ gives \eqref{eq:weak-bv-upper}. We provide the complete calculation in Appendix~\ref{sec:app-wbv-upper-bound}.

\paragraph{Proof sketch of the lower bound.}
We construct an instance in which the oracle combines empirical discrepancies across items into a constant loss under the true distribution. Let $\mathbf D=\prod_{i=1}^n\operatorname{Unif}[0,1]$, and partition $[0,1]$ into $k=\Theta(\sqrt n)$ equal intervals, with $k$ even. Write $b(x)$ for the interval containing $x$. Every policy probes all $n$ items, and its reward contains the term $\varepsilon\cos(\theta+a\sum_i s_i(b(x_i)))$, where $a=1/k$ and each $s_i:[k]\to\{-1,1\}$ is balanced, meaning $\sum_r s_i(r)=0$.

We also allow nondecreasing downward maps of the interval indices, which make the problem up-monotone. An additional increasing term makes identity maps optimal when the sum of the mean interval indices is sufficiently close to its value under $\mathbf D$. Choosing $\varepsilon>0$ depending only on $V_0$ ensures that all rewards lie in $[0,1]$ and every policy has separate total variation at most $V_0$. The construction admits an exact oracle on all product distributions.

Under the true distributions, every balanced sign has mean zero. In contrast, let $q_{i,m}$ be the largest balanced signed mean under item $i$'s shaded empirical distribution after $m$ observations. For $m\ge k$ and $k\ge C\ell_T$, we prove that $q_{i,m}=\Theta(\sqrt{k/m})$ with probability at least $1-Ce^{-ck}$; see \Cref{lem:app-wbv-shift-tv,lem:app-wbv-balanced-empirical}. For inputs satisfying the centering condition above, we choose the exact oracle to return identity maps and signs attaining these means. Its optimizing phase is
\[
\theta=-\Theta_m\pmod{2\pi},\qquad \Theta_m:=\sum_{i=1}^n\arctan(q_{i,m}\tan a)=\Theta\left(\frac{n}{\sqrt{km}}\right).
\]

Set $m_\star=\lfloor C_\star n^2/k\rfloor$ for a sufficiently large fixed constant $C_\star$, and take $T_n=2m_\star+1$. We show that the centering condition and discrepancy bounds hold simultaneously for all items and all $m\in[m_\star,2m_\star]$ with probability $1-o(1)$. Throughout this block, $\Theta_m$ is bounded away from zero and is at most $1$. Changing only the selected policy's phase to zero leaves its increasing term unchanged and improves its expected reward under $\mathbf D$ by
\[
\varepsilon(\cos a)^n(1-\cos\Theta_m)=\Omega_{V_0}(1).
\]
Here $(\cos a)^n$ is bounded below by a positive constant because $k=\Theta(\sqrt n)$.

Since every policy probes every item, $N_i^t=t-1$. The block therefore contains $\Theta(n^2/k)=\Theta(n^{3/2})$ rounds, each with constant ordinary pseudo-regret on the simultaneous event. Taking expectations gives \eqref{eq:weak-bv-lower}. The complete proof appears in \Cref{thm:app-wbv-lower-bound}. This establishes the dimension exponent $3/4$ for ordinary-regret guarantees that hold uniformly over admissible instances and exact oracle selections of \Cref{alg:strong-up}.

\paragraph{Comparison with prior work.}
For general distributions, \citet{agarwal2026semibandit} assume that the oracle returns $k$-threshold policies whose leaf reward functions are coordinate-wise continuous and have uniformly bounded coordinate variation. These assumptions imply our separate-variation condition. Indeed, for rewards in $[0,1]$ and leaf variation bounded by $V_{\rm leaf}$, fixing all but one coordinate gives at most $k$ pieces, each with variation at most $V_{\rm leaf}$, and at most $k-1$ jumps of size at most $1$. Thus $\operatorname{TV}_{\rm sep}(r_\sigma)\le kV_{\rm leaf}+(k-1)$. Our regularity condition is strictly more general: it also permits bounded-variation rewards with infinitely many jumps and requires no finite threshold representation.

Their Kolmogorov--Smirnov stability analysis also extends directly to policies with separate variation at most $V_0$. Under the assumptions of \Cref{thm:weak-bv-exact-exponent}, this gives expected $\alpha$-regret $O(V_0n\sqrt{T\log(2nT)}+1)$ for the same algorithm, using reverse shading for down-monotone problems; see \Cref{cor:agarwal-bv-regret}. This expresses the earlier analysis in terms of the same regularity parameter used in our upper bound.

For fixed $V_0>0$, comparison with \eqref{eq:weak-bv-upper} therefore shows that our analysis improves the polynomial dependence on $n$ from $n$ to $n^{3/4}$, up to logarithmic factors. We prove the implication between the regularity assumptions, the extended stability bound, and the resulting regret comparison in Appendix~\ref{sec:app-agarwal-comparison}.

%% file: apps.tex
\section{Applications}\label{sec:apps}

We now apply our results to several fundamental stochastic optimization problems. For most applications, we verify strong monotonicity for the chosen offline algorithm. For stochastic matching and the adaptive \((1/2-\varepsilon)\)-approximation for stochastic knapsack, we use our bounded-variation result under monotonicity, without requiring a transfer property. The improvement for this knapsack oracle holds for fixed support size. Throughout, the input distributions are unknown, all deterministic parameters are known, and probing an item reveals its full realization. \Cref{tab:application-summary} summarizes our guarantees and compares them with prior work; the corresponding subsections give the details and exact bounds.

\begin{table}[h!]
\centering
\small
\setlength{\tabcolsep}{3pt}
\renewcommand{\arraystretch}{1.5}
\begin{tabularx}{\linewidth}{|>{\centering\arraybackslash}X|c|c|c|c|}
\hline
\multirow{2}{*}{Application} & Approx. & Our \(\alpha\)-Regret & Prior Bound & Our Distributional \\
& Factor (\(\alpha\)) & \(\widetilde O(\cdot)\) & \(\widetilde O(\cdot)\) & Assumptions \\
\hline
Series Testing & \(1\) & \(C\sqrt{nT}\) & \(Cn\sqrt T\)\textsuperscript{G} & Bernoulli \\
\hline
Best-of-\(K\) Selection & \(1-\varepsilon\) & \(\sqrt{nT}\) & \(\sqrt{nKT}\)\textsuperscript{C} & Arbitrary on \([0,1]\) \\
\hline
Single-Choice Prophet Inequality & \(1\) & \(U\sqrt{nT}\) & \(U\sqrt{nT}\)\textsuperscript{L} & Arbitrary bounded \\
Multiple-Choice Prophet Inequality & \(1\) & \(kU\sqrt{nT}\) & \(knU\sqrt T\)\textsuperscript{G} & Arbitrary bounded \\
\hline
Pandora's Box & \(1\) & \((C+U)\sqrt{nT}\) & \(U\sqrt{nT}\)\textsuperscript{L} & Arbitrary bounded \\
\hline
\multirow{2}{*}{Stochastic Knapsack} & \(1/4\) & \(R\sqrt{nT}\) & \(nR\sqrt{kT}\)\textsuperscript{G} & Arbitrary bounded \\
& \(1/2-\varepsilon\) & \(kR n^{3/4}\sqrt T\) & \(nR\sqrt{kT}\)\textsuperscript{G} & Discrete \\
\hline
Stochastic Orienteering & \(\Omega(1/\max\{1,\log\log B\})\) & \(R\sqrt{nT}\) & \(nR\sqrt {kT}\) \textsuperscript{G} & Arbitrary bounded \\
\hline
Stochastic Submodular Maximization & \(1-1/e-\varepsilon\) & \(G\sqrt{nT}\) & \(nG\sqrt T\)\textsuperscript{G} & Bernoulli \\
\hline
Unweighted Stochastic Matching & \(1/2\) & \multirow{3}{*}{\(Wn^{3/4}\sqrt T\)} & \multirow{3}{*}{\(Wn\sqrt T\)\textsuperscript{G}} & \multirow{3}{*}{Bernoulli} \\
Weighted Stochastic Matching (\(b_v\ge 2\) for every \(v\)) & \(0.382\) & & & \\
\hline
\end{tabularx}
\caption{Summary of expected regret bounds. The Best-of-$K$ comparison concerns general marginals; Bernoulli inputs already admit $\widetilde O(\sqrt{nT})$ regret \citep{pmlr-v99-merlis19a}. Prior knapsack and orienteering bounds assume discrete distributions with at most \(k\) support values per item. The prior knapsack bounds use an adaptive \((1/2-\varepsilon)\)-approximation oracle, whereas our \(1/4\)-regret guarantee uses a nonadaptive oracle. The prior Pandora bound assumes \(c_i\le U\). For stochastic matching, \(n=|E|\) and \(W=\sum_{e\in E}w_e\).}
\label{tab:application-summary}
\par\smallskip
{\footnotesize\raggedright
Sources: G = \citet{agarwal2026semibandit};
C = \citet{chen2016generalreward}; L = \citet{liu2025improved}.
\par}
\end{table}

\subsection{Series Testing}\label{subsec:app-series}
We start with a simple problem. 
There are \(n\) components, where component \(i\) works independently with probability \(p_i\), and testing it costs \(c_i\ge0\). The system is functional if and only if every component works. The goal is to test components sequentially and determine whether the system is functional at minimum expected cost. 
Testing stops when the first failed component is found, so only the tested components are observed and the feedback is semi-bandit. 
Let \(Y_i\) be the indicator that component \(i\) fails and \(q_i:=\Pr(Y_i=1)=1-p_i\). Every deterministic policy is an ordering \(\pi\), whose expected cost under a product distribution \(\mathbf P\) is
\[
J(\pi\mid\mathbf P)=\sum_{k=1}^n c_{\pi(k)}\prod_{\ell<k}(1-q_{\pi(\ell)}).
\]
If \(\mathbf P\sd\mathbf Q\) with respect to the failure indicators, then \(q_i^{\mathbf P}\le q_i^{\mathbf Q}\) for every \(i\), and hence \(J(\pi\mid\mathbf Q)\le J(\pi\mid\mathbf P)\) for every ordering \(\pi\). Thus series-system testing is strongly up-monotone as a minimization problem. The optimal ordering sorts components by increasing \(c_i/q_i\), placing components with \(q_i=0\) last \citep{butterworth1972fault}. The resulting exact oracle runs in polynomial time. Since the realized testing cost lies in an interval of length at most \(C:=\sum_{i=1}^n c_i\), \Cref{thm:strong-min} gives a polynomial-time learning algorithm with
\( \EE[R(T)]=O\!\left(C\sqrt{nT}\,\log^{3/2}(2nT)\right)\), improving over the $O(n C\sqrt{T \log T})$ result of \citet{agarwal2026semibandit}.

\subsection{Best-of-$K$ Selection}\label{subsec:app-best-of-k}

The Best-of-$K$ selection problem asks for a set of \(K\) random variables
whose maximum has the largest expected value; see
\citet{pmlr-v49-simchowitz16}.  The same max-reward objective is called
\(K\)-MAX by \citet{chen2016generalreward}.  We use the former terminology
throughout.

There are \(n\) independent random variables \(X_i\sim D_i\), each supported
on \([0,1]\), and a fixed \(1\le K\le n\).  A policy chooses a set
\(S\subseteq[n]\) with \(|S|=K\) before observing the current outcomes,
receives reward
\(
    r(S,\xvec)=\max_{i\in S}x_i,
\)
and observes \(X_i\) for every \(i\in S\).  Thus the feedback is semi-bandit.
For a product distribution
\(\mathbf P=\prod_{i=1}^n P_i\), with marginal CDFs
\(F_i^{\mathbf P}\), the expected reward is
\[
    R(S\mid\mathbf P)
    =
    \EE_{\xvec\sim\mathbf P}
    \left[\max_{i\in S}X_i\right] =
    \int_0^1
    \left(
        1-\prod_{i\in S}F_i^{\mathbf P}(z)
    \right)\,dz.
\]

Best-of-\(K\) selection is strongly up-monotone.  Indeed, if
\(\mathbf P\sd\mathbf Q\), then for every fixed feasible set \(S\),
\(
    R(S\mid\mathbf Q)\ge R(S\mid\mathbf P),
\)
because \(r(S,\xvec)\) is coordinatewise nondecreasing.  Consequently,
the optimal value is up-monotone, and every exact optimization oracle
witnesses strong up-monotonicity.  Since the realized reward lies in
\([0,1]\), \Cref{thm:main} therefore gives an exact-oracle semi-bandit
algorithm satisfying
\(
    \EE[R(T)]
    =
    O\!\left(
        \sqrt{nT}\,\log^{3/2}(2nT)
    \right).
\)

For general marginal distributions, exact offline Best-of-\(K\) optimization
is NP-hard.  However, \citet{chen2016generalreward} give a
polynomial-time approximation scheme for the offline problem: for every
fixed \(\varepsilon>0\), it returns a
\((1-\varepsilon)\)-approximate solution in polynomial time.  Since every
fixed set has a coordinatewise nondecreasing realized reward, this oracle
also satisfies the transfer condition required by
\Cref{thm:approximate-oracle}.  Combining their offline approximation scheme
with our online algorithm therefore yields, for every fixed
\(\varepsilon>0\), a polynomial-time semi-bandit algorithm with
\(
    \EE[(1-\varepsilon)\text{-}R(T)]
    =
    O\!\left(
        \sqrt{nT}\,\log^{3/2}(2nT)
    \right).
\)
For independent Bernoulli rewards, \citet{pmlr-v49-simchowitz16} study fixed-confidence identification of the optimal $K$-set under semi-bandit feedback, while the results of \citet{pmlr-v99-merlis19a} give $\widetilde O(\sqrt{nT})$ cumulative regret. For arbitrary independent marginals on $[0,1]$, \citet{chen2016generalreward} obtain expected $(1-\varepsilon)$-regret $\widetilde O(\sqrt{nKT})$. Our bound removes the polynomial dependence on $K$ in this guarantee while preserving their approximation factor.

\subsection{Prophet Inequality}\label{subsec:app-prophet}
The prophet inequality is a fundamental problem in optimal stopping with applications in algorithmic game theory \citep{krengel1977semiamarts, samuelcahn1984comparison}. 
The input consists of independent rewards \(X_i\sim D_i\), supported on \([0,U]\), which arrive in a fixed order. After observing each reward, a policy must accept it and stop or reject it irrevocably; it may accept at most one reward. 
Thus only a prefix of the rewards is observed, giving semi-bandit feedback. Classical prophet inequalities compare a stopping policy with the clairvoyant benchmark \(\EE[\max_iX_i]\) and obtain a tight \(1/2\)-approximation. Here we instead compare with the optimal nonclairvoyant policy subject to the same arrival order. For known distributions, this policy is computed exactly by backward induction: setting \(V_{n+1}=0\) and \(V_i=\EE[\max\{X_i,V_{i+1}\}]\), it accepts \(X_i\) if and only if \(X_i\ge V_{i+1}\). Under our convention that \(\mathbf P\sd\mathbf Q\) means that \(\mathbf Q\) stochastically dominates \(\mathbf P\), \citet[Lemma~28]{guo2021generalizing} show that this exact oracle witnesses strong up-monotonicity. Since the selected reward lies in \([0,U]\), \Cref{thm:main} gives a polynomial-time learning algorithm with
\(\EE[R(T)]=O\!\left(U\sqrt{nT}\,\log^{3/2}(2nT)\right).\)
This improves the dimension dependence of the \(O(nU\sqrt{T\log T})\) guarantee of \citet{agarwal2026semibandit} from \(n\) to \(\sqrt n\), and matches up to logarithmic factors the problem-specific \(\widetilde O(U\sqrt{nT})\) guarantee of \citet{liu2025improved}.

\paragraph{Multiple-Choice Prophet Inequality.}
The multiple-choice prophet inequality, also commonly called the
$k$-unit prophet inequality, extends the classical prophet inequality by
allowing up to $k$ selections
\citep{hajiaghayi2007automated,alaei2014bayesian,kleinberg2019matroid}.
There are $n$ independent rewards $X_i\sim D_i$, each supported on
$[0,U]$, which arrive in a fixed order.  After observing $X_i$, a policy
must irrevocably accept or reject it, and may accept at most $k$ rewards
in total.  Its realized reward is the sum of the accepted rewards.  Once
the $k$th reward is accepted, the policy may stop, so only a prefix of
the rewards is observed; hence the feedback is semi-bandit.  The case
$k=1$ reduces to the prophet-inequality problem considered above.

As in the preceding subsection, our benchmark is the optimal
nonclairvoyant policy subject to the same arrival order.  For a product distribution
$\mathbf P=\prod_{i=1}^n P_i$, let $V_{i,j}^{\mathbf P}$ denote the
optimal expected reward from items $i,i+1,\ldots,n$ when $j$ selections
remain.  The boundary conditions are
\(    V_{n+1,j}^{\mathbf P}=0 \)
and \(V_{i,0}^{\mathbf P}=0,\)
and backward induction gives, for $j\ge 1$,
\(
    V_{i,j}^{\mathbf P}
    =
    \EE_{X_i\sim P_i}
    \left[
        \max\left\{
            V_{i+1,j}^{\mathbf P},
            X_i+V_{i+1,j-1}^{\mathbf P}
        \right\}
    \right].
\)
Thus an optimal policy accepts the realization $X_i=x$ in state $(i,j)$
if and only if
\(
    x\ge
    \theta_{i,j}^{\mathbf P}
    :=
    V_{i+1,j}^{\mathbf P}
    -
    V_{i+1,j-1}^{\mathbf P}.
\)
For finitely supported marginals, all values
$V_{i,j}^{\mathbf P}$ and thresholds
$\theta_{i,j}^{\mathbf P}$ can be computed by a polynomial-time dynamic
program, so this gives an exact oracle of the type required by our
online algorithm.

We next verify strong up-monotonicity.  Suppose that
$\mathbf P\sd\mathbf Q$, so every marginal under $\mathbf Q$
stochastically dominates the corresponding marginal under $\mathbf P$.
Fix the optimal dynamic-programming policy
$\mathcal O_k(\mathbf P)$ defined above, and let
$W_{i,j}^{\mathbf Q}$ be its expected reward from state $(i,j)$ when the
realizations are instead drawn from $\mathbf Q$.  We claim by backward
induction that
\(
    W_{i,j}^{\mathbf Q}
    \ge
    V_{i,j}^{\mathbf P}
    \,,
    \text{for every }i,j.
\)
The claim is immediate at the boundary states.  For the induction step,
write
\(
    \theta
    =
    V_{i+1,j}^{\mathbf P}
    -
    V_{i+1,j-1}^{\mathbf P}.
\)
Using the induction hypothesis after either rejecting or accepting item
$i$ gives
\[
    W_{i,j}^{\mathbf Q}
    \ge
    \EE_{Y_i\sim Q_i}
    \left[
        g_{i,j}^{\mathbf P}(Y_i)
    \right],
\text{where }
    g_{i,j}^{\mathbf P}(y)
    :=
    \begin{cases}
        V_{i+1,j}^{\mathbf P}, & y<\theta,\\[2mm]
        y+V_{i+1,j-1}^{\mathbf P}, & y\ge\theta.
    \end{cases}
\]
By the definition of $\theta$, the two pieces agree at the threshold, so
$g_{i,j}^{\mathbf P}$ is nondecreasing.  Stochastic dominance therefore
implies
\(
    \EE_{Y_i\sim Q_i}
    \left[
        g_{i,j}^{\mathbf P}(Y_i)
    \right]
    \ge
    \EE_{X_i\sim P_i}
    \left[
        g_{i,j}^{\mathbf P}(X_i)
    \right]
    =
    V_{i,j}^{\mathbf P}.
\)
Consequently,
\(
    R(\mathcal O_k(\mathbf P)\mid\mathbf Q)
    \ge
    R(\mathcal O_k(\mathbf P)\mid\mathbf P)
    =
    \OPT(\mathbf P),
\)
so the dynamic-programming oracle witnesses strong up-monotonicity.
Since the realized reward lies in $[0,kU]$, \Cref{thm:main} yields a
polynomial-time semi-bandit learning algorithm satisfying
\[
    \EE[R(T)]
    =
    O\!\left(
        kU\sqrt{nT}\,\log^{3/2}(2nT)
    \right).
\]

Applying the framework of \citet{agarwal2026semibandit} to the same dynamic-programming oracle gives expected regret \(O(knU\sqrt{T\log(2nT)})\) against the same nonclairvoyant benchmark, so our bound improves the dependence on \(n\) from \(n\) to \(\sqrt n\), up to logarithmic factors.
Classical multiple-choice prophet inequalities instead compare an online policy with a clairvoyant prophet that selects the $k$ largest realized rewards.  The equal-cardinality problem was studied by \citet{hajiaghayi2007automated}, and \citet{alaei2014bayesian} obtained the asymptotically optimal-order guarantee $1-1/\sqrt{k+3}$. 

\subsection{Pandora's Box}\label{subsec:app-pandora}
In Pandora's box, there are \(n\) boxes with independent rewards \(X_i\sim D_i\), supported on \([0,U]\), and known inspection costs \(c_i>0\), and outside option zero \citep{weitzman1979optimal}. 
A policy adaptively inspects subset \(S\subseteq[n]\), observing \(X_i\) only after paying \(c_i\), and seeks to maximize \(\EE[\max_{i\in S}X_i-\sum_{i\in S}c_i]\). 
Thus only inspected rewards are observed, giving semi-bandit feedback. For known distributions, Weitzman's optimal policy assigns each box a reservation value \(\rho_i\) satisfying \(\EE[(X_i-\rho_i)^+]=c_i\), inspects boxes in decreasing order of \(\rho_i\), and stops once the largest observed reward is at least every remaining reservation value. 
Under our convention, \citet[Lemma~31]{guo2021generalizing} show that if \(\mathbf P\sd\mathbf Q\), then this policy \(\mathcal O(\mathbf P)\) satisfies \(R(\mathcal O(\mathbf P)\mid\mathbf Q)\ge R(\mathcal O(\mathbf P)\mid\mathbf P)\); hence the Weitzman oracle witnesses strong up-monotonicity. 
Since the realized utility lies in \([-C,U]\), where \(C:=\sum_{i=1}^n c_i\), \Cref{thm:main} gives a polynomial-time learning algorithm with \(\EE[R(T)]=O((C+U)\sqrt{nT}\,\log^{3/2}(2nT))\). 
This improves the dimension dependence of the \(O(n(C+U)\sqrt{T\log T})\) guarantee of \citet{agarwal2026semibandit} from \(n\) to \(\sqrt n\). 
The problem-specific analysis of \citet{liu2025improved} is sharper: even when \(X_i, c_i\in[0,1]\), so that \(C\) may be as large as \(n\), it obtains \(\widetilde O(\sqrt{nT})\) regret without the \(C+1\) range factor retained by our generic theorem.

\subsection{Stochastic Knapsack}\label{subsec:app-knapsack}
There are \(n\) items with deterministic rewards \(r_i\ge 0\) and independent random costs \(C_i\sim D_i\). The realized cost \(C_i\) is observed only after item \(i\) is selected. Given a knapsack budget \(B>0\), a policy selects items sequentially until the cumulative cost exceeds \(B\), and maximizes the expected total reward of the items that fit in the knapsack. The item that causes an overflow earns no reward, while the rewards of previously completed items are retained. Since only the costs of selected items are observed, the feedback is semi-bandit. Costs exceeding \(B\) may be identified with a common overflow value and rescaled to fit our bounded-support setting. Write \(R:=\sum_{i=1}^n r_i\).

\paragraph{Exact regret.} Consider product distributions \(\mathbf P\sd\mathbf Q\), so item costs under \(\mathbf Q\) are stochastically larger. For a set \(S\) of untried items and remaining budget \(b\), let
\[
V_{\mathbf Q}(S,b)=\max\left\{0,\max_{i\in S} \EE_{Y_i\sim Q_i} \left[\one\{Y_i \le b\} \cdot \bigl(r_i+V_{\mathbf Q}(S\setminus\{i\},b-Y_i)\bigr)\right]\right\}.
\]
The function \(b\mapsto V_{\mathbf Q}(S,b)\) is non-decreasing. Let \(\mathcal O_{\mathrm{DP}}(\mathbf Q)\) choose a maximizing Bellman action at every state, and let \(W_{\mathbf P}(S,b)\) be the expected reward obtained by this policy under \(\mathbf P\). We claim that \(W_{\mathbf P}(S,b)\ge V_{\mathbf Q}(S,b)\) for every \(S,b\). The claim is immediate for \(S=\varnothing\), and the induction step is trivial if the policy stops. Otherwise, if it attempts item \(i\), the induction hypothesis gives
\[
W_{\mathbf P}(S,b)\ge\EE_{X_i\sim P_i}\!\left[\one\{X_i\le b\}\bigl(r_i+V_{\mathbf Q}(S\setminus\{i\},b-X_i)\bigr)\right].
\]
The integrand is non-increasing in the realized cost. Since \(P_i\sd Q_i\), its expectation under \(P_i\) is at least its expectation under \(Q_i\), which equals \(V_{\mathbf Q}(S,b)\) by the Bellman choice. Consequently,
\[
R(\mathcal O_{\mathrm{DP}}(\mathbf Q)\mid\mathbf P)\ge R(\mathcal O_{\mathrm{DP}}(\mathbf Q)\mid\mathbf Q)=\OPT(\mathbf Q).
\]
This also implies \(\OPT(\mathbf P)\ge\OPT(\mathbf Q)\), so stochastic knapsack is down-monotone and \((\cP,\mathcal O_{\mathrm{DP}})\) has oracle-compatible strong down-monotonicity. Therefore, \Cref{thm:strong-down-max} gives an oracle-efficient learning algorithm with
\(
\EE[R(T)]=O \left(R\sqrt{nT}\,\log^{3/2}(2nT)\right).
\)
The exact dynamic program is generally exponential; this is an oracle-efficient rather than necessarily polynomial-time guarantee.

\paragraph{Approximate regret.} The preceding argument already shows that stochastic knapsack is down-monotone for arbitrary, possibly continuous, cost distributions. It remains to verify the transfer property for an approximation oracle. \citet{dean2008stochastic} give a polynomial-time nonadaptive \(1/4\)-approximation: their oracle returns a fixed ordering \(\pi_{\mathbf Q}\) whose expected reward under \(\mathbf Q\) is at least \(\OPT(\mathbf Q)/4\). For \(\mathbf P\sd\mathbf Q\), couple the item costs \(X_i\sim P_i\) and \(Y_i\sim Q_i\) so that \(X_i\le Y_i\) almost surely for every \(i\). When the same ordering is executed on the coupled realizations, every item that fits under \(\mathbf Y\) also fits under \(\mathbf X\). Therefore,
\(
R(\pi_{\mathbf Q}\mid\mathbf P)\ge R(\pi_{\mathbf Q}\mid\mathbf Q).
\)
Thus the DGV oracle satisfies the required transfer property. The approximation extension following \Cref{thm:strong-down-max} gives a polynomial-time learning algorithm with
\(\EE[\tfrac14\text{-}R(T)]=O \left(R\sqrt{nT}\,\log^{3/2}(2nT)\right).\)

For discrete cost distributions with maximum support size \(k\), \citet{agarwal2026semibandit} combine their framework with the adaptive oracle of \citet{ma2018improvements} to obtain \((1/2-\varepsilon)\)-regret \(O(nR\sqrt{kT\log T})\). Our \(1/4\)-regret guarantee uses the nonadaptive oracle of \citet{dean2008stochastic}. It improves the dependence on \(n\) from \(n\) to \(\sqrt n\), up to logarithmic factors, removes the factor \(\sqrt k\), and applies to arbitrary, possibly continuous, cost distributions, at the cost of a weaker approximation factor.
However, our bounded-variation result also improves the guarantee for the same adaptive oracle. For the analysis, work with rescaled costs and let \(A_i:=\operatorname{supp}(D_i)\cup\{0\}\) and \(q_i(x):=\max\{a\in A_i:a\le x\}\). Define an auxiliary execution that runs each oracle policy on the rounded costs \(q_i(x_i)\), using these costs for both budget updates and reward calculation. Its normalized reward
\[
\bar r_\sigma(\mathbf x):=\frac{r_\sigma(q_1(x_1),\ldots,q_n(x_n))}{R}
\]
has at most \(k+1\) constant pieces in each coordinate and hence separate total variation at most \(k\). Since the true and optimistic marginals are supported on \(A_i\), the auxiliary and actual executions agree under both distributions, including their rewards and probing probabilities. The sets \(A_i\) are used only in the analysis and need not be known to the learner. Applying \Cref{prop:app-wbv-one-round} to the auxiliary execution and using the original oracle guarantee, down-monotonicity, and the counting argument in the proof of \Cref{thm:weak-bv-exact-exponent} gives expected \((1/2-\varepsilon)\)-regret \(\widetilde O(kRn^{3/4}\sqrt T)\), without requiring a transfer property. 
For fixed \(k\), this improves the dependence on \(n\) from \(n\) to \(n^{3/4}\) while preserving the approximation factor.

\paragraph{Stochastic Orienteering.} The argument also extends to stochastic orienteering. There are \(n\) jobs located at the vertices of a metric space with a designated root. Job \(i\) has deterministic reward \(r_i\ge0\) and an independent random processing cost \(C_i\). Starting from the root, a policy travels between vertices and processes jobs, subject to a budget \(B\) on the sum of the deterministic travel costs and realized processing costs; a job earns its reward only if it is completed within the budget. We assume that the full processing cost of every attempted job is observed, including the final job that exceeds the remaining budget, so the feedback is semi-bandit.

The Bellman argument above continues to apply after augmenting the state by the current vertex and subtracting the deterministic travel cost before processing the next job. Hence stochastic orienteering is down-monotone with respect to the processing costs. 
For integral travel costs, \citet{gupta2015orienteering} give a polynomial-time nonadaptive \(\alpha_B\)-approximation to the optimal adaptive policy, where
$\alpha_B = \Omega\!\left(\frac{1}{\max\{1,\log\log B\}}\right).$
Their result allows arbitrary processing-cost distributions. The oracle accesses these distributions through tail probabilities and truncated expectations, which are explicitly computable for the finitely supported distributions supplied by our learner.
Their oracle fixes a path before any processing costs are observed. For \(\mathbf P\sd\mathbf Q\), couple costs \(X_i\sim P_i\) and \(Y_i\sim Q_i\) so that \(X_i\le Y_i\) almost surely. Along any fixed path, the travel cost of each prefix is unchanged and its processing cost under \(\mathbf P\) is no larger; consequently, every job completed under \(\mathbf Q\) is also completed under \(\mathbf P\). Conditioning on any internal randomization used to select the path and then averaging gives
$R(\mathcal O_{\mathrm{SO}}(\mathbf Q)\mid\mathbf P)\ge R(\mathcal O_{\mathrm{SO}}(\mathbf Q)\mid\mathbf Q).$ 
Thus the oracle satisfies the required transfer property, and the approximation extension following \Cref{thm:strong-down-max} gives a polynomial-time learning algorithm with
$\EE[\alpha_B\text{-}R(T)]=O\left(R \sqrt{nT}\, \log^{3/2}(2nT)\right)$.
For discrete processing-cost distributions with maximum support size \(k\), \citet{agarwal2026semibandit} obtain the same approximation factor \(\alpha_B\) with additive regret \(O(nR\sqrt{kT\log T})\). Our bound improves the dependence on \(n\) from \(n\) to \(\sqrt n\), up to logarithmic factors, removes the factor \(\sqrt k\), and applies to arbitrary, possibly continuous, processing-cost distributions.

\subsection{Stochastic Submodular Maximization}\label{subsec:stochastic-submodular}
We follow the model in \citet{agarwal2026semibandit}. Let \(E=[n]\), let \(f:2^E\to\mathbb Z_{\ge0}\) be a known monotone submodular function, and let \(\mathcal I\subseteq2^E\) be the independent sets of a known matroid. Each item \(i\in E\) is active independently with unknown probability \(p_i\), and its activation status is revealed only after it is selected. Write \(X_i\sim\operatorname{Bernoulli}(p_i)\) and \(A:=\{i\in E:X_i=1\}\) for the random set of active items. A policy selects a feasible set \(S\in\mathcal I\) sequentially and adaptively, seeking to maximize \(\EE[f(S\cap A)]\). Only selected activation indicators are observed, giving semi-bandit feedback. A nonadaptive policy fixes \(S\) before observing any outcomes. We take \(G:=f(E)-f(\varnothing)\) as an upper bound on the objective range.

\paragraph{Monotonicity and transfer.}
Monotonicity of the optimal value follows from \citet{agarwal2026semibandit}. For every fixed feasible set \(S\), the realized reward \(f(S\cap A)\) is nondecreasing in each activation indicator. Thus \Cref{lem:nonadaptive-transfer} gives the required transfer property for any nonadaptive approximation oracle returning a feasible set.

\paragraph{Approximation regret.}
Fix \(0<\varepsilon<1-1/e\) and let \(\alpha_\varepsilon:=1-1/e-\varepsilon\). Assuming exact access to the expectations required by the offline algorithm, \citet{asadpour2016maximizing} provide a randomized nonadaptive \(\alpha_\varepsilon\)-approximation to the optimal adaptive policy, with oracle complexity polynomial in \(n\) and \(1/\varepsilon\). 
The expected guarantee follows by applying their algorithm to \(f-f(\varnothing)\), with approximation tolerance and failure probability at most \(\varepsilon/2\). Since shading preserves Bernoulli marginals, \Cref{lem:nonadaptive-transfer,thm:approximate-oracle} yield an oracle-efficient learning algorithm with
\(
\EE[\alpha_\varepsilon\text{-}R(T)] = O( G\sqrt{nT}\,\log^{3/2}(2nT)).
\)
Applying \citet{agarwal2026semibandit} to the shifted objective \(f-f(\varnothing)\) gives regret \(O(nG\sqrt{T\log T})\) for the same approximation factor and optimal adaptive benchmark. Our bound improves the dimension dependence from \(n\) to \(\sqrt n\), up to logarithmic factors.

\subsection{Stochastic Matching}\label{subsec:stochastic-matching}
Consider an undirected graph \((V,E)\) with known nonnegative edge weights \(w_e\) and vertex probing limits \(b_v\). Each edge \(e\) is active independently with unknown probability \(p_e\), revealed only when the edge is probed. A policy may probe an edge whose endpoints are unmatched, subject to probing at most \(b_v\) edges incident to each vertex \(v\). Every active probed edge must be added to the matching. The objective is to maximize the expected total weight of the resulting matching. Only probed activation indicators are observed, giving semi-bandit feedback. Write \(n:=|E|\) and \(W:=\sum_{e\in E}w_e\).

The offline algorithms of \citet{adamczyk2011greedy} and \citet{brubach2021offline} give approximation factors \(1/2\) for unweighted matching and \(0.382\) for weighted matching, respectively, where the latter assumes that each vertex has patience at least \(2\). \citet{agarwal2026semibandit} establish monotonicity and combine these algorithms with their framework to obtain expected \(\alpha\)-regret \(\widetilde O(Wn\sqrt T)\).

To apply our bounded-variation result, normalize rewards by \(W\) and extend each binary observation to \([0,1]\) by thresholding at \(1/2\). Every coordinate section of a fixed policy's normalized reward then has at most two constant pieces and total variation at most one. Since the true and optimistic marginals are Bernoulli, this extension leaves all relevant executions unchanged. Thus \Cref{thm:weak-bv-exact-exponent} gives expected \(\alpha\)-regret \(\widetilde O(Wn^{3/4}\sqrt T)\) for the same approximation factors, improving the dependence on \(n\) from \(n\) to \(n^{3/4}\).

%% file: computations-syn.tex
\section{Computational Experiments}
\label{sec:computations}

We evaluate our framework on Bernoulli Best-of-$K$ selection, prophet inequality, and Pandora's box, using both finite-support and continuous distributions. We averaged results over 50 independent paired replications for the experiments involving finite-support and heterogeneous-Uniform cases. We averaged results over 12 independent paired replications for the experiments involving Uniform/Beta and dimension cases. Across all problem instances, our algorithm computes shaded empirical distributions using Bernstein-type methods and uses a problem-specific offline oracle. We evaluated the performance of the framework by computing cumulative regret. The framework performed well on finite-support instances of all three applications, whereas the empirical plug-in outperformed, on average, on the continuous cases at the given horizons. Additional implementation details and plots are given in Appendix~\ref{app:computations}.

We compare Algorithm~\ref{alg:strong-up}, labeled \emph{Ours}, with the algorithm of \citet{agarwal2026semibandit}, labeled \emph{AGNZ}, using their original confidence settings and initialization rules without tuning. 

We also include an \emph{empirical plug-in} method, which applies the same offline oracle to the empirical product distribution without a confidence adjustment, and an application-specific \emph{Uniform} baseline. Before an item is first observed, empirical plug-in uses a point mass at $1$ for Bernoulli Best-of-$K$, equal mass on the four specified support points for discrete prophet and Pandora instances, and $\operatorname{Unif}[0,1]$ for continuous instances. It uses the empirical distribution thereafter and has no forced initialization phase. All initialization rounds are counted as online rounds, and their regret is included.

For a policy $\sigma^t$ executed in round $t$, we report cumulative pseudo-regret
\[\Reg(h):=\sum_{t=1}^{h}\left(\OPT(\mathbf D)-R(\sigma^t\mid\mathbf D)\right).\]
The benchmark is the optimal policy with knowledge of the distributions and the same within-round observation constraints. The true distributions are used to evaluate policies and the benchmark, and are not revealed to the learner. Within each replication, all methods share the same realization of the items, while each learner observes only the realizations revealed by its own policy. The curves also show mean pseudo-regret with pointwise $95\%$ Student-t intervals; the dimension plot uses percentile bootstrap intervals obtained by resampling whole replications. Some intervals are narrower than the plotted curves.



\subsection{Bernoulli Best-of-$K$ Selection}

\paragraph{Instances.} There are $n$ independent Bernoulli items, exactly $K$ of which are chosen uniformly at random to be good. A good item has success probability $1/K$, while every other item has success probability $1/(2K)$. In each round, the learner selects $K$ items before observing their outcomes, receives their maximum realization, and observes all $K$ selected outcomes. A set containing $g$ good items therefore has expected reward $1-(1-1/K)^g(1-1/(2K))^{K-g}$. The exact offline oracle selects the $K$ items with the largest estimated success probabilities. The Uniform baseline selects a uniformly random $K$-subset.

We consider $K=n/2$ and $K=5$, with $n\in\{50,100,500,1000\}$. These settings allow the number of observations per round to grow with the dimension or remain fixed. In a separate dimension experiment, we fix $K=n/2$ and $T=100{,}000$, and vary $n\in\{16,32,64,\ldots,16384\}$.

\begin{figure*}[t]
\centering
\begin{subfigure}[t]{.485\textwidth}
\centering
\includegraphics[width=\linewidth]{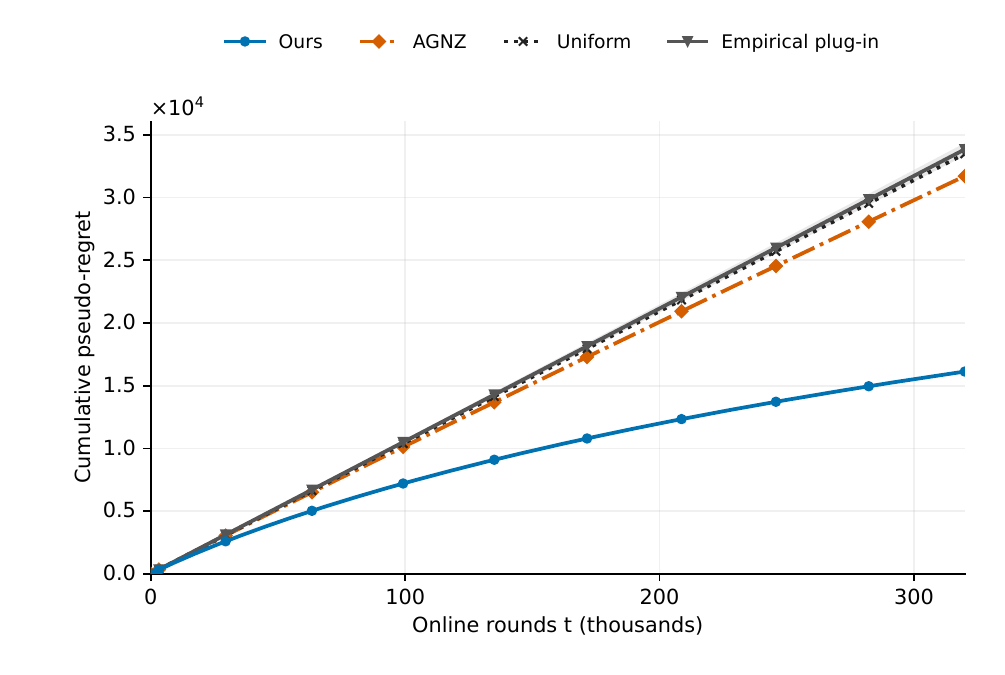}
\caption{$K=n/2$, $T=320{,}000$.}
\end{subfigure}\hfill
\begin{subfigure}[t]{.485\textwidth}
\centering
\includegraphics[width=\linewidth]{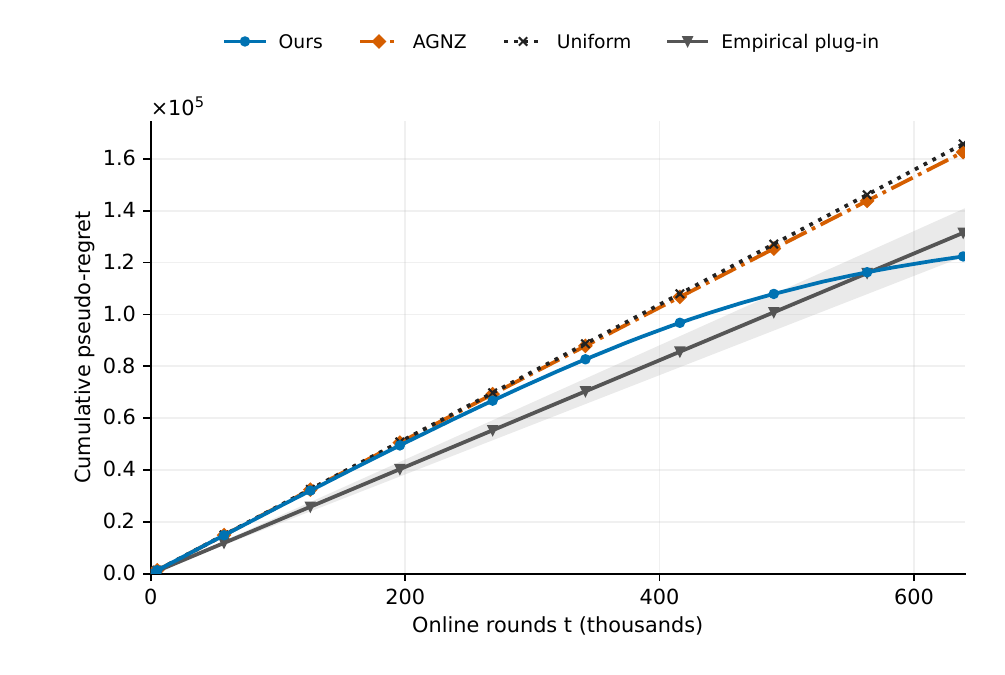}
\caption{$K=5$, $T=640{,}000$.}
\end{subfigure}
\caption{Computational results for Bernoulli Best-of-$K$ selection at $n=500$.}\label{fig:best-k-time}
\end{figure*}

\paragraph{Results.} Our algorithm has the lowest mean endpoint regret among the four methods in every tested configuration of both regimes. \Cref{fig:best-k-time} shows the regret curves at $n=500$: relative to AGNZ, our algorithm reduces mean endpoint regret by approximately $49\%$ for $K=n/2$ and $25\%$ for $K=5$, at the respective displayed horizons. For $K=5$ at $n=500$, the empirical plug-in method has lower mean regret during the early part of the horizon, but our algorithm ultimately achieves a lower mean regret by the end with a more stable behavior. 

\begin{figure*}[t]
\centering
\includegraphics[width=.78\textwidth]{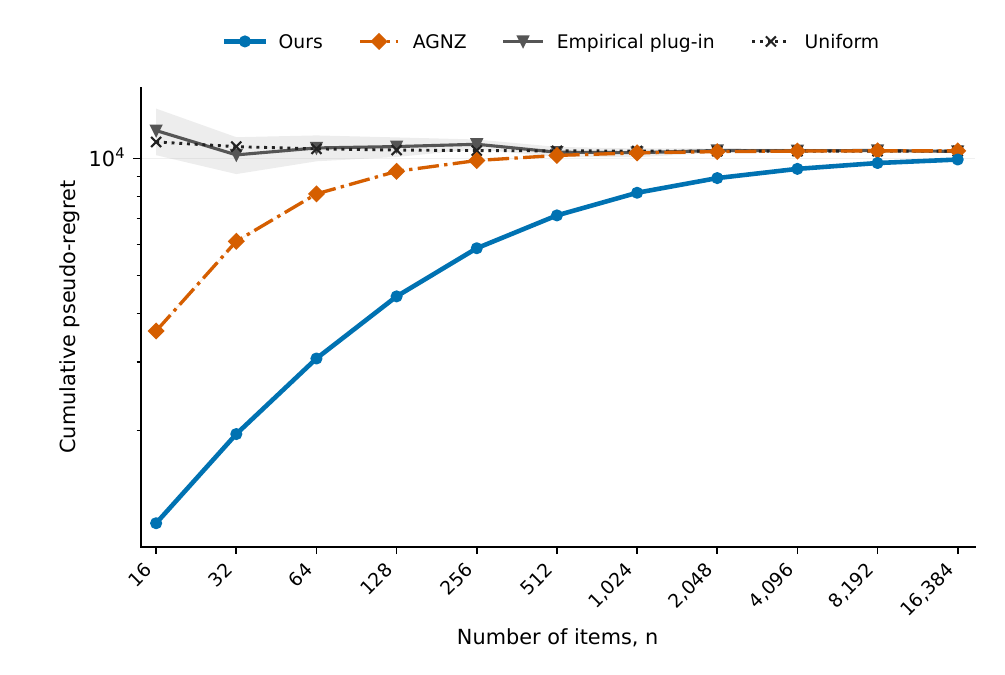}
\caption{Regret versus $n$ for Bernoulli Best-of-$K$ selection with $K=n/2$ and $T=100{,}000$.}\label{fig:best-k-dimension}
\end{figure*}

In the dimension experiment, which tests a broader range of $n$, our algorithm also has the lowest mean regret at every tested \(n\); see \Cref{fig:best-k-dimension}. The observation that the regret of \emph{Ours} approaches that of Uniform is consistent with our theory. As \(n\) grows to the order of \(T\), \(\sqrt{nT}\) grows to the trivial regret upper bound \(T\). The regret of \emph{AGNZ} approaches that of Uniform earlier when \(n \asymp \sqrt{T}\), which is consistent with its \(O(n\sqrt{T})\) regret bound. This experiment illustrates the dependence of the regret on \(n\). The remaining regret plots are reported in Appendix~\ref{app:best-k}.

\subsection{Multiple-Choice Prophet Inequality}

\paragraph{Instances.} In each round, $n$ independent rewards arrive in a fixed order, and the learner may accept at most $k$ of them and get the total reward of the $k$ items. The policy stops after its $k$th acceptance, and only items in the visited prefix are observed. We use the optimal known-distribution nonclairvoyant policy as our benchmark. The offline oracle computes optimal acceptance thresholds by dynamic programming. The Uniform baseline selects a uniformly random $k$-subset of arrival positions in advance and accepts those positions.

For the discrete instances, rewards have support $\{0,1/4,1/2,1\}$. There are four approximately equally represented item types, randomly permuted once per replication. Writing $q=1/(4n)$, their probabilities at the three positive support points are $q(16,0,0)$, $q(0,8,0)$, $q(0,0,4)$, and $q(4,2,2)$, respectively, with the remaining mass at zero. We use $k\in\{1,2,5\}$ and $n\in\{50,100,500,1000\}$.

For the continuous instances, $X_i\sim\operatorname{Unif}[0,U_i]$, where $U_i\in\{0.2,0.4,0.6,0.8,1\}$. The endpoints and arrival order are fixed across replications, and the endpoints are hidden from the learner. Both optimistic algorithms use raw empirical CDFs without discretization. We use $k\in\{1,2,5\}$ and $n\in\{50,100,500,1000\}$.
We also consider the Uniform/Beta family: the first $k$ rewards are $\operatorname{Unif}[0,1]$, and the remaining $n-k$ rewards are $\operatorname{Beta}(1,16n)$, independently, in this fixed order. For $k=1$, this is the distribution family in \Cref{thm:agn-continuous-lower-bound}.

\begin{figure*}[t]
\centering
\begin{subfigure}[t]{.485\textwidth}
\centering
\includegraphics[width=\linewidth]{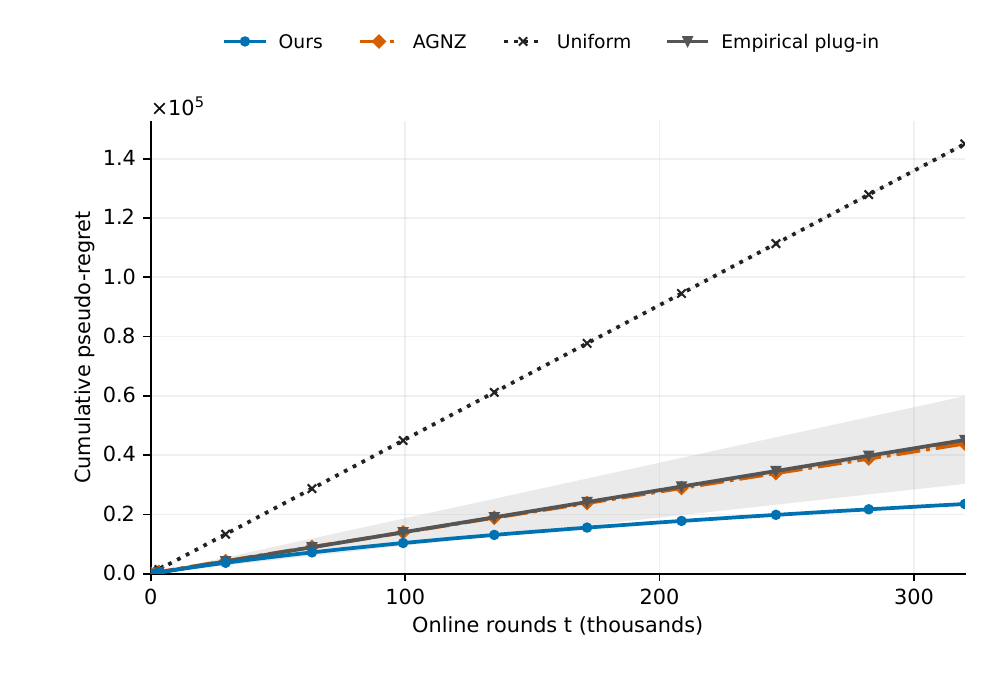}
\caption{Discrete, $n=500$, $k=1$.}\label{fig:prophet-discrete}
\end{subfigure}\hfill
\begin{subfigure}[t]{.485\textwidth}
\centering
\includegraphics[width=\linewidth]{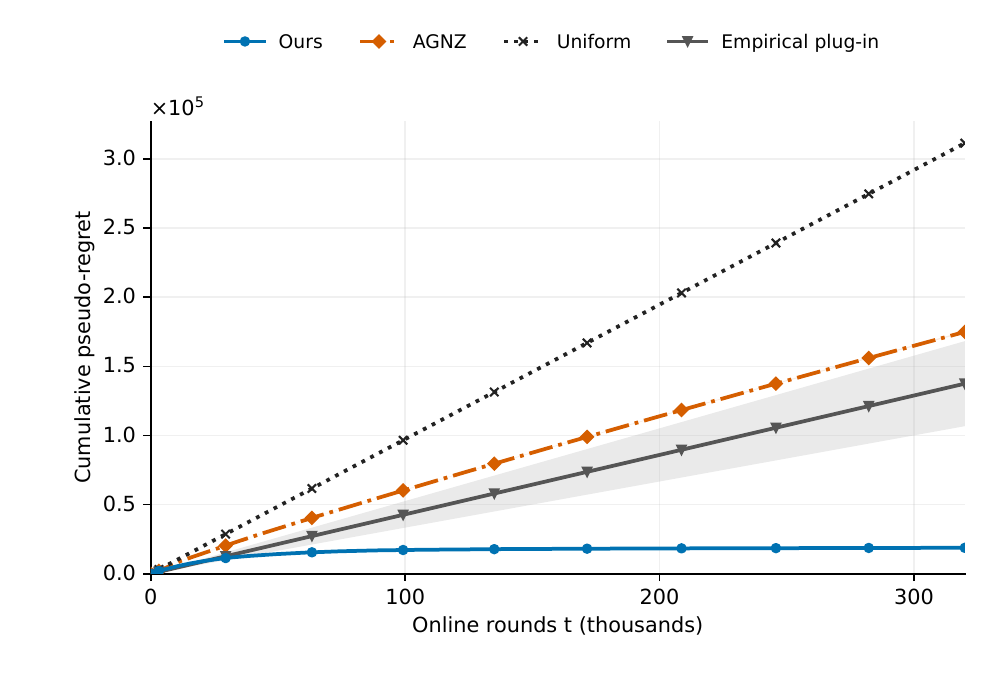}
\caption{Discrete, $n=500$, $k=5$.}\label{fig:prophet-multiple}
\end{subfigure}

\medskip
\begin{subfigure}[t]{.485\textwidth}
\centering
\includegraphics[width=\linewidth]{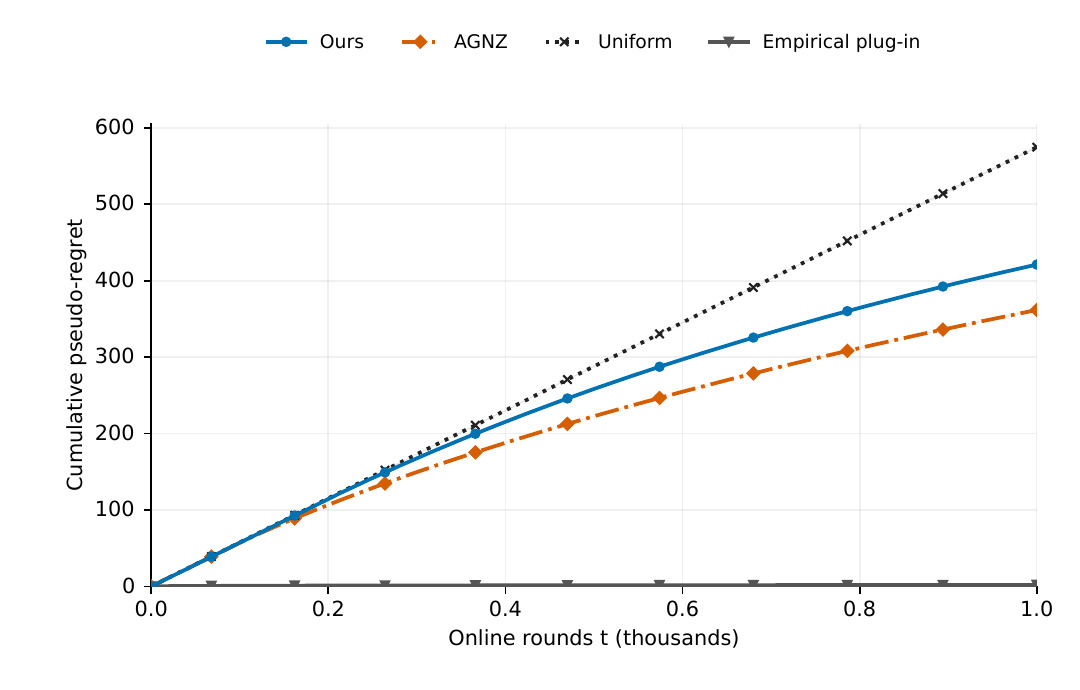}
\caption{heterogeneous-Uniform, $n=50$, $k=1$.}\label{fig:cont-prophet-k1}
\end{subfigure}\hfill
\begin{subfigure}[t]{.485\textwidth}
\centering
\includegraphics[width=\linewidth]{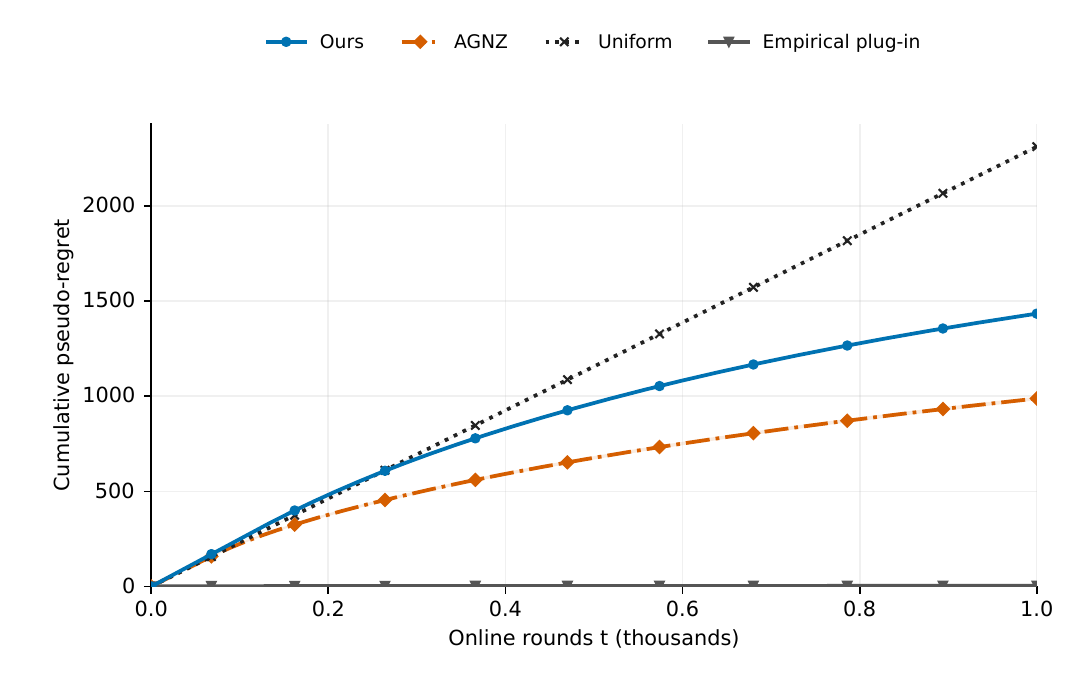}
\caption{heterogeneous-Uniform, $n=50$, $k=5$.}\label{fig:cont-prophet-k5}
\end{subfigure}
\caption{Computational results for prophet inequalities.}\label{fig:prophet-results}
\end{figure*}

\paragraph{Results.} On the discrete instances, our algorithm has the lowest mean endpoint regret among the four methods in every tested configuration; see \Cref{fig:prophet-discrete,fig:prophet-multiple} for representative curves. At $n=500$ and $T=320{,}000$, it reduces mean endpoint regret relative to AGNZ by approximately $46\%$ for $k=1$ and $89\%$ for $k=5$. The same confidence construction therefore performs well for both single-choice and multiple-choice stopping problems.

\begin{figure*}[t]
\centering
\includegraphics[width=.95\textwidth]{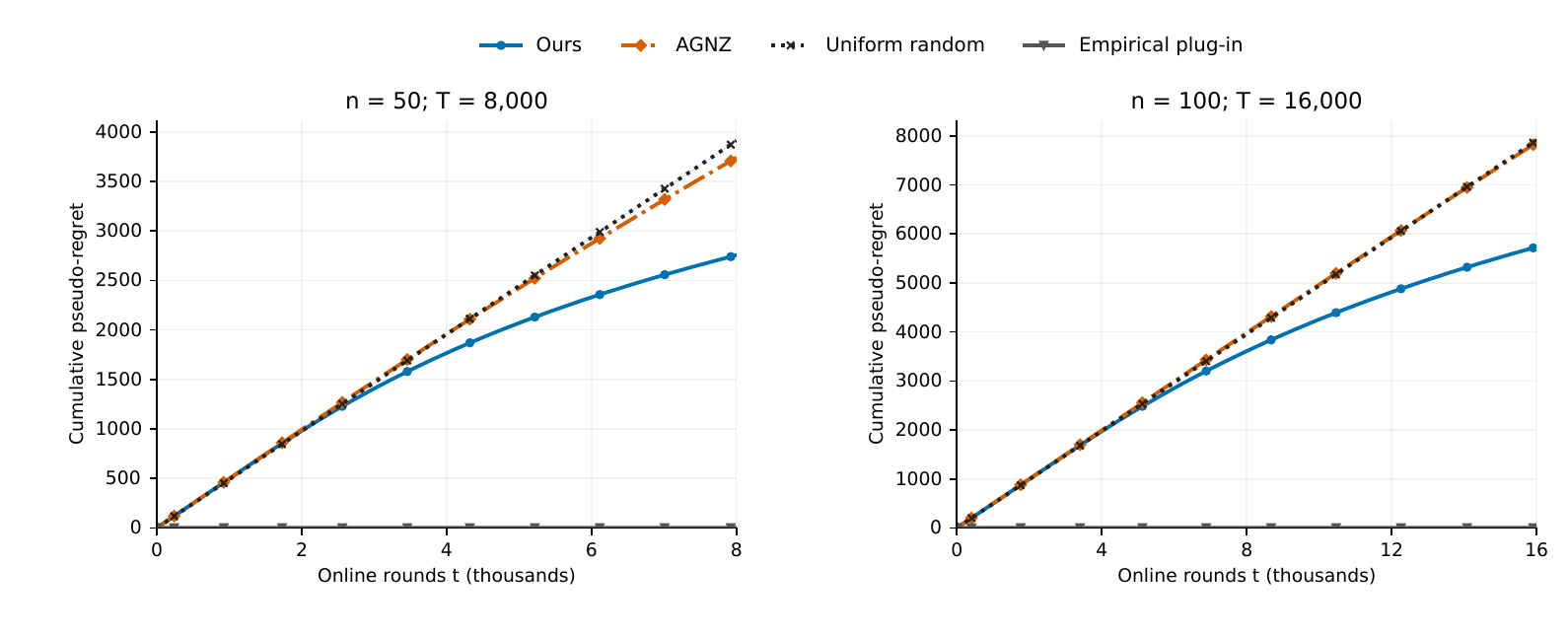}
\caption{Computational results for Uniform/Beta prophet instances with $k=1$.}\label{fig:uniform-beta-prophet}
\end{figure*}

Empirical plug-in performs better on both continuous families (\Cref{fig:cont-prophet-k1,fig:cont-prophet-k5,fig:uniform-beta-prophet}), possibly reflecting the particular distributions selected, while our framework achieves lower mean endpoint regret than both empirical plug-in and AGNZ on every tested finite-support instance. Among the two optimistic methods, AGNZ has lower mean endpoint regret than our algorithm at $n=50$ and $n=100$ for every tested capacity in the heterogeneous-Uniform family. At $n=50$ and $T=1000$, our algorithm incurs approximately $16\%$ more regret for $k=1$ and $45\%$ more for $k=5$, while both methods improve on Uniform. Our algorithm has lower mean endpoint regret at the larger tested dimensions, so these comparisons describe early learning behavior and do not establish a long-run ranking.

On the Uniform/Beta family, our algorithm has lower mean endpoint regret than AGNZ and Uniform in all seven reported configurations. For $k=1$, the reduction relative to AGNZ is approximately $26\%$ at $n=50$, $T=8000$, and $27\%$ at $n=100$, $T=16000$; see \Cref{fig:uniform-beta-prophet}. These experiments illustrate finite-horizon behavior on the lower-bound family. Additional plots are reported in Appendix~\ref{app:prophet}.

\subsection{Pandora's Box}

\paragraph{Instances.} There are $n$ independent boxes, each with inspection cost $1.5/n$. The learner sequentially chooses which boxes to open and when to stop, observing only the opened values. Its reward is the maximum value among the opened boxes minus the total inspection cost among them, with recall and outside option zero. The exact offline oracle uses Weitzman's reservation-value policy \citep{weitzman1979optimal}: the reservation value $z_i$ satisfies $\EE[(X_i-z_i)^+]=1.5/n$, and boxes are opened in decreasing order of $z_i$ until the best available value is at least the next reservation value. We use the same confidence construction as in the preceding applications.
We note that we allow the additional policy that stops immediately for reward zero. Adding this policy preserves strong monotonicity and the regret guarantee. Every instance below has a box with $\EE[X_i]>1.5/n$, so the optimal policy inspects at least one box and the two formulations have the same benchmark.

For the discrete instances, rewards have support $\{0, 0.25, 0.6, 1\}$. There are four approximately equally represented box types, randomly permuted once per replication. Their probabilities at the three positive support points are $(2,1,4)/n$, $(2,6,1)/n$, $(8,3,0.5)/n$, and $(4,4,1.7)/n$, respectively, with the remaining mass at zero. We use $n\in\{50,100,500,1000\}$. The Uniform baseline opens boxes in a uniformly random order, stopping upon observing value $1$ or exhausting the boxes.

For the continuous instances, we use the same heterogeneous-Uniform and Uniform/Beta distributions as in the prophet experiments, with $n\in\{50,100,500,1000\}$. In the Uniform/Beta family, $k\in\{1,2,5\}$ specifies the number of Uniform boxes. The Uniform baseline opens every box and serves as a weak control. 

\begin{figure*}[t]
\centering
\begin{subfigure}[t]{.485\textwidth}
\centering
\includegraphics[width=\linewidth]{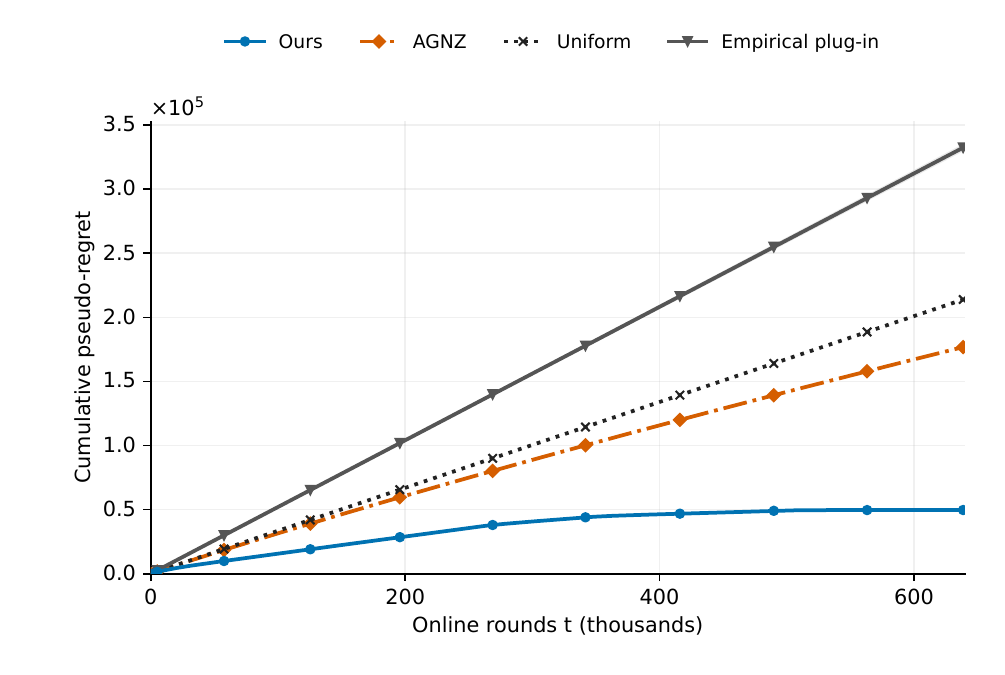}
\caption{Discrete, $n=500$.}\label{fig:pandora-discrete}
\end{subfigure}\hfill
\begin{subfigure}[t]{.485\textwidth}
\centering
\includegraphics[width=\linewidth]{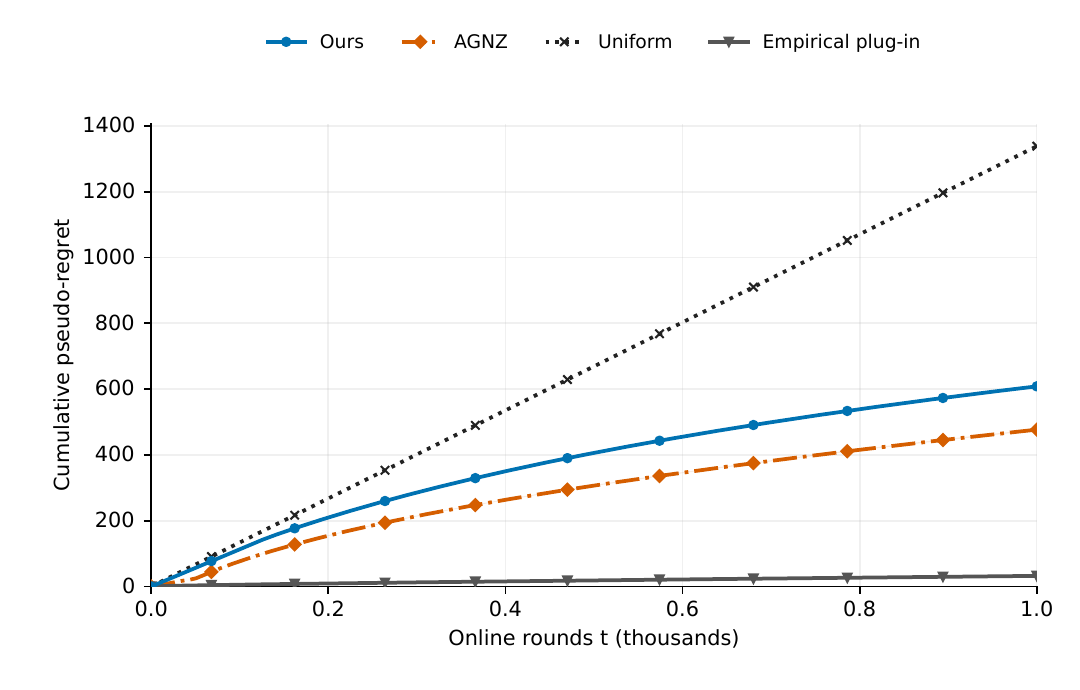}
\caption{heterogeneous-Uniform, $n=50$.}\label{fig:pandora-continuous}
\end{subfigure}
\caption{Computational results for Pandora's box.}\label{fig:pandora-results}
\end{figure*}

\paragraph{Results.} On the discrete instances, our algorithm has the lowest mean endpoint regret among the four methods at every tested dimension; see \Cref{fig:pandora-discrete}. At $n=500$ and $T=640{,}000$, its mean endpoint regret is approximately $72\%$ lower than AGNZ's. The framework therefore also performs well in an application where the oracle determines both the inspection order and the stopping decision.

Empirical plug-in also has the lowest mean endpoint regret on every reported continuous Pandora instance. Among the two optimistic methods, AGNZ has lower mean endpoint regret than our algorithm at three of the four tested dimensions in the heterogeneous-Uniform family; see \Cref{fig:pandora-continuous}. 
At $n=50$ and $T=1000$, our algorithm incurs approximately $28\%$ more regret than AGNZ. At $n=500$, the two mean endpoint regrets differ by less than $1\%$. All four configurations use $T=1000$, so these comparisons concern early learning; at $n=1000$, AGNZ spends the entire horizon in initialization.

\begin{figure*}[t]
\centering
\includegraphics[width=.95\textwidth]{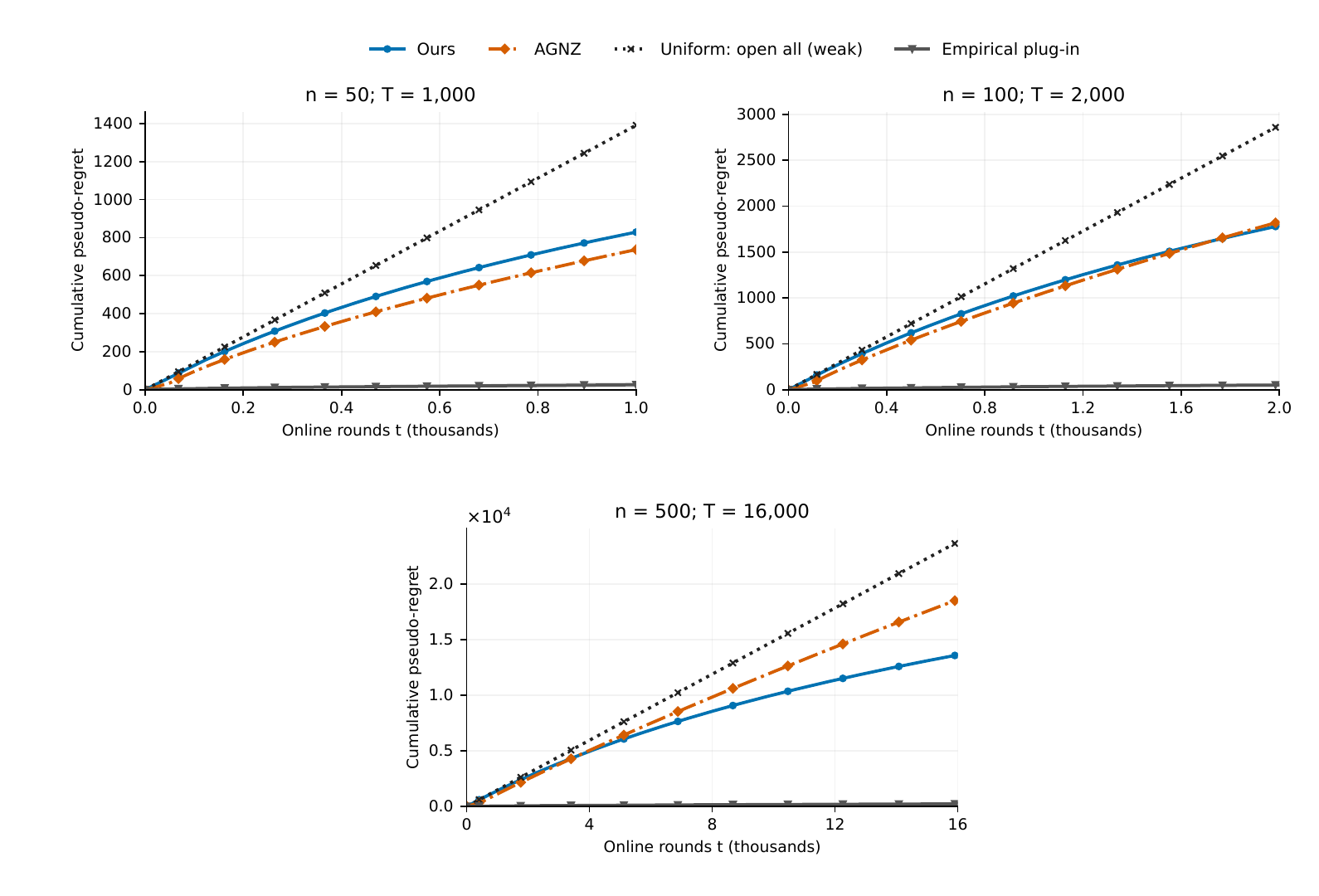}
\caption{Computational results for Uniform/Beta Pandora instances with $k=5$.}\label{fig:uniform-beta-pandora}
\end{figure*}

The Uniform/Beta family also exhibits both outcomes; see \Cref{fig:uniform-beta-pandora}. Our algorithm has lower mean endpoint regret than AGNZ in six of the seven reported configurations. With $k=5$, it incurs approximately $12\%$ more regret at $n=50$, $T=1000$, but approximately $27\%$ less regret at $n=500$, $T=16000$.
Additional plots are reported in Appendix~\ref{app:pandora}.

%% file: conclusion.tex
\section{Concluding Remarks}
We give oracle-efficient regret guarantees for monotone stochastic optimization under semi-bandit feedback. With an exact oracle witnessing strong monotonicity, \Cref{thm:main} gives expected regret \(\widetilde O(G\sqrt{nT})\), directly resolving the open question of \citet{agarwal2026semibandit} about improved dimension dependence under strong monotonicity. Our bound improves their linear dependence on \(n\) to \(\sqrt n\).
The matching lower bound in \Cref{thm:strong-regret-lower-bound} holds even under full feedback, so this dependence on \(n\) and \(T\) is minimax optimal up to logarithmic factors for \(T\ge n\). The same upper bound extends to expected \(\alpha\)-regret for approximation oracles satisfying the corresponding transfer condition. These guarantees allow arbitrary distributions on bounded support, require oracle calls only on finitely supported product distributions, and use no additional initialization samples.

We also improve the regret guarantee under monotonicity without an oracle transfer assumption. For monotone maximization and minimization problems with objectives in \([0,1]\), a uniform separate-variation bound \(V_0\) on the oracle's output policies gives expected \(\alpha\)-regret \(\widetilde O((1+V_0)n^{3/4}\sqrt T)\). Expressing our analysis and that of \citet{agarwal2026semibandit} under the same variation condition shows that, for fixed \(V_0>0\), our bound improves the polynomial dimension dependence from \(n\) to \(n^{3/4}\). The lower bound in \Cref{thm:weak-bv-exact-exponent}, at horizons \(T=\Theta(n^{3/2})\), shows that the exponent \(3/4\) is sharp for ordinary-regret guarantees of this form that hold uniformly for \Cref{alg:strong-up} over admissible problems and exact oracle selections. Without the variation condition, an exact oracle selection can cause the same algorithm to incur linear regret even on a one-item instance (\Cref{thm:weak-monotonicity-linear-regret}).

Two natural questions remain. First, what is the minimax regret under weak monotonicity and uniform separate bounded variation? For objectives in \([0,1]\) and fixed \(V_0>0\), can a different learning algorithm with access to an exact oracle satisfying this variation condition achieve expected regret \(\widetilde O(\sqrt{nT})\)? Second, under strong monotonicity, can the logarithmic gap between our upper and lower bounds be removed? In particular, can one achieve expected regret \(O(G\sqrt{nT})\) for \(T\ge n\)?

%% file: AIUseStatement.tex
\section*{AI Use Statement}
In this paper, the vast majority of mathematical arguments were developed by the authors. The paper was written by the authors. AI models, including ChatGPT and Claude, were used in the following ways:

\begin{itemize}
    \item to assist with the implementation of numerical experiments designed by the authors, including package selection, code organization, debugging, and restructuring of experimental code;

    \item to transcribe into \LaTeX{} mathematical arguments about the regret upper bounds for strongly and weakly monotone instances that had already been fully developed in the authors' manuscript;

    \item to propose the ramp term used in Equation~\ref{eq:app-wbv-lower-reward} and Lemma~\ref{lem:app-wbv-identity-optimal}, which makes the constructed reward function weakly monotone while preserving sufficient variation to keep the local regret bounded below over \(O(n^{3/2})\) rounds;

    \item to assist with literature searches and comparisons relevant to the topic of this paper;

    \item to check and simplify other proofs developed by the authors;

    \item to provide advice on expressions and phrasing during the writing of this paper.
\end{itemize}

All AI-assisted mathematical arguments, computations, literature references, and textual revisions were subsequently examined and rigorously checked by the authors. The authors take full responsibility for the final mathematical claims, proofs, citations, code, computational results, and presentation.

%% file: weak_bv_appendix.tex
\section{Proofs for Monotonicity with Bounded Variation}
\label{app:weak-bv-proof}

We first prove the upper and lower bounds in \Cref{thm:weak-bv-exact-exponent}, then establish the comparison with the analysis of \citet{agarwal2026semibandit}.

\subsection{Upper bound under Separate Bounded Variation}\label{sec:app-wbv-upper-bound}
We prove the up-monotone maximization case of the upper bound in \Cref{thm:weak-bv-exact-exponent}; the other cases follow as described in \Cref{sec:weak-monotone}. The oracle $\mathcal O$ is a possibly randomized $\alpha$-approximation oracle taking values in $\mathcal N\subseteq\cC$, where $\sup_{\sigma\in\mathcal N}\operatorname{TV}_{\rm sep}(r_\sigma)\le V_0$. Its approximation guarantee holds in expectation over its internal randomness, and $\OPT$ is defined over the full feasible class $\cC$. Throughout the proof, $C$ denotes a universal positive constant whose value may change between occurrences (we do this to keep the exposition clean).

Fix $T\ge\max\{n,2\}$ and set $\delta=1/(T+1)$. For the analysis, associate each item $i$ with an independent sequence $(X_{i,m})_{m\ge1}$ of i.i.d.\ samples from $D_i$, with the sequences independent across items. Whenever the algorithm probes item $i$, reveal the next unused sample in its sequence. By nonanticipation, this generates the same distribution of observed transcripts as the original model.

Let $\widehat D_{i,m}$ be the empirical distribution of $X_{i,1},\ldots,X_{i,m}$. Define
\begin{equation}\label{eq:app-wbv-optimistic-marginal}
\begin{aligned}
\ell_T&:=\log\bigl(2nT^2(T+1)\bigr),\\
E_{i,0}&:=\delta_1,\qquad E_{i,m}:=\mathsf{Sh}_{4\ell_T/m}(\widehat D_{i,m})\quad(m\ge1).
\end{aligned}
\end{equation}
These are the optimistic marginals used by \Cref{alg:strong-up}. Writing $N_i^t$ for the number of observations of item $i$ before round $t$, the algorithm selects
\begin{equation}\notag
\widetilde{\mathbf E}^{\,t}:=\prod_{i=1}^n E_{i,N_i^t},
\qquad
\sigma^t:=\mathcal O(\widetilde{\mathbf E}^{\,t};\xi^t),
\end{equation}
where each $\xi^t$ is fresh oracle randomness, independent of the past and the sample sequences. Initially $N_i^1=0$. If $S^t$ is the set of items probed in round $t$, then $N_i^{t+1}=N_i^t+\one\{i\in S^t\}$.

Let $\mathcal H_{t-1}$ denote the information available before the oracle call in round $t$, and write $\EE_t[\cdot]:=\EE[\cdot\mid\mathcal H_{t-1}]$. Thus $\widetilde{\mathbf E}^{\,t}$ is determined by $\mathcal H_{t-1}$, while $\sigma^t$ also depends on the fresh randomness $\xi^t$.

For a sufficiently large universal constant $C_0$, put
\begin{equation}\label{eq:app-wbv-def-rho}
\rho_0:=1,\qquad \rho_m:=\min\left\{1,\frac{C_0\ell_T}{m}\right\}\quad(m\ge1).
\end{equation}

The following lemma controls the optimistic marginals uniformly over coordinates, sample sizes, and intervals. Intervals may have arbitrary open or closed endpoints, including singletons.

\begin{lemma}[Relative interval concentration]
\label{lem:app-wbv-shifted-interval}
There is a universal choice of $C_0\ge1$ in \eqref{eq:app-wbv-def-rho} such that, with probability at least $1-1/T$, simultaneously for every $i\in[n]$ and $0\le m\le T$,
\begin{itemize}
    \item $D_i\sd E_{i,m},$ and 
    \item $|E_{i,m}(I)-D_i(I)|\le C\left(\sqrt{\rho_mD_i(I)}+\rho_m\right)$ for every interval $I\subseteq[0,1]$, where $C$ is universal.
\end{itemize}
\end{lemma}

\begin{proof}
Applying \Cref{thm:optimistic-distributions} with $\delta=1/(T+1)$ gives $\lambda=\ell_T$ and hence $D_i\sd E_{i,m}$ simultaneously for all $i\in[n]$ and $m\in[T]$, except on an event of probability at most $1/(T+1)$. For $m=0$, the stochastic order holds deterministically because $E_{i,0}=\delta_1$.

We next establish the interval estimate. Put $\eta_m:=\ell_T/m$ for $m\ge1$ and $\delta_0:=2e^{-\ell_T}=1/[nT^2(T+1)]$. The class of intervals has shatter coefficient at most $(2m+1)^2$ on $2m$ points. The relative VC inequalities in \citet[Corollary~2]{portier2026relative}, together with
\[
\log\left(\frac{6(2m+1)^2}{\delta_0}\right)\le C\ell_T\qquad(m\le T),
\]
therefore imply that, for each fixed $i\in[n]$ and $m\in[T]$, with failure probability at most $\delta_0$,
\begin{equation}\label{eq:app-wbv-relative-empirical-interval}
|\widehat D_{i,m}(I)-D_i(I)|\le C\left(\sqrt{\eta_mD_i(I)}+\eta_m\right)
\end{equation}
simultaneously for every interval $I$. Taking limits of intervals with rational endpoints covers all endpoint conventions, including singletons. A union bound over $i$ and $m$ makes \eqref{eq:app-wbv-relative-empirical-interval} simultaneous for all $i,m,I$, with failure probability at most $nT\delta_0=1/[T(T+1)]$.

It remains to control the change caused by shading. Recall from \eqref{eq:shade-scalar} that $s_\gamma(u) = \max\{0, u - \sqrt{2\gamma u(1-u)} - \gamma\}$. For $\gamma > 0$ and $u\in[0,1]$, define $b_\gamma(u) := \gamma + \sqrt{2\gamma u(1-u)}$ and $d_\gamma(u) := u -s_\gamma(u) = \min\{u,b_\gamma(u)\}.$
Since $\sqrt{2\gamma u}\le u+\gamma/2$, we have
\[
0\le b_\gamma(u)-d_\gamma(u)=(b_\gamma(u)-u)_+\le \frac{3\gamma}{2}.
\]
Moreover, observe that for $0\le u\le v\le1$, we have $|b_\gamma(v) - b_\gamma(u)| \le \sqrt{2 \gamma(v-u)}$,
because $u\mapsto u(1-u)$ is $1$-Lipschitz and $|\sqrt a-\sqrt b|\le\sqrt{|a-b|}$. Consequently,
\begin{equation}\label{eq:app-wbv-shading-modulus}
|d_\gamma(v)-d_\gamma(u)|\le C\left(\sqrt{\gamma(v-u)}+\gamma\right).
\end{equation}

Let $P$ be any distribution on $[0,1]$. If an interval $I$ does not contain $1$, write $P(I)=v-u$ using the appropriate CDF values or left limits. Its probability changes under shading by $d_\gamma(u)-d_\gamma(v)$, which is controlled by \eqref{eq:app-wbv-shading-modulus}. If $I$ contains $1$, write $P(I)=1-u$. The endpoint convention in \eqref{eq:shade-cdf} makes the change equal to $d_\gamma(u)$, and $d_\gamma(u) \le b_\gamma(u) \le \gamma + \sqrt{2\gamma(1-u)}$.
Thus, in either case,
\begin{equation}\label{eq:app-wbv-shading-relative-interval}
|\mathsf{Sh}_\gamma(P)(I)-P(I)|\le C\left(\sqrt{\gamma P(I)}+\gamma\right).
\end{equation}

Work now on the event where \eqref{eq:app-wbv-relative-empirical-interval} holds uniformly. If $\rho_m=1$, including $m=0$, the desired bound is immediate because interval probabilities lie in $[0,1]$. Otherwise, fix $i,m,I$ and write $\eta:=\eta_m=\rho_m/C_0$, $p:=D_i(I)$, and $\widehat p:=\widehat D_{i,m}(I)$. Equation~\eqref{eq:app-wbv-relative-empirical-interval} gives
\[
\widehat p\le p+C\sqrt{\eta p}+C\eta\le C(p+\eta).
\]
Applying \eqref{eq:app-wbv-shading-relative-interval} with $P=\widehat D_{i,m}$ and $\gamma=4\eta$, followed by the triangle inequality, yields
\[|E_{i,m}(I)-D_i(I)|\le C \left(\sqrt{\eta \widehat p} + \sqrt{\eta p} + \eta\right) \le C\left(\sqrt{\eta p} + \eta\right) \le C\left(\sqrt{\rho_m p} + \rho_m\right).\]
Finally, the stochastic-order and empirical-concentration events fail with total probability at most
$\frac1{T+1}+\frac1{T(T+1)}=\frac1T.$
This proves the lemma.
\end{proof}

The next lemma converts an interval-probability bound into a bound on expected values for functions of bounded variation.

\begin{lemma}[BV interpolation]
\label{lem:app-wbv-bv-fourth-root}
Let $P,Q$ be distributions on $[0,1]$. Suppose that $K>0$ and $0<\rho\le1$ satisfy
\begin{equation}\label{eq:app-wbv-abstract-relative-interval}
|Q(I) - P(I)| \le K\left(\sqrt{\rho P(I)} + \rho\right)
\end{equation}
for every interval $I\subseteq[0,1]$, including singletons. If $g:[0,1] \to [0,1]$ has pointwise total variation at most $V<\infty$, then
\begin{equation}\notag
\left|\int g\,d(Q-P)\right| \le CK \left(\sqrt{V\rho}\, \operatorname{Var}_P(g)^{1/4} + V\rho\right),
\end{equation}
where $C$ is a universal constant and $\operatorname{Var}_P(g)$ is the variance of $g$ under distribution $P$.
\end{lemma}

\begin{proof}
First suppose that $g$ is constant on each cell of a finite ordered partition of $[0,1]$ into intervals, allowing singleton cells. Put $c:=\int g\,dP$, $h_+:=(g-c)_+$, and $h_-:=(c-g)_+$. Since $c$ lies between the minimum and maximum of $g$, each $h_\pm$ vanishes somewhere and has total variation at most $V$.

For either $h=h_+$ or $h=h_-$, let $N_h(u)$ be the number of interval components of $\{h>u\}$. Counting the changes of $\one\{h>u\}$ along the partition and integrating over $u$ gives
\begin{equation}\notag
2\int_0^\infty N_h(u)\,du=\operatorname{TV}(h)+h(0)+h(1)\le2\operatorname{TV}(h)\le2V.
\end{equation}
Here $h(0)+h(1)\le\operatorname{TV}(h)$ because $h$ vanishes at some point of $[0,1]$.

Apply \eqref{eq:app-wbv-abstract-relative-interval} to each component of $\{h>u\}$. Summing and using Cauchy--Schwarz gives
\begin{equation}\notag
|(Q-P)(\{h>u\})|\le K\left(\sqrt{\rho N_h(u)P(h>u)}+\rho N_h(u)\right).
\end{equation}
Using $h(x)=\int_0^\infty\one\{h(x)>u\}\,du$, integrate this bound over $u$ and apply Cauchy--Schwarz again. Since $\int_0^\infty N_h(u)\,du\le V$ and $\int_0^\infty P(h>u)\,du=\EE_Ph$, we obtain
\begin{equation}\notag
\left|\int h\,d(Q-P)\right| \le K\left(\sqrt{V\rho}\,(\EE_Ph)^{1/2}+V\rho\right) \le K\left(\sqrt{V\rho}\,\operatorname{Var}_P(g)^{1/4}+V\rho\right).
\end{equation}
The last inequality uses $\EE_Ph_\pm\le\EE_P|g-c|\le\operatorname{Var}_P(g)^{1/2}$. Since $g=c+h_+-h_-$ and $P,Q$ have equal total mass, adding the two bounds proves the result for these step functions.

For a general $g$, put $v(x):=\operatorname{TV}_{[0,x]}(g)$, with $v(0)=0$. Additivity of variation gives $|g(y)-g(x)|\le v(y)-v(x)$ whenever $x\le y$. Partition $[0,v(1)]$ into finitely many intervals of length at most $\varepsilon$. Their nonempty inverse images under $v$ form an ordered interval partition of $[0,1]$, possibly with singleton cells. Choose one point in each cell and define $g_\varepsilon$ on that cell to equal $g$ at the chosen point. Then $\|g_\varepsilon-g\|_\infty\le\varepsilon$ and $\operatorname{TV}(g_\varepsilon)\le\operatorname{TV}(g)$; the latter follows because the chosen points are ordered. Apply the step-function bound to $g_\varepsilon$ and let $\varepsilon\downarrow0$. Uniform convergence gives convergence of both integrals and of the variance under $P$, proving the result.
\end{proof}

We now use the BV interpolation lemma to compare the expected reward of a fixed policy under two product distributions.

\begin{proposition}[Policy stability]
\label{prop:app-wbv-one-round}
Let $\mathbf D=\prod_{i=1}^nD_i$ and $\mathbf E=\prod_{i=1}^nE_i$ be product distributions on $[0,1]^n$. Fix a feasible policy $\sigma$ with rewards in $[0,1]$ and $\operatorname{TV}_{\rm sep}(r_\sigma)\le V_0$. Suppose that, for every $i\in[n]$, the pair $(D_i,E_i)$ satisfies \eqref{eq:app-wbv-abstract-relative-interval} with parameter $\rho_i\in(0,1]$ and a common constant $K>0$. Let $q_i$ be the probability that $\sigma$ probes item $i$ under $\mathbf D$, and put $A:=\sum_iq_i\rho_i$ and $U:=\sum_iq_i\rho_i^{2/3}$. Then
\[
|R(\sigma\mid\mathbf E)-R(\sigma\mid\mathbf D)|\le C(1+K)^{4/3}(1+V_0)\left(U^{3/4}+U+A\right),
\]
where $C$ is a universal constant.
\end{proposition}

\begin{proof}
For every history $h$ consistent with $\sigma$, let $W(h)$ be the expected terminal reward obtained by continuing $\sigma$ from $h$, with the unobserved coordinates sampled independently from their marginals in $\mathbf E$. This defines $W(h)$ even when $h$ has zero probability under $\mathbf E$. At a terminal history, $W(h)$ is the reward determined by that history. In particular, $0\le W(h)\le1$ and $W(\emptyset) = R(\sigma\mid\mathbf E)$.

If the next item probed at $h$ is $i$, write $h\oplus(i,y)$ for the history obtained by appending the observation $(i,y)$, and put $g_h(y):=W(h\oplus(i,y))$. Then $W(h)=\int g_h\,dE_i$. Because item $i$ is unobserved at $h$, varying its value leaves the preceding decisions unchanged. Thus $g_h$ is an average of coordinate sections of $r_\sigma$, with the observations in $h$ fixed. Summing absolute increments along any partition and moving the sum inside this average gives $\operatorname{TV}_{[0,1]}(g_h)\le V_0$. We also have $0\le g_h\le1$.

Run an auxiliary execution of $\sigma$ under $\mathbf D$. Let $H_\ell$ be its history after $\ell$ probes, let $\tau\le n$ be its total number of probes, and write $W_\ell:=W(H_\ell)$. For $\ell<\tau$, let $i_\ell$ be the next item probed and define $e_\ell:=\int g_{H_\ell}\,d(E_{i_\ell}-D_{i_\ell})$ and $v_\ell:=\operatorname{Var}_{D_{i_\ell}}(g_{H_\ell})$. Throughout this proof, $\EE_{\mathbf D}$ denotes expectation over this auxiliary execution.

Conditionally on $H_\ell$, the next observation has distribution $D_{i_\ell}$, so $\EE_{\mathbf D}[W_{\ell+1}\mid H_\ell]=W_\ell-e_\ell$. Telescoping and using that $W_\tau$ is the realized reward gives
\begin{equation}\label{eq:app-wbv-simulation-telescoping}
R(\sigma\mid\mathbf E)-R(\sigma\mid\mathbf D)=\EE_{\mathbf D}\sum_{\ell<\tau}e_\ell.
\end{equation}
Put $S:=\EE_{\mathbf D}\sum_{\ell<\tau}|e_\ell|$ and $B:=\EE_{\mathbf D}\sum_{\ell<\tau}v_\ell$. The corresponding second-moment identity is $\EE_{\mathbf D}[W_{\ell+1}^2\mid H_\ell]=v_\ell+(W_\ell-e_\ell)^2$. Summing this identity gives
\[
B=\EE_{\mathbf D}[W_\tau^2]-W_0^2+\EE_{\mathbf D}\sum_{\ell<\tau}(2W_\ell e_\ell-e_\ell^2)\le1+2S,
\]
where the inequality uses $0\le W_\ell\le1$.

Applying \Cref{lem:app-wbv-bv-fourth-root} to each $g_{H_\ell}$ yields
$|e_\ell|\le CK\left(\sqrt{V_0}\,\rho_{i_\ell}^{1/2}v_\ell^{1/4}+V_0\rho_{i_\ell}\right).$
Since each item is probed at most once, $\EE_{\mathbf D}\sum_{\ell<\tau}\rho_{i_\ell}=A$ and $\EE_{\mathbf D}\sum_{\ell<\tau}\rho_{i_\ell}^{2/3}=U$. Applying H\"older's inequality with exponents $4/3$ and $4$, first along each execution and then over the auxiliary randomness, therefore gives $S \le CK\sqrt{V_0}\, U^{3/4}B^{1/4} + CKV_0A$.
Substituting $B\le1+2S$ and using $(1+2S)^{1/4}\le1+2^{1/4}S^{1/4}$, Young's inequality gives
\[
S\le \frac12S+CK\sqrt{V_0}\,U^{3/4}+CK^{4/3}V_0^{2/3}U+CKV_0A.
\]
Absorb $S/2$ into the left-hand side. Since $K,K^{4/3}\le(1+K)^{4/3}$ and $\sqrt{V_0},V_0^{2/3},V_0\le1+V_0$, the resulting bound on $S$ is the right-hand side of the proposition. Finally, \eqref{eq:app-wbv-simulation-telescoping} gives $|R(\sigma\mid\mathbf E)-R(\sigma\mid\mathbf D)|\le S$, completing the proof.
\end{proof}

\paragraph{Completing the upper bound.}
For a fixed policy $\sigma$, let $q_i(\sigma)$ be its probability of probing item $i$ under $\mathbf D$. In round $t$, put $q_{t,i}:=\EE_t[q_i(\sigma^t)]$ and $\Delta_t:=\alpha\OPT(\mathbf D)-\EE_t[R(\sigma^t\mid\mathbf D)]$, and define
\[
A_t:=\sum_{i=1}^n q_{t,i}\rho_{N_i^t},\qquad U_t:=\sum_{i=1}^n q_{t,i}\rho_{N_i^t}^{2/3}.
\]
These quantities are determined by $\mathcal H_{t-1}$; in particular, $q_{t,i}$ averages over the fresh oracle randomness.

Let $G_t$ be the event that the stochastic-order and interval inequalities in \Cref{lem:app-wbv-shifted-interval} hold for every item at its current sample size $N_i^t$, with the universal constant from that lemma. This event is determined by $\mathcal H_{t-1}$, and the lemma gives $\mathbb P(G_t^c)\le1/T$. We apply the oracle guarantee conditional on $\mathcal H_{t-1}$; the uniform concentration event is used only to bound $\mathbb P(G_t^c)$.

On $G_t$, up-monotonicity gives $\OPT(\mathbf D)\le\OPT(\widetilde{\mathbf E}^{\,t})$. The oracle guarantee, \Cref{prop:app-wbv-one-round}, and Jensen's inequality over the oracle randomness therefore yield
\[
\Delta_t \le \EE_t\!\left[R(\sigma^t\mid\widetilde{\mathbf E}^{\,t})-R(\sigma^t\mid\mathbf D)\right] \le C(1+V_0)\left(U_t^{3/4}+U_t+A_t\right).
\]
Here the proposition applies to every returned policy because $\sigma^t\in\mathcal N$. Since $\Delta_t\le1$ always, taking expectations and summing gives
\begin{equation}\label{eq:app-wbv-regret-before-counting}
\EE[\alpha\text{-}R(T)]=\sum_{t=1}^T\EE[\Delta_t]\le C(1+V_0)\sum_{t=1}^T\EE\!\left[U_t^{3/4}+U_t+A_t\right]+1.
\end{equation}

It remains to bound these sums. Put $I_{t,i}:=\one\{i\in S^t\}$. By nonanticipation and independence of the sample sequences, conditional on $\mathcal H_{t-1}$ and the fresh oracle randomness, the vector of next unused samples has product distribution $\mathbf D$. Consequently, $\EE_t[I_{t,i}]=q_{t,i}$. For every nonnegative deterministic function $a$ on the nonnegative integers,
\begin{equation}\label{eq:app-wbv-adaptive-counting}
\sum_{t=1}^T\EE[q_{t,i}a(N_i^t)] = \EE\sum_{t=1}^T I_{t,i}a(N_i^t) = \EE\sum_{m=0}^{N_i^{T+1}-1}a(m)
\le\sum_{m=0}^{T-1}a(m).
\end{equation}
The second equality holds because each probe advances the observation count by exactly one.

Applying \eqref{eq:app-wbv-adaptive-counting} with $a(m)=\rho_m$ and $a(m)=\rho_m^{2/3}$, and using \eqref{eq:app-wbv-def-rho}, gives
\[
\sum_{t=1}^T\EE[A_t]\le Cn\ell_T\log(eT),\qquad
\sum_{t=1}^T\EE[U_t]\le Cn\ell_T^{2/3}T^{1/3}.
\]
For the second estimate, use $\rho_m^{2/3}\le(C_0\ell_T/m)^{2/3}$ for $m\ge1$ and $\sum_{m=1}^{T-1}m^{-2/3}\le3T^{1/3}$; the $m=0$ contribution is absorbed into the constant. We apply Jensen's inequality in each round and then H\"older's inequality across rounds to obtain
\[
\sum_{t=1}^T\EE[U_t^{3/4}]
\le T^{1/4}\left(\sum_{t=1}^T\EE[U_t]\right)^{3/4}
\le Cn^{3/4}\ell_T^{1/2}\sqrt T.
\]
Substituting these estimates into \eqref{eq:app-wbv-regret-before-counting} yields
\begin{equation}\label{eq:app-wbv-detailed-upper-bound}
\EE[\alpha\text{-}R(T)]\le C(1+V_0)\left[n^{3/4}\ell_T^{1/2}\sqrt T+n\ell_T^{2/3}T^{1/3}+n\ell_T\log(eT)\right]+1.
\end{equation}

Finally, put $P_T:=\log(2nT)$, so $\ell_T\le CP_T$. If $T\le n^{3/2}P_T^4$, the trivial bound gives $\EE[\alpha\text{-}R(T)]\le T\le n^{3/4}\sqrt T\,P_T^2$. Otherwise, the first and third terms in \eqref{eq:app-wbv-detailed-upper-bound} are at most $Cn^{3/4}\sqrt T\,P_T^2$, using $T\ge n$ for the third term. For the remaining term,
\[
\frac{n\ell_T^{2/3}T^{1/3}}{n^{3/4}\sqrt T\,P_T^2}
\le C\frac{n^{1/4}}{T^{1/6}P_T^{4/3}}
\le C.
\]
Thus, in both cases,
\[
\EE[\alpha\text{-}R(T)]\le C(1+V_0)n^{3/4}\sqrt T\,\log^2(2nT)+1.
\]
This proves the upper bound in \Cref{thm:weak-bv-exact-exponent}.

\subsection{Lower Bound under Separate Bounded Variation}

We now prove the lower bound in \Cref{thm:weak-bv-exact-exponent}.

\begin{theorem}\label{thm:app-wbv-lower-bound}
For every fixed $V_0>0$, there exist constants $c_{V_0}>0$ and $n_0(V_0)$ such that the following holds for every $n\ge n_0(V_0)$. There is an up-monotone maximization problem with $n$ items, rewards in $[0,1]$, and a distribution-independent feasible policy class $\cC_n$ satisfying
\(
\sup_{\sigma\in\cC_n}\operatorname{TV}_{\rm sep}(r_\sigma)\le V_0,
\)
together with an exact oracle $\mathcal O_n$ defined on all product distributions on $[0,1]^n$, such that under $\mathbf D=\prod_{i=1}^n\operatorname{Unif}[0,1]$, at a horizon $T_n=\Theta(n^{3/2})$, \Cref{alg:strong-up}, using $\mathcal O_n$ and $\delta=1/(T_n+1)$, satisfies
$\EE[R(T_n)]\ge c_{V_0}n^{3/2}.$
\end{theorem}

We prove the theorem by constructing policies that probe all $n$ items in a fixed order. Their rewards depend on which of $k=\Theta(\sqrt n)$ intervals contains each observation. We first define the policy class and verify up-monotonicity, the reward range, and the variation bound. We then specify an exact oracle on all product distributions and characterize its output on the optimistic distributions encountered by the algorithm. Finally, we show that, with probability tending to one, the selected policies incur ordinary pseudo-regret bounded below by a positive constant depending only on $V_0$ in each of $\Theta(n^{3/2})$ rounds. The oracle selection is part of the construction.

\paragraph{The policy class.}
We use a cosine term to combine empirical errors across items. We also allow downward maps of the interval indices to ensure up-monotonicity. An additional increasing term will make the identity maps optimal for the distributions used in the regret calculation.

Let $k$ be the largest even integer not exceeding $\sqrt n$. We assume that $n$ is sufficiently large that $k\ge2$ and $k\ge\sqrt n/2$. Partition $[0,1]$ into the intervals $I_1:=[0,1/k]$ and $I_r:=((r-1)/k,r/k]$ for $2\le r\le k$, and let $b(x)=r$ when $x\in I_r$. Define the balanced sign functions and downward maps by
\[
\mathcal S_k:=\left\{s:[k]\to\{-1,1\}:\sum_{r=1}^k s(r)=0\right\}, \,
\mathcal T_k:=\left\{\tau:[k]\to[k]:\tau\text{ is nondecreasing and }\tau(r)\le r\right\}.
\]
In particular, $\mathcal T_k$ contains the identity map.
Then, we define
\begin{equation}\label{eq:app-wbv-lower-parameters}
a:=\frac1k,\qquad m_0:=\frac{n(k+1)}2,\qquad W:=A_0k\sqrt n,
\end{equation}
where $A_0$ is an absolute constant chosen below. Define the increasing function
\begin{equation}\label{eq:app-wbv-ramp-definition}
H(z):=\left[\frac12+\frac{z-m_0}{2W}\right]_0^1,\qquad [u]_0^1:=\min\{1,\max\{0,u\}\}.
\end{equation}

For each $\mathbf s\in\mathcal S_k^n$, $\theta\in[0,2\pi]$, and $\boldsymbol\tau\in\mathcal T_k^n$, define a policy $\sigma=\sigma_{\mathbf s,\theta,\boldsymbol\tau}$ that probes items $1,\ldots,n$ in that order and receives terminal reward
\begin{equation}\label{eq:app-wbv-lower-reward}
\begin{aligned}
r_\sigma(\xvec)
={}&\frac12+\varepsilon\cos\left(\theta+a\sum_{i=1}^n s_i(\tau_i(b(x_i)))\right)\\
&+\Lambda_{\rm ramp}\varepsilon\left(H\left(\sum_{i=1}^n\tau_i(b(x_i))\right)-\frac12\right).
\end{aligned}
\end{equation}
Let $\cC_n$ consist of all these policies, each including its terminal reward rule in \eqref{eq:app-wbv-lower-reward}.

We first choose $A_0$ sufficiently large as required by \Cref{lem:app-wbv-center-ramp}. We then choose $\Lambda_{\rm ramp}>0$, depending only on $A_0$, as required by \Cref{lem:app-wbv-identity-optimal}, and set
\begin{equation}\label{eq:app-wbv-epsilon-choice}
\varepsilon:=\min\left\{\frac1{2+\Lambda_{\rm ramp}},\frac{V_0}{2+\Lambda_{\rm ramp}/(2A_0)}\right\}>0.
\end{equation}
These constants are independent of $n$, the horizon, and the input distributions.

\begin{lemma}\label{lem:app-wbv-lower-structural-properties}
The class $\cC_n$ is distribution-independent and consists of deterministic, nonanticipating policies with rewards in $[0,1]$ and $\sup_{\sigma\in\cC_n}\operatorname{TV}_{\rm sep}(r_\sigma)\le V_0$. The maximum defining $\OPT(\mathbf P)$ is attained for every product distribution $\mathbf P$ on $[0,1]^n$, and the problem is up-monotone.
\end{lemma}

\begin{proof}
Every policy probes each item exactly once and then receives a reward determined by the observed values. The class is fixed independently of the input distributions. Since the cosine lies in $[-1,1]$ and $H-1/2$ lies in $[-1/2,1/2]$, the choice of $\varepsilon$ in \eqref{eq:app-wbv-epsilon-choice} gives $0\le r_\sigma(\xvec)\le1$.

Fix a policy $\sigma=\sigma_{\mathbf s,\theta,\boldsymbol\tau}$ and all coordinates except coordinate $i$. As $x_i$ increases from $0$ to $1$, the value $s_i(\tau_i(b(x_i)))$ changes at most $k-1$ times, with each change having magnitude $2$. Each such change alters the cosine by at most $2\sin a$, so the cosine section has variation at most $2(k-1)\sin(1/k)\le2$. Moreover, $\tau_i\circ b$ is nondecreasing and $H$ is nondecreasing and $1/(2W)$-Lipschitz. The section involving $H$ therefore has variation at most $(k-1)/(2W)\le1/(2A_0\sqrt n)$. Hence
\[
\operatorname{TV}_{\rm sep}(r_\sigma)\le\varepsilon\left(2+\frac{\Lambda_{\rm ramp}}{2A_0\sqrt n}\right)\le V_0.
\]

For every fixed pair $(\mathbf s,\boldsymbol\tau)$ and every product distribution $\mathbf P$, the expected reward is continuous in $\theta\in[0,2\pi]$, so it attains a maximum. Since there are finitely many choices of $(\mathbf s,\boldsymbol\tau)$, the maximum over $\cC_n$ is also attained.

It remains to prove up-monotonicity. We first show how to transform a larger categorical distribution into a smaller one using downward maps. Let $p\sd q$ be distributions on $[k]$, so $F_p(r)\ge F_q(r)$ for every $r$. For $v\in[0,1]$, define $F_p^{-1}(v):=\min\{r\in[k]:F_p(r)\ge v\}$. For $u\in[0,1]$, put
\[
\kappa_u(r):=F_p^{-1}\bigl(F_q(r-1)+u q_r\bigr),\qquad F_q(0):=0.
\]
The argument of $F_p^{-1}$ is nondecreasing in $r$ and is at most $F_q(r)\le F_p(r)$. Thus $\kappa_u$ is nondecreasing and $\kappa_u(r)\le r$, so $\kappa_u\in\mathcal T_k$. If $Z\sim q$ and $U\sim\operatorname{Unif}[0,1]$ are independent, then $F_q(Z-1)+Uq_Z$ is uniform on $[0,1]$. Consequently, $\kappa_U(Z)$ has distribution $p$.

Now consider product distributions $\mathbf P\sd\mathbf Q$, and let $p_i=(P_i(I_r))_{r=1}^k$ and $q_i=(Q_i(I_r))_{r=1}^k$. Since $b$ is nondecreasing, $p_i\sd q_i$. Fix any policy $\sigma=\sigma_{\mathbf s,\theta,\boldsymbol\tau}\in\cC_n$. For each item $i$, let $\kappa_{i,u}$ be the map just constructed for $(p_i,q_i)$. Take independent $U_i\sim\operatorname{Unif}[0,1]$, independently of the observations. For each fixed realization of $\mathbf U$, replace $\tau_i$ by $\tau_i\circ\kappa_{i,U_i}$ to obtain a policy $\sigma_{\mathbf U}$. Compositions of maps in $\mathcal T_k$ remain in $\mathcal T_k$, so every $\sigma_{\mathbf U}$ belongs to $\cC_n$.

Under $\mathbf Q$, the transformed interval indices $\kappa_{i,U_i}(b(X_i))$ are independent and have distributions $p_i$. Averaging over the auxiliary variables $\mathbf U$ therefore gives
\(
R(\sigma\mid\mathbf P)=\EE_{\mathbf U}\!\left[R(\sigma_{\mathbf U}\mid\mathbf Q)\right]\le\OPT(\mathbf Q).
\)
Taking the maximum over $\sigma\in\cC_n$ proves $\OPT(\mathbf P)\le\OPT(\mathbf Q)$.
\end{proof}

To show that the identity maps are optimal in the relevant regime, we first bound how much the increasing term decreases when the interval indices are reduced.

\begin{lemma}[Loss in the increasing term]\label{lem:app-wbv-center-ramp}
Let $A_0\ge16$, with $m_0,W,H$ as defined in \eqref{eq:app-wbv-lower-parameters} and \eqref{eq:app-wbv-ramp-definition}. Let $(U_i,V_i)$, $i\in[n]$, be independent pairs satisfying $U_i,V_i\in[k]$ and $V_i\le U_i$ almost surely. Define $S:=\sum_iU_i$, $L:=\sum_i(U_i-V_i)$, and $\Delta:=\EE[L]$. If $|\EE[S]-m_0|\le W/8$, then
\[
\EE[H(S)-H(S-L)]\ge c_0\min\{\Delta/W,1\},
\]
where $c_0:=1/32$.
\end{lemma}

\begin{proof}
The claim is immediate if $\Delta=0$, so assume $\Delta>0$. Let $\mathcal G:=\{|S-m_0|\le W/4\}$ and $L_i:=U_i-V_i$. On $\mathcal G$, we have $H(S)\in[3/8,5/8]$. Since $L\ge0$, the definition of $H$ gives
\begin{equation}\label{eq:app-wbv-ramp-pointwise}
H(S)-H(S-L)=\min\left\{\frac{L}{2W},H(S)\right\}\ge\frac38\min\{L/W,1\}\qquad\text{on }\mathcal G.
\end{equation}

We next show that $\EE[L\mathbf1_{\mathcal G}]\ge\Delta/2$. Fix $i$ and a value $u$ with $\PP(U_i=u)>0$. By independence and the assumption on $\EE[S]$,
\(
|\EE[S\mid U_i=u]-m_0|\le W/8+k\le3W/16.
\)
The other coordinates remain independent under this conditioning. Applying Hoeffding's inequality at deviation $W/16$ therefore gives
\[
\PP(\mathcal G^c\mid U_i=u)\le2\exp\left(-\frac{W^2}{128nk^2}\right)=2e^{-A_0^2/128}\le\frac12,
\]
where the last inequality uses $A_0\ge16$. Independence of the pairs also gives $\PP(\mathcal G\mid U_i,V_i)=\PP(\mathcal G\mid U_i)$. Consequently,
\[
\EE[L\mathbf1_{\mathcal G}]
=\sum_i\EE\!\left[L_i\PP(\mathcal G\mid U_i,V_i)\right]
\ge\frac12\sum_i\EE[L_i]
=\frac{\Delta}{2}.
\]

Moreover, independence and $0\le L_i\le k$ give $\EE[L^2]=\Delta^2+\sum_i\operatorname{Var}(L_i)\le\Delta^2+k\Delta$. We use these two estimates and Cauchy--Schwarz to obtain
\[
\begin{aligned}
\EE[\min\{L/W,1\}\mathbf1_{\mathcal G}]
&\ge\EE\left[\frac{L}{W+L}\mathbf1_{\mathcal G}\right] \ge \frac{(\EE[L\mathbf1_{\mathcal G}])^2}{\EE[L(W+L)\mathbf1_{\mathcal G}]}\\
&\ge\frac{\Delta}{4(W+k+\Delta)}
\ge\frac1{12}\min\{\Delta/W,1\}.
\end{aligned}
\]
For the last inequality, use $k\le W$ and consider separately $\Delta\le W$ and $\Delta>W$. Finally, $H(S)-H(S-L)\ge0$ everywhere because $H$ is nondecreasing. Combining the preceding bound with \eqref{eq:app-wbv-ramp-pointwise} proves the result with $c_0=(3/8)(1/12)=1/32$.
\end{proof}

To describe the signs chosen by the oracle, we define the balanced discrepancy of a probability vector $p$ on $[k]$ as
\begin{equation}\label{eq:app-wbv-eta-definition}
\eta(p):=\max_{s\in\mathcal S_k}\sum_{r=1}^k s(r)p_r.
\end{equation}
Since $-s\in\mathcal S_k$ whenever $s\in\mathcal S_k$, we also have $\eta(p)=\max_{s\in\mathcal S_k}|\sum_r s(r)p_r|\ge0$.

\begin{lemma}[Optimality of the identity maps]\label{lem:app-wbv-identity-optimal}
For a sufficiently large choice of $\Lambda_{\rm ramp}$ depending only on $A_0$, the following holds. Let $\mathbf Q=\prod_{i=1}^nQ_i$ be any product distribution on $[0,1]^n$, let $U_i=b(X_i)$ for $\xvec\sim\mathbf Q$, and write $p_i=(Q_i(I_r))_{r=1}^k$. Suppose that
\begin{equation}\label{eq:app-wbv-Q-center-condition}
\left|\sum_{i=1}^n\EE[U_i]-m_0\right|\le W/8.
\end{equation}
Then an optimal policy in $\cC_n$ is obtained by choosing $\tau_i$ to be the identity map for every $i$, choosing $s_i\in\mathcal S_k$ with $q_i:=\EE[s_i(U_i)]=\eta(p_i)$, and taking
\begin{equation}\label{eq:app-wbv-optimal-phase}
\theta_{\mathbf Q}=-\sum_{i=1}^n\arctan(q_i\tan a)\pmod{2\pi},
\end{equation}
with the representative in $[0,2\pi)$.
\end{lemma}

\begin{proof}
First fix the signs $\mathbf s$ and consider identity maps. Put $q_i:=\EE[s_i(U_i)]$. Since the $U_i$ are independent and each $s_i$ takes values in $\{-1,1\}$,
\[
\EE\left[e^{\mathrm i a\sum_i s_i(U_i)}\right]
=\prod_i\EE\left[e^{\mathrm i a s_i(U_i)}\right]
=\prod_i(\cos a+\mathrm i q_i\sin a).
\]
Thus, maximizing over $\theta$ makes the expected cosine equal to $M(\mathbf q)$, where
\begin{equation}\label{eq:app-wbv-M-definition}
M(\mathbf z):=\prod_{i=1}^n(\cos^2a+z_i^2\sin^2a)^{1/2},\qquad \mathbf z\in[-1,1]^n.
\end{equation}

Now fix arbitrary maps $\tau_i\in\mathcal T_k$. Let $V_i:=\tau_i(U_i)$, $q_i':=\EE[s_i(V_i)]$, and $\Delta:=\sum_i\EE[U_i-V_i]$. After optimizing the phase, the expected cosine for these maps is $M(\mathbf q')$. Since $V_i\le U_i$ and the indices are integers,
\[
|q_i'-q_i|\le2\PP(U_i\ne V_i)\le2\EE[U_i-V_i].
\]
Moreover, $0<M(\mathbf z)\le1$. Differentiating \eqref{eq:app-wbv-M-definition} gives
\[
\left|\frac{\partial M}{\partial z_i}(\mathbf z)\right|
=M(\mathbf z)\frac{|z_i|\sin^2a}{\cos^2a+z_i^2\sin^2a}
\le\tan^2a\le C/k^2,
\]
where $C$ is an absolute constant. Applying this derivative bound between $\mathbf q$ and $\mathbf q'$, and using $0<M\le1$, we obtain
\begin{equation}\label{eq:app-wbv-cosine-improvement-bound}
[M(\mathbf q')-M(\mathbf q)]_+\le C\min\{\Delta/k^2,1\}.
\end{equation}

The pairs $(U_i,V_i)$ are independent, so \eqref{eq:app-wbv-Q-center-condition} and \Cref{lem:app-wbv-center-ramp} give
\[
\EE\left[H\left(\sum_iU_i\right)-H\left(\sum_iV_i\right)\right]\ge c_0\min\{\Delta/W,1\}.
\]
Since $k\ge\sqrt n/2$, we have $W/k^2=A_0\sqrt n/k\le2A_0$, and hence $\min\{\Delta/k^2,1\}\le2A_0\min\{\Delta/W,1\}$. Combining this with \eqref{eq:app-wbv-cosine-improvement-bound}, after optimizing the phase for each choice of maps, the difference between the expected reward with $\boldsymbol\tau$ and with identity maps is at most
$\varepsilon(2CA_0-\Lambda_{\rm ramp}c_0)\min\{\Delta/W,1\}.$
Choosing $\Lambda_{\rm ramp}\ge2CA_0/c_0$ makes this difference nonpositive. Thus identity maps are optimal for every fixed choice of signs.

It remains to choose the signs and phase. With identity maps, the increasing term is independent of $\mathbf s$, and each factor in $M(\mathbf q)$ is nondecreasing in $|q_i|$. By \eqref{eq:app-wbv-eta-definition} and the symmetry of $\mathcal S_k$, we can therefore maximize each factor by choosing $q_i=\eta(p_i)\ge0$. Finally, $\cos a>0$, so the complex factor $\cos a+\mathrm i q_i\sin a$ has argument $\arctan(q_i\tan a)$. The phase in \eqref{eq:app-wbv-optimal-phase} makes their product, after multiplication by $e^{\mathrm i\theta_{\mathbf Q}}$, positive real. It therefore attains the maximum expected cosine $M(\mathbf q)$, proving the lemma.
\end{proof}

\begin{lemma}[Exact oracle]\label{lem:app-wbv-measurable-selector}
There is a deterministic, measurable exact oracle $\mathcal O_n$ taking values in $\cC_n$, defined on all product distributions on $[0,1]^n$. Whenever its input satisfies \eqref{eq:app-wbv-Q-center-condition}, it returns an optimal policy with identity maps and the signs and phase specified in \Cref{lem:app-wbv-identity-optimal}.
\end{lemma}

\begin{proof}
Fix orderings of the finite sets $\mathcal S_k$ and $\mathcal S_k^n\times\mathcal T_k^n$. Consider any input $\mathbf Q=\prod_{i=1}^nQ_i$. For each pair $(\mathbf s,\boldsymbol\tau)$, put $q_i:=\sum_{r=1}^k Q_i(I_r)s_i(\tau_i(r))$. The phase calculation in the preceding proof shows that
\[
\theta_{\mathbf s,\boldsymbol\tau}(\mathbf Q):=-\sum_{i=1}^n\arctan(q_i\tan a)\pmod{2\pi},
\]
with the representative in $[0,2\pi)$, maximizes the expected reward for these signs and maps. The resulting expected reward is
\[
\frac12+\varepsilon M(\mathbf q)+\Lambda_{\rm ramp}\varepsilon\left(\EE_{\xvec\sim\mathbf Q}\left[H\left(\sum_i\tau_i(b(X_i))\right)\right]-\frac12\right),
\]
where $M$ is defined in \eqref{eq:app-wbv-M-definition}. Select the first pair attaining the largest of these values and return its policy with the optimizing phase. Every policy in $\cC_n$ is specified by one of these pairs and a phase, so this selection is exact for every product distribution $\mathbf Q$.

When \eqref{eq:app-wbv-Q-center-condition} holds, we instead use identity maps. For each item $i$, write $p_i=(Q_i(I_r))_{r=1}^k$ and select the first $s_i\in\mathcal S_k$ attaining $\eta(p_i)$. Choose the phase in \eqref{eq:app-wbv-optimal-phase}. By \Cref{lem:app-wbv-identity-optimal}, this selection is also exact.

Finally, represent each output policy by its parameters $(\mathbf s,\theta,\boldsymbol\tau)$. The candidate values and the quantity in \eqref{eq:app-wbv-Q-center-condition} are continuous functions of the cell probabilities $Q_i(I_r)$, and the phases are measurable functions of these probabilities. Since the cell probabilities are measurable functions of the input distributions, the finite comparisons and fixed tie-breaking rules give a measurable oracle.
\end{proof}

We use this fixed oracle $\mathcal O_n$ in the remaining analysis. Its domain includes the initial distribution $\prod_{i=1}^n\delta_1$ and every shaded empirical distribution produced by \Cref{alg:strong-up}.

We next estimate the balanced discrepancy of the shaded empirical distributions. We first bound the effect of shading in total variation, where $d_{\mathrm{TV}}(P,Q):=\sup_B|P(B)-Q(B)|$ and the supremum is over measurable sets.

\begin{lemma}[Total variation of shading]\label{lem:app-wbv-shift-tv}
For every finitely supported distribution $P$ on $[0,1]$ and every $\gamma>0$,
\[
d_{\mathrm{TV}}\bigl(P,\mathsf{Sh}_\gamma(P)\bigr)\le\sqrt{\gamma/2}+\gamma.
\]
\end{lemma}

\begin{proof}
Recall that $d_\gamma(u):=u-s_\gamma(u)=\min\{u,\sqrt{2\gamma u(1-u)}+\gamma\}$. Both functions in this minimum are concave, so $d_\gamma$ is concave. Consequently, it is nondecreasing up to a maximum and nonincreasing thereafter.

For $x<1$, the CDF gap is $F_P(x)-F_{\mathsf{Sh}_\gamma(P)}(x)=d_\gamma(F_P(x))$, and at $x=1$ it is zero. Thus the CDF gap also increases to a maximum and then decreases, including its final drop to zero at $1$. Reading the atoms of $P$ together with $1$ in increasing order, the signed masses of $P-\mathsf{Sh}_\gamma(P)$ therefore change sign at most once, from positive to negative. Their total positive mass equals the largest CDF gap, giving
\[
d_{\mathrm{TV}}\bigl(P,\mathsf{Sh}_\gamma(P)\bigr)
=\sup_x\bigl(F_P(x)-F_{\mathsf{Sh}_\gamma(P)}(x)\bigr)
\le\max_{u\in[0,1]}d_\gamma(u)
\le\sqrt{\gamma/2}+\gamma.
\]
\end{proof}

\begin{lemma}[Balanced empirical discrepancy]\label{lem:app-wbv-balanced-empirical}
Let $k\ge2$ be even, and let $\widehat p$ be the empirical distribution of $m\ge k$ independent uniform samples from $[k]$. There are universal constants $c,C>0$ such that
\[
\PP\left(c\sqrt{k/m}\le\eta(\widehat p)\le C\sqrt{k/m}\right)\ge1-Ce^{-ck}.
\]
\end{lemma}

\begin{proof}
Let $C_r$ be the number of samples in category $r$. Pair the categories as $(1,2),(3,4),\ldots,(k-1,k)$, assigning the positive sign to the more frequent category in each pair and the negative sign to the other. These signs are balanced, so
\[
\eta(\widehat p)\ge\frac{Z}{m},\qquad Z:=\sum_{j=1}^{k/2}|C_{2j-1}-C_{2j}|.
\]
For each pair, $Y_j:=C_{2j-1}-C_{2j}$ is a sum of $m$ independent random variables taking values $1$ and $-1$ with probability $1/k$ each, and $0$ otherwise. Thus
\[
\EE[Y_j^2]=\frac{2m}{k},\qquad \EE[Y_j^4]=\frac{2m}{k}+\frac{12m(m-1)}{k^2}\le14(m/k)^2,
\]
where the last inequality uses $m\ge k$. Applying the Paley--Zygmund inequality to $Y_j^2$ gives $\PP(|Y_j|\ge\sqrt{m/k})\ge1/14$, and hence $\EE[Z]\ge\sqrt{mk}/28$. Replacing one sample changes $Z$ by at most $2$. The bounded-differences inequality \citep[Theorem~1]{warnke2016typical} therefore gives $\PP(Z<c\sqrt{mk})\le e^{-ck}$ for a universal constant $c>0$~\citep{mcdiarmid1989method}. Since $\eta(\widehat p)\ge Z/m$, this proves the lower bound.

For the upper bound, let $u$ be the uniform distribution on $[k]$. Every balanced sign vector has mean zero under $u$, so $\eta(\widehat p)\le\|\widehat p-u\|_1$. Moreover,
\[
\EE[\eta(\widehat p)]\le\sum_{r=1}^k\EE[|\widehat p_r-1/k|]\le\sum_{r=1}^k\sqrt{\operatorname{Var}(\widehat p_r)}\le\sqrt{k/m}.
\]
Replacing one sample changes $\eta(\widehat p)$ by at most $2/m$. Another application of the same bounded-differences inequality gives $\PP(\eta(\widehat p)>2\sqrt{k/m})\le e^{-k/2}$. Combining the two probability bounds proves the lemma.
\end{proof}

Take $\mathbf D=\prod_{i=1}^nD_i$, where each $D_i$ is uniform on $[0,1]$. Let $(X_{i,j})_{i,j}$ be independent samples with $X_{i,j}\sim D_i$, and define the empirical distributions $\widehat D_{i,m}$ and their shaded versions $E_{i,m}$ as in \eqref{eq:app-wbv-optimistic-marginal}. For $m\ge1$, write $\widehat p_{i,m}:=(\widehat D_{i,m}(I_r))_{r=1}^k$ and $e_{i,m}:=(E_{i,m}(I_r))_{r=1}^k$.

For probability vectors $p,q$, the definition of $\eta$ gives $|\eta(p)-\eta(q)|\le\|p-q\|_1$. Passing to cell probabilities cannot increase total variation distance, so \Cref{lem:app-wbv-shift-tv} gives
\[
|\eta(e_{i,m})-\eta(\widehat p_{i,m})|
\le2d_{\mathrm{TV}}(E_{i,m},\widehat D_{i,m})
\le2\left(\sqrt{\frac{2\ell_T}{m}}+\frac{4\ell_T}{m}\right).
\]
For $m\ge k$, the last expression is at most $(2\sqrt{2\ell_T/k}+8\ell_T/k)\sqrt{k/m}$. Consequently, \Cref{lem:app-wbv-balanced-empirical} implies that there are universal constants $c,C,C_1>0$ such that, whenever $k\ge C_1\ell_T$, for every item $i$ and sample size $m\ge k$,
\begin{equation}\label{eq:app-wbv-shifted-balanced-bound}
\PP\left(c\sqrt{k/m}\le\eta(e_{i,m})\le C\sqrt{k/m}\right)\ge1-Ce^{-ck}.
\end{equation}

We now choose the horizon and verify these estimates over a block of rounds. Fix a sufficiently large universal constant $C_\star$ and set
\begin{equation}\label{eq:app-wbv-lower-horizon}
m_\star:=\left\lfloor C_\star\frac{n^2}{k}\right\rfloor,\qquad T_n:=2m_\star+1,\qquad \ell_n:=\log\bigl(2nT_n^2(T_n+1)\bigr).
\end{equation}
Run \Cref{alg:strong-up} with $\delta=1/(T_n+1)$, so $\ell_T=\ell_n$. Since $k\asymp\sqrt n$ and $\ell_n=O(\log n)$, we have $m_\star\ge k$ for all sufficiently large $n$ and $k/\ell_n\to\infty$. Every policy in $\cC_n$ probes all $n$ items, so $N_i^t=t-1$ deterministically and the oracle input in round $t$ is $\widetilde{\mathbf E}^{\,t}=\prod_{i=1}^nE_{i,t-1}$.

Let $\mathcal G_n$ be the event that, simultaneously for every integer $m\in[m_\star,2m_\star]$ and every $i\in[n]$,
\begin{equation}\label{eq:app-wbv-good-q-bounds}
c\sqrt{k/m}\le q_{i,m}:=\eta(e_{i,m})\le C\sqrt{k/m},
\end{equation}
and, for every such $m$,
\begin{equation}\label{eq:app-wbv-empirical-center}
\left|\sum_{i=1}^n\EE_{X\sim E_{i,m}}[b(X)]-m_0\right|\le W/8.
\end{equation}
We show that $\PP(\mathcal G_n)=1-o(1)$. For \eqref{eq:app-wbv-good-q-bounds}, a union bound applied to \eqref{eq:app-wbv-shifted-balanced-bound} gives failure probability at most $Cn(m_\star+1)e^{-ck}=o(1)$.

To prove \eqref{eq:app-wbv-empirical-center}, put $\overline U_{i,m}:=m^{-1}\sum_{j=1}^m b(X_{i,j})$. Since $b$ takes values in $[1,k]$, \Cref{lem:app-wbv-shift-tv} gives
\[
\left|\sum_{i=1}^n\EE_{X\sim E_{i,m}}[b(X)]-m_0\right|
\le\left|\sum_{i=1}^n\overline U_{i,m}-m_0\right|
+nk\left(\sqrt{\frac{2\ell_n}{m}}+\frac{4\ell_n}{m}\right).
\]
Each $b(X_{i,j})$ is uniform on $[k]$, so $\EE[\sum_i\overline U_{i,m}]=m_0$. Applying Hoeffding's inequality to the $nm$ independent observations yields
\[
\PP\left(\left|\sum_i\overline U_{i,m}-m_0\right|>W/16\right)
\le2\exp\left(-\frac{W^2m}{128nk^2}\right)
=2e^{-A_0^2m/128}.
\]
The deterministic shading term decreases with $m$, and at $m=m_\star$ it satisfies
\[
\frac{nk}{W}\left(\sqrt{\frac{2\ell_n}{m_\star}}+\frac{4\ell_n}{m_\star}\right)
\le C\left(\sqrt{\frac{k\ell_n}{n}}+\frac{k\ell_n}{n^{3/2}}\right)=o(1).
\]
Thus this term is at most $W/16$ throughout the block for all sufficiently large $n$. A union bound over the $m_\star+1$ sample sizes proves \eqref{eq:app-wbv-empirical-center} with probability $1-o(1)$. Together with the discrepancy bounds, this proves $\PP(\mathcal G_n)=1-o(1)$.

On $\mathcal G_n$, consider a round with $m=t-1\in[m_\star,2m_\star]$. By \eqref{eq:app-wbv-empirical-center} and \Cref{lem:app-wbv-measurable-selector}, the oracle returns identity maps, balanced signs $\mathbf s^t$ satisfying $\EE_{X\sim E_{i,m}}[s_i^t(b(X))]=q_{i,m}$, and phase
\[
\theta_t=-\Theta_m\pmod{2\pi},\qquad \Theta_m:=\sum_{i=1}^n\arctan(q_{i,m}\tan a).
\]
Since $0\le q_{i,m}\le1$ and $k\ge2$, we have $q_{i,m}\tan a<1$. Using $\arctan z\asymp z$ for $z\in[0,1]$, $\tan a\asymp1/k$, and \eqref{eq:app-wbv-good-q-bounds}, we obtain $cn/\sqrt{km}\le\Theta_m\le Cn/\sqrt{km}$. The choice of $m_\star$ therefore gives
\begin{equation}\label{eq:app-wbv-phase-window}
\frac{c}{\sqrt{C_\star}}\le\Theta_m\le\frac{C}{\sqrt{C_\star}}\le1,
\end{equation}
where the last inequality holds for a sufficiently large fixed choice of $C_\star$.

To evaluate the selected policy under $\mathbf D$, hold its parameters fixed. Each balanced sign $s_i^t(b(X_i))$ has mean zero under $D_i$, and the coordinates are independent. Consequently,
\[
\EE_{\xvec\sim\mathbf D}\left[\cos\left(\theta_t+a\sum_{i=1}^n s_i^t(b(X_i))\right)\right]=(\cos a)^n\cos\Theta_m.
\]
The policy with the same signs and identity maps but phase $0$ belongs to $\cC_n$. Its increasing term is unchanged, while its expected cosine term is $(\cos a)^n$. Thus
\[
\OPT(\mathbf D)-R(\sigma^t\mid\mathbf D)\ge\varepsilon(\cos a)^n(1-\cos\Theta_m).
\]
Since $n/k^2\to1$, the factor $(\cos a)^n$ is bounded below by a positive universal constant for all sufficiently large $n$. Together with \eqref{eq:app-wbv-phase-window} and $1-\cos z\ge z^2/4$ for $z\in[0,1]$, this gives $\OPT(\mathbf D)-R(\sigma^t\mid\mathbf D)\ge c\varepsilon/C_\star$ in every round of the block on $\mathcal G_n$.

There are $m_\star+1$ rounds in this block, and ordinary pseudo-regret is nonnegative in every round. Hence, for every sample path,
$R(T_n)\ge\frac{c\varepsilon}{C_\star}m_\star\mathbf1_{\mathcal G_n}.$
Taking expectations and using $\PP(\mathcal G_n)=1-o(1)$ yields
\[
\EE[R(T_n)]\ge\frac{c\varepsilon}{C_\star}m_\star\PP(\mathcal G_n)\ge c_{V_0}n^{3/2}.
\]
Here $c_{V_0}>0$ depends only on $V_0$, since all construction constants are fixed. This proves \Cref{thm:app-wbv-lower-bound}. By \eqref{eq:app-wbv-lower-horizon}, $T_n=\Theta(n^{3/2})$, so the bound is also $\Omega_{V_0}(n^{3/4}\sqrt{T_n})$. This establishes the lower-bound claim in \Cref{thm:weak-bv-exact-exponent}.

\subsection{Comparison with the Analysis of \citet{agarwal2026semibandit}}
\label{sec:app-agarwal-comparison}

We justify the comparison with \citet{agarwal2026semibandit} in \Cref{subsec:weak-bv-exact-exponent}. We first relate their regularity assumptions to separate bounded variation, then establish the corresponding stability and regret bounds. We use the paper's notation, with distributions supported on $[0,1]$. For distributions $P,Q$, write $d_{\rm KS}(P,Q):=\sup_x|F_P(x)-F_Q(x)|$.

A $k$-threshold policy is represented by a decision tree in which each probe partitions the item's possible values into at most $k$ intervals, with one branch for each interval. Each leaf $\ell$ has a reward function $r_{\sigma,\ell}$ that agrees with $r_\sigma$ on realizations reaching that leaf. Assumption~1 of \citet{agarwal2026semibandit} requires these leaf functions to be coordinate-wise continuous and to have uniformly bounded coordinate variation. The following implication uses their variation bound and the range of the realized reward.

\begin{lemma}[Separate variation of threshold policies]\label{lem:agarwal-threshold-bv}
Let $\sigma$ be a $k$-threshold policy whose realized reward takes values in an interval of length $G$. Suppose every leaf reward function has separate total variation at most $V_{\rm leaf}$. Then
\begin{equation}\label{eq:agarwal-implies-tv-sep}
\operatorname{TV}_{\rm sep}(r_\sigma)\le kV_{\rm leaf}+(k-1)G.
\end{equation}
In particular, for rewards in $[0,1]$, we have $\operatorname{TV}_{\rm sep}(r_\sigma)\le kV_{\rm leaf}+(k-1)$.
\end{lemma}

\begin{proof}
Fix a coordinate $i$ and the remaining values $x_{-i}$. Before item $i$ is probed, the execution is independent of $x_i$. If the policy stops without probing $i$, its reward is constant as $x_i$ varies. Otherwise, it reaches the same node labeled by $i$ for every value of $x_i$. The thresholds at this node partition $[0,1]$ into at most $k$ intervals. On each interval, the execution reaches a fixed leaf: item $i$ is not probed again, and all remaining observations are fixed by $x_{-i}$. Thus the reward section agrees with a section of one leaf function on each interval.

Consider any finite partition of $[0,1]$ used to calculate the pointwise variation of this reward section. The increments whose endpoints lie in the same threshold interval contribute at most $V_{\rm leaf}$ per interval. At most $k-1$ increments cross between intervals, and each has magnitude at most $G$. The total is therefore at most $kV_{\rm leaf}+(k-1)G$. Taking the supremum over partitions, coordinates, and fixed values $x_{-i}$ proves \eqref{eq:agarwal-implies-tv-sep}.
\end{proof}

The converse does not hold. Consider a one-item policy that probes the item and receives reward
\[
g(x):=\sum_{j=1}^{\infty}2^{-j}\one\{x\ge1/j\},\qquad x\in[0,1].
\]
This function is nondecreasing, with $g(0)=0$ and $g(1)=1$, so its pointwise total variation is $1$. However, it has a jump of size $2^{-j}$ at every $1/j$ for $j\ge2$. Any finite partition into intervals leaves at least one of these jumps in the interior of a piece. Hence $g$ cannot agree with a continuous leaf function on every piece of any finite threshold partition. Separate bounded variation therefore permits rewards excluded by the threshold and continuous-leaf assumptions.

We next prove a policy-stability bound directly from separate bounded variation.

\begin{lemma}[Bounded variation and KS distance]\label{lem:agarwal-bv-ks}
Let $P,Q$ be distributions on $[0,1]$, and let $g:[0,1]\to\mathbb R$ have finite pointwise total variation. Then
\begin{equation}\label{eq:agarwal-bv-ks}
\left|\int g\,dP-\int g\,dQ\right|\le\operatorname{TV}_{[0,1]}(g)\,d_{\rm KS}(P,Q).
\end{equation}
\end{lemma}

\begin{proof}
Put $v(x):=\operatorname{TV}_{[0,x]}(g)$, with $v(0)=0$, and define $g_\pm(x):=(v(x)\pm(g(x)-g(0)))/2$. Since $|g(y)-g(x)|\le v(y)-v(x)$ for $x\le y$, both $g_+$ and $g_-$ are nonnegative and nondecreasing. Moreover, $g=g(0)+g_+-g_-$ and $g_+(1)+g_-(1)=\operatorname{TV}_{[0,1]}(g)$.

For any nonnegative nondecreasing function $h$, each superlevel set $\{h>u\}$ is an upper interval, possibly empty. Its probabilities under $P$ and $Q$ differ by at most $d_{\rm KS}(P,Q)$, using the CDFs or their left limits according to whether the lower endpoint is excluded or included. Integrating over $u$ gives
\[
\left|\int h\,d(P-Q)\right|
\le\int_0^{h(1)}|P(h>u)-Q(h>u)|\,du
\le h(1)d_{\rm KS}(P,Q).
\]
Apply this bound to $g_+$ and $g_-$. The constant $g(0)$ cancels because $P$ and $Q$ have equal total mass, proving \eqref{eq:agarwal-bv-ks}.
\end{proof}

\begin{proposition}[Policy stability in KS distance]\label{prop:agarwal-tv-sep-stability}
Let $\mathbf D=\prod_{i=1}^nD_i$ and $\mathbf E=\prod_{i=1}^nE_i$ be product distributions on $[0,1]^n$. For every feasible policy $\sigma$ with $\operatorname{TV}_{\rm sep}(r_\sigma)\le V_0$,
\begin{equation}\label{eq:agarwal-tv-sep-stability}
|R(\sigma\mid\mathbf E)-R(\sigma\mid\mathbf D)|
\le V_0\sum_{i=1}^n\pi_i^{\mathbf D}(\sigma)\,d_{\rm KS}(D_i,E_i).
\end{equation}
\end{proposition}

\begin{proof}
As in the proof of \Cref{prop:app-wbv-one-round}, let $W(h)$ be the expected reward obtained by continuing $\sigma$ from a consistent history $h$, sampling the unobserved coordinates independently from their marginals in $\mathbf E$. This defines $W(h)$ even when $h$ has zero probability under $\mathbf E$. If the next item probed at $h$ is $i$, put $g_h(y):=W(h\oplus(i,y))$.

Nonanticipation ensures that changing the unobserved value of item $i$ leaves the preceding history unchanged. Thus $g_h$ is an average of coordinate sections of $r_\sigma$, with the observations in $h$ fixed. Averaging cannot increase total variation, so $\operatorname{TV}_{[0,1]}(g_h)\le V_0$. By \eqref{eq:agarwal-bv-ks},
$\left|\int g_h\,d(E_i-D_i)\right|\le V_0d_{\rm KS}(D_i,E_i)$.

Run $\sigma$ under $\mathbf D$. Let $H_\ell$ be the history after $\ell$ probes, let $\tau\le n$ be the total number of probes, and let $i_\ell$ be the next item probed when $\ell<\tau$. The continuation-value identity \eqref{eq:app-wbv-simulation-telescoping} gives
\[
\begin{aligned}
|R(\sigma\mid\mathbf E)-R(\sigma\mid\mathbf D)|
&\le\EE_{\mathbf D}\sum_{\ell=0}^{\tau-1}\left|\int g_{H_\ell}\,d(E_{i_\ell}-D_{i_\ell})\right|\\
&\le V_0\EE_{\mathbf D}\sum_{\ell=0}^{\tau-1}d_{\rm KS}(D_{i_\ell},E_{i_\ell}) = V_0\sum_{i=1}^n\pi_i^{\mathbf D}(\sigma)\,d_{\rm KS}(D_i,E_i).
\end{aligned}
\]
The last equality uses that each item is probed at most once.
\end{proof}

We now derive the regret bound obtained from this KS stability estimate under the same assumptions as our upper bound.

\begin{corollary}\label{cor:agarwal-bv-regret}
Under the upper-bound assumptions of \Cref{thm:weak-bv-exact-exponent}, for every $T\ge\max\{n,2\}$, \Cref{alg:strong-up}, using $\mathcal O$ and $\delta=1/(T+1)$, satisfies
\begin{equation}\label{eq:agarwal-bv-regret}
\EE[\alpha\text{-}R(T)]\le CV_0n\sqrt{T\log(2nT)}+1,
\end{equation}
where $C$ is universal.
\end{corollary}

\begin{proof}
We again give the proof for up-monotone maximization; the extensions described in \Cref{sec:weak-monotone} apply to the KS stability bound as well.
Use the notation of Appendix~\ref{sec:app-wbv-upper-bound}, including $\ell_T$, $E_{i,m}$, and $\rho_m$ from \eqref{eq:app-wbv-optimistic-marginal} and \eqref{eq:app-wbv-def-rho}. Applying \Cref{lem:app-wbv-shifted-interval} to intervals $[0,x]$ and using $\rho_m\le1$ gives
$d_{\rm KS}(D_i,E_{i,m})\le C\sqrt{\rho_m}.$
Together with $D_i\sd E_{i,m}$, this holds simultaneously for all items and sample sizes with probability at least $1-1/T$. The case $m=0$ is included because $E_{i,0}=\delta_1$ and $\rho_0=1$.

Let $\mathcal G_t$ be the event that these two inequalities hold for the marginals used in round $t$. This event is determined by $\mathcal H_{t-1}$ and satisfies $\PP(\mathcal G_t^c)\le1/T$. Write $\bar\pi_i^t:=\EE_t[\pi_i^{\mathbf D}(\sigma^t)]$ and $\Delta_t:=\alpha\OPT(\mathbf D)-\EE_t[R(\sigma^t\mid\mathbf D)]$. On $\mathcal G_t$, up-monotonicity and the oracle's approximation guarantee give
\[
\Delta_t \le \EE_t\!\left[R(\sigma^t\mid\widetilde{\mathbf E}^{\,t})-R(\sigma^t\mid\mathbf D)\right] \le CV_0\sum_{i=1}^n\bar\pi_i^t\sqrt{\rho_{N_i^t}}.
\]
The second inequality applies \Cref{prop:agarwal-tv-sep-stability} to each returned policy and averages over the oracle randomness.

Since $\Delta_t\le1$ in every round, the failure events contribute at most $1$ to expected regret. Applying the counting identity \eqref{eq:app-wbv-adaptive-counting} with $a(m)=\sqrt{\rho_m}$ therefore gives
\[
\begin{aligned}
\EE[\alpha\text{-}R(T)]
&\le CV_0\sum_{t=1}^T\sum_{i=1}^n\EE\!\left[\bar\pi_i^t\sqrt{\rho_{N_i^t}}\right]+1\\
&\le CV_0n\sum_{m=0}^{T-1}\sqrt{\rho_m}+1 \le CV_0n\sqrt{T\ell_T}+1.
\end{aligned}
\]
The last inequality uses $\rho_0=1$, $\sqrt{\rho_m}\le\sqrt{C_0\ell_T/m}$ for $m\ge1$, and $\sum_{m=1}^{T-1}m^{-1/2}\le2\sqrt T$. Since $\ell_T\le C\log(2nT)$, this proves \eqref{eq:agarwal-bv-regret}.
\end{proof}

For rewards in $[0,1]$, \Cref{lem:agarwal-threshold-bv} allows us to take $V_0=kV_{\rm leaf}+(k-1)$ under the threshold and leaf-variation assumptions of \citet{agarwal2026semibandit}. Substituting this bound into \eqref{eq:agarwal-bv-regret} gives
\[
O\left(k(1+V_{\rm leaf})n\sqrt{T\log(2nT)}+1\right),
\]
recovering their dependence on the threshold parameter, leaf variation, and number of items in our normalized reward setting.

More generally, \eqref{eq:agarwal-bv-regret} and \eqref{eq:weak-bv-upper} apply under the same output-policy variation assumption. For fixed $V_0>0$, our bound improves the polynomial dependence on $n$ from $n$ to $n^{3/4}$, up to logarithmic factors. Under the threshold assumptions, our bound becomes
\[
O\left(k(1+V_{\rm leaf})n^{3/4}\sqrt T\,\log^2(2nT)+1\right).
\]
Thus the dependence on $k$ remains through the variation bound, while separate bounded variation also covers policies with no finite threshold representation.

%% file: computations-appendix-syn.tex
\section{Additional Computational Results}
\label{app:computations}

\paragraph{Implementation details.}
Time-curve bands use pointwise $95\%$  intervals across independent replications. 
Dimension intervals use the $2.5\%$ and $97.5\%$ quantiles of $10{,}000$ whole-replication bootstrap resamples. Continuous-instance intervals are conditional on the fixed instance. Each candidate uses fresh runs and confidence parameters computed with that $T$. Uniform-endpoint instances use prefixes of one saved vector drawn independently and uniformly from $\{0.2,0.4,0.6,0.8,1\}$, preserving its order across replications. 

\subsection{Bernoulli Best-of-$K$ Selection}
\label{app:best-k}

\begin{figure*}[htbp]
    \centering

    \begin{subfigure}[t]{.485\textwidth}
        \centering
        \includegraphics[width=\linewidth]
        {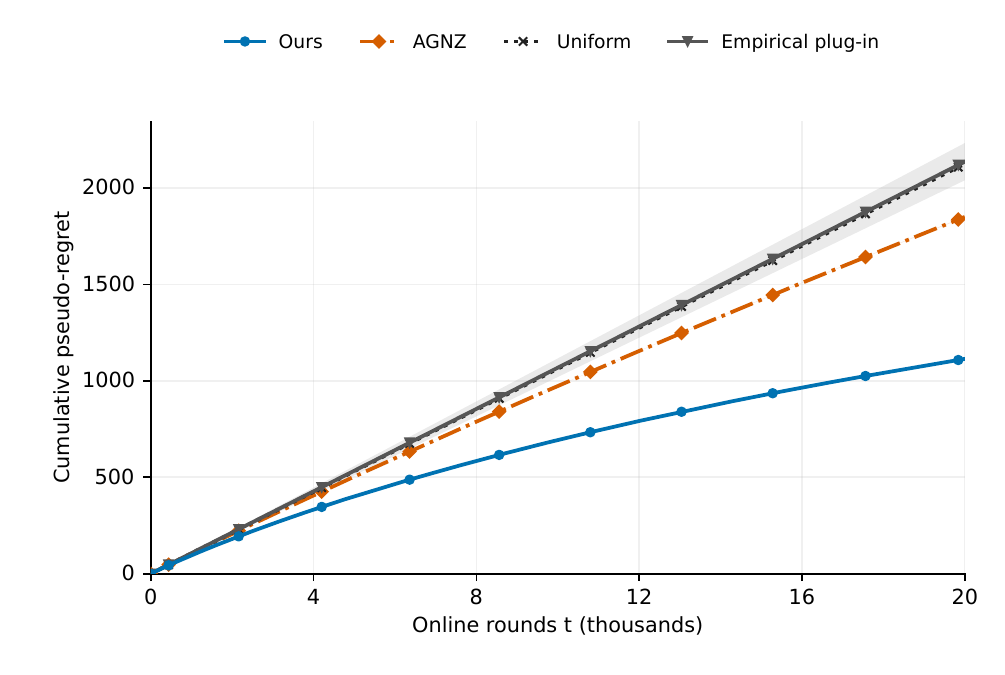}
        \caption{$n=50$.}
    \end{subfigure}
    \hfill
    \begin{subfigure}[t]{.485\textwidth}
        \centering
        \includegraphics[width=\linewidth]
        {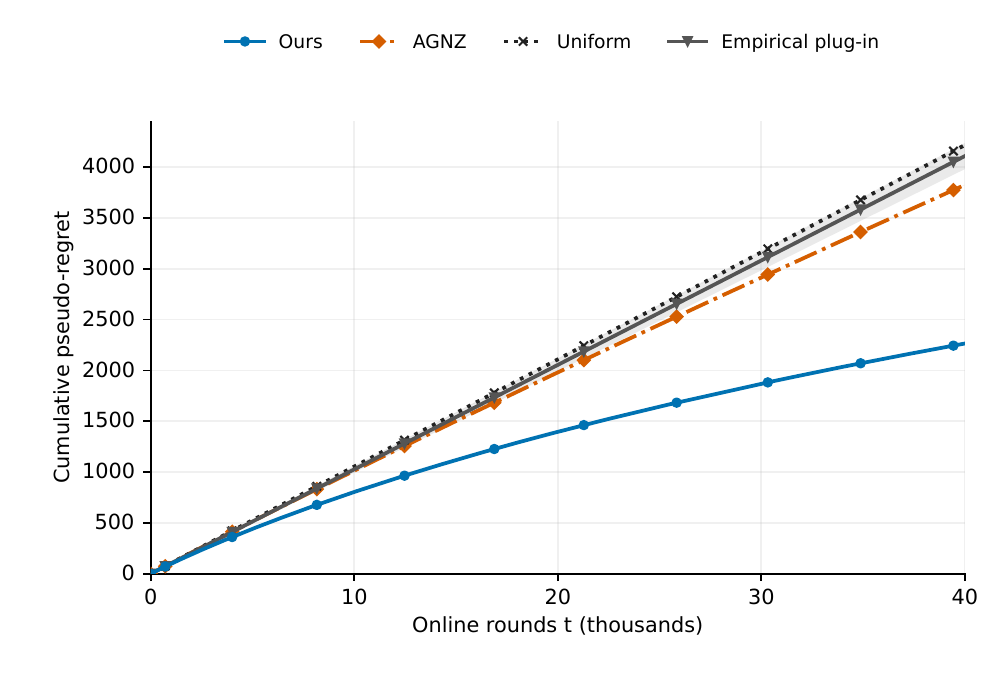}
        \caption{$n=100$.}
    \end{subfigure}

    \medskip

    \begin{subfigure}[t]{.485\textwidth}
        \centering
        \includegraphics[width=\linewidth]
        {plots/bernoulli_best_of_k_n500_K_n_over_2.pdf}
        \caption{$n=500$.}
    \end{subfigure}
    \hfill
    \begin{subfigure}[t]{.485\textwidth}
        \centering
        \includegraphics[width=\linewidth]
        {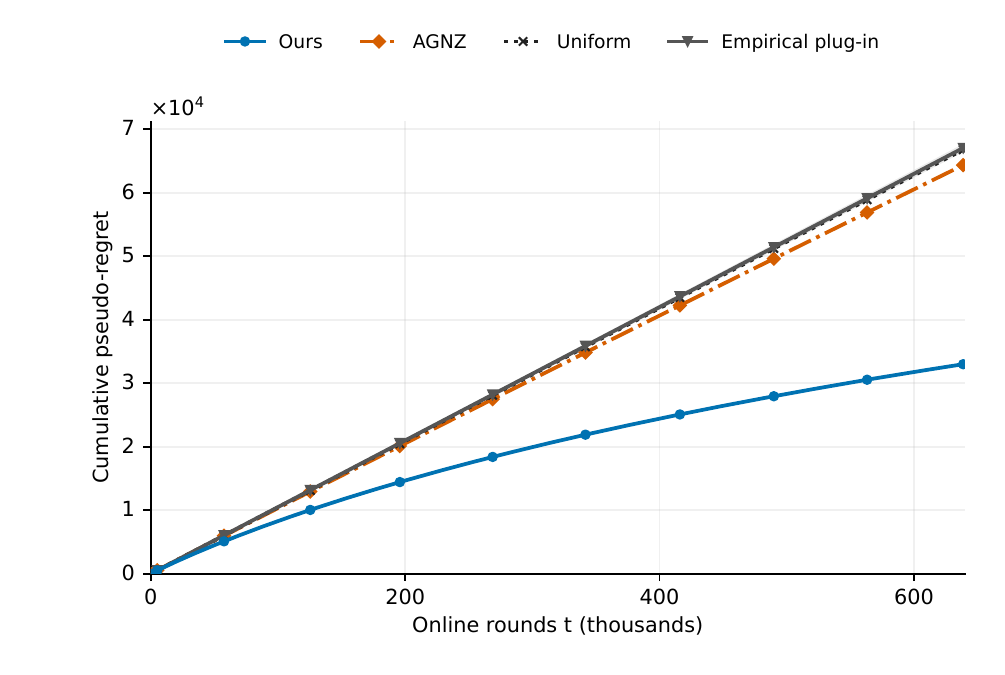}
        \caption{$n=1000$.}
    \end{subfigure}

    \caption{Computational results for Bernoulli Best-of-$K$ selection with $K=n/2$.}
    \label{fig:app-best-k-half}
\end{figure*}

\begin{figure*}[htbp]
\centering
\begin{subfigure}[t]{.485\textwidth}
\centering
\includegraphics[width=\linewidth]{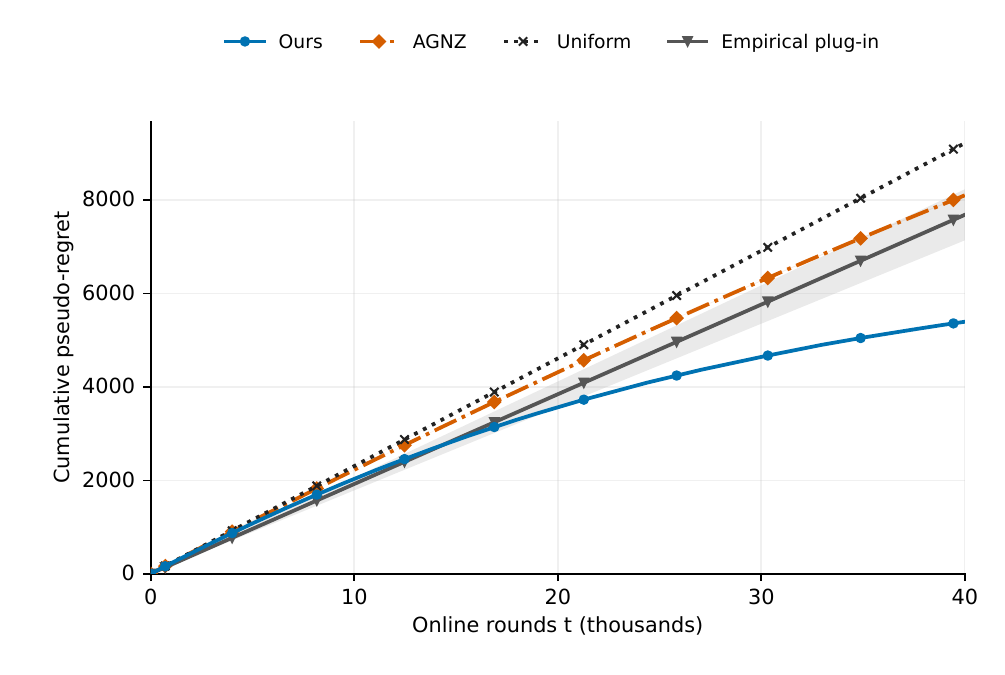}
\caption{$n=50$.}
\end{subfigure}\hfill
\begin{subfigure}[t]{.485\textwidth}
\centering
\includegraphics[width=\linewidth]{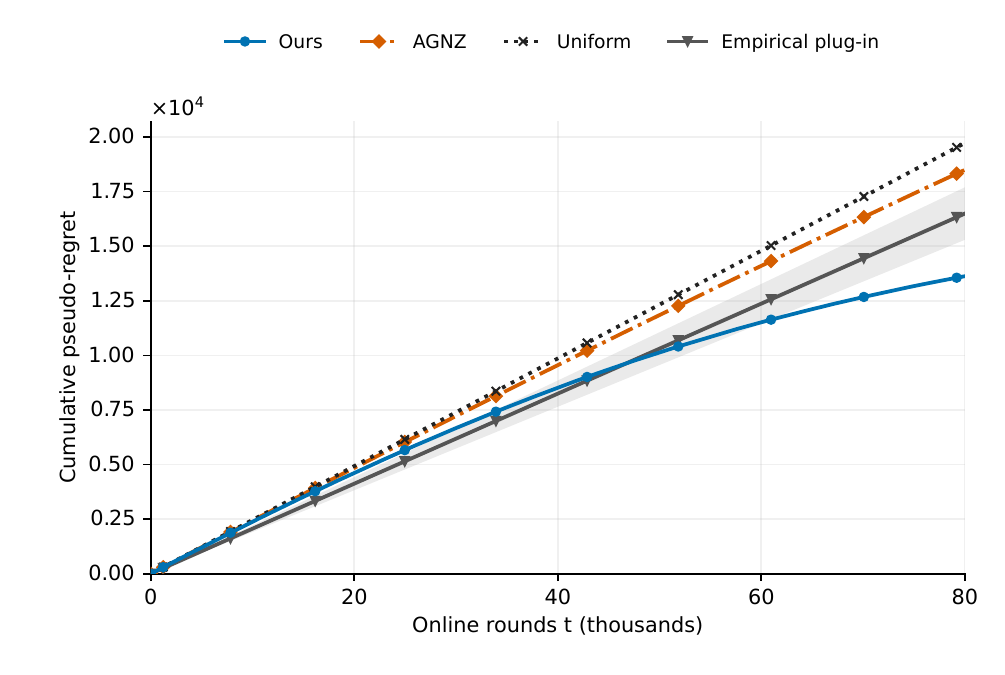}
\caption{$n=100$.}
\end{subfigure}

\medskip
\begin{subfigure}[t]{.485\textwidth}
\centering
\includegraphics[width=\linewidth]{plots/bernoulli_best_of_k_n500_K_5.pdf}
\caption{$n=500$.}
\end{subfigure}\hfill
\begin{subfigure}[t]{.485\textwidth}
\centering
\includegraphics[width=\linewidth]{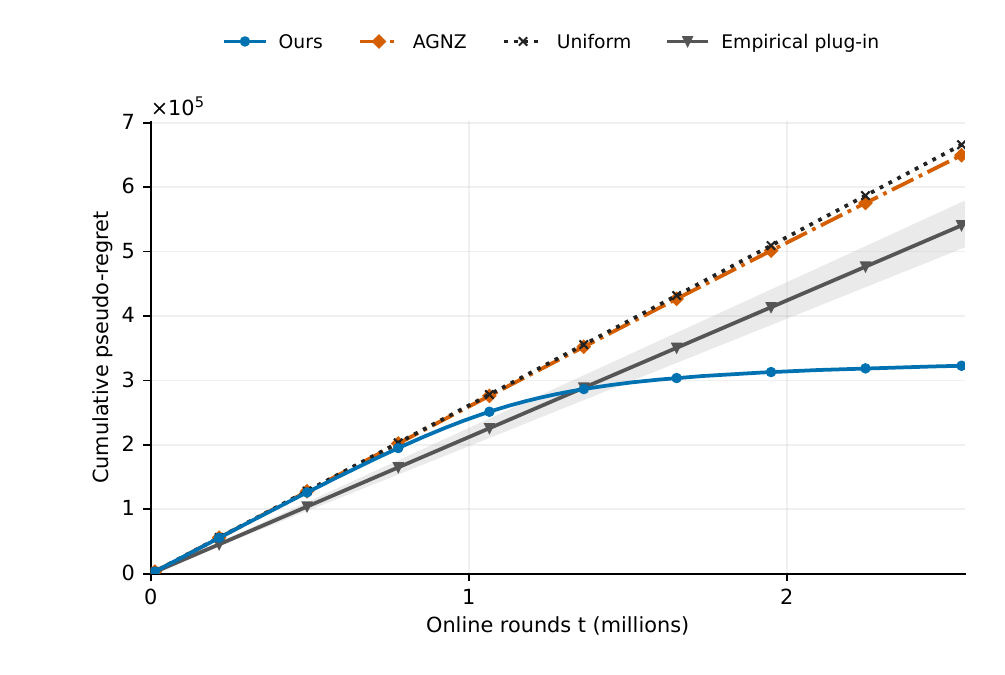}
\caption{$n=1000$.}
\end{subfigure}
\caption{Computational results for Bernoulli Best-of-$K$ selection with $K=5$.}
\label{fig:app-best-k-five}
\end{figure*}

\Cref{fig:app-best-k-half,fig:app-best-k-five} report the remaining regret-versus-time plots for the instances described in \Cref{sec:computations}.

\clearpage
\subsection{Multiple-Choice Prophet Inequality}
\label{app:prophet}

We report the remaining prophet experiments described in \Cref{sec:computations}. 

\begin{figure*}[htbp]
\centering
\begin{subfigure}[t]{.485\textwidth}
\centering
\includegraphics[width=\linewidth]{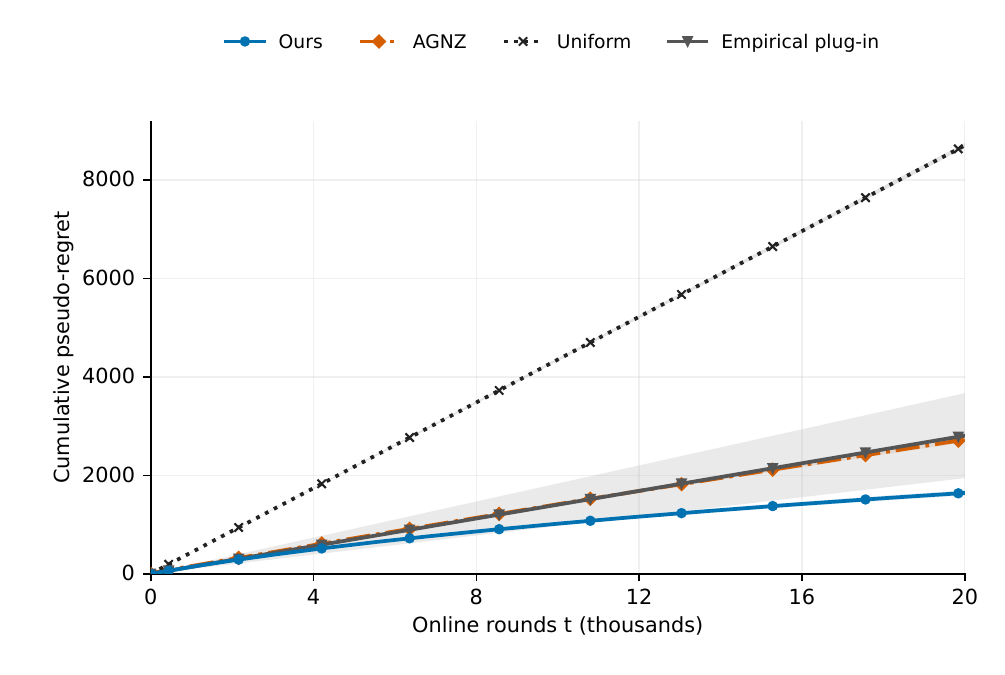}
\caption{$n=50$.}
\end{subfigure}\hfill
\begin{subfigure}[t]{.485\textwidth}
\centering
\includegraphics[width=\linewidth]{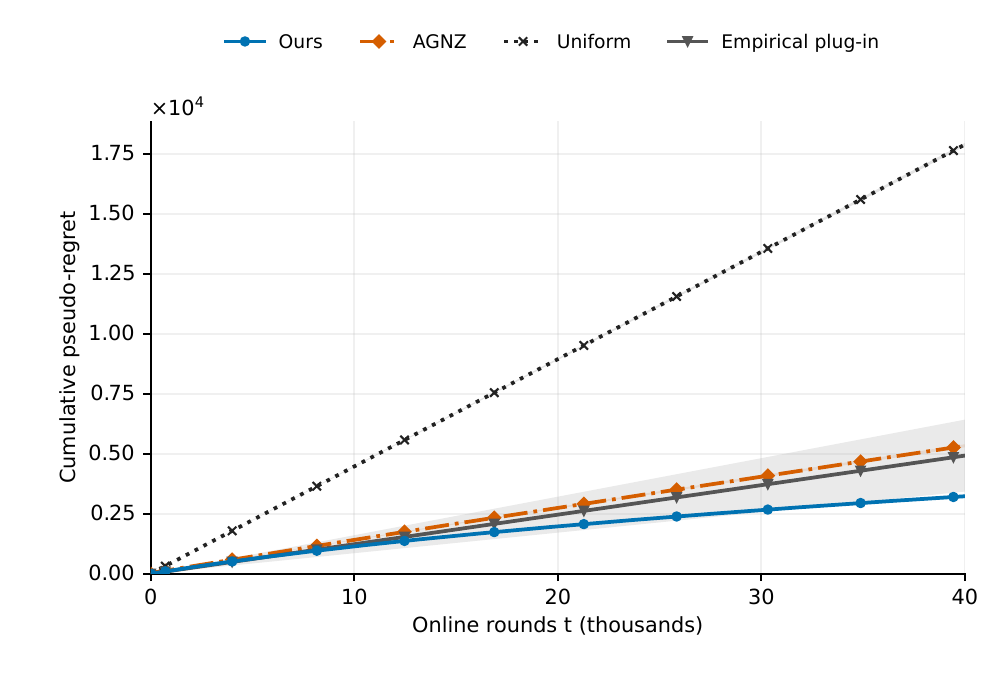}
\caption{$n=100$.}
\end{subfigure}

\medskip
\begin{subfigure}[t]{.485\textwidth}
\centering
\includegraphics[width=\linewidth]{plots/prophet_n500.pdf}
\caption{$n=500$.}
\end{subfigure}\hfill
\begin{subfigure}[t]{.485\textwidth}
\centering
\includegraphics[width=\linewidth]{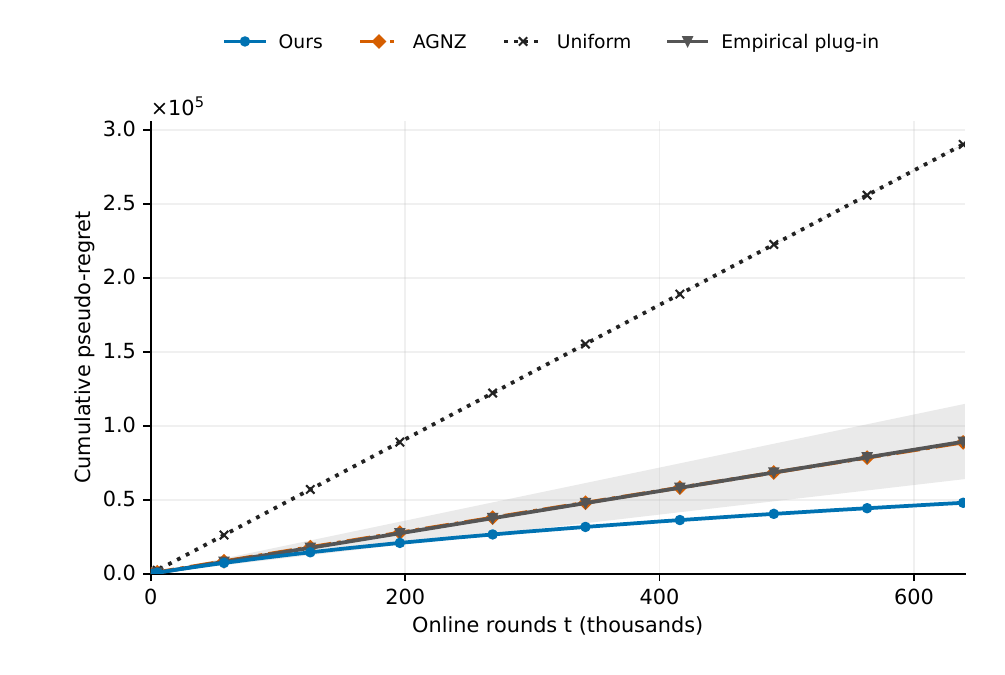}
\caption{$n=1000$.}
\end{subfigure}
\caption{Computational results for discrete prophet instances with $k=1$.}
\label{fig:app-prophet-discrete-one}
\end{figure*}

\begin{figure*}[htbp]
\centering
\begin{subfigure}[t]{.485\textwidth}\centering\includegraphics[width=\linewidth]{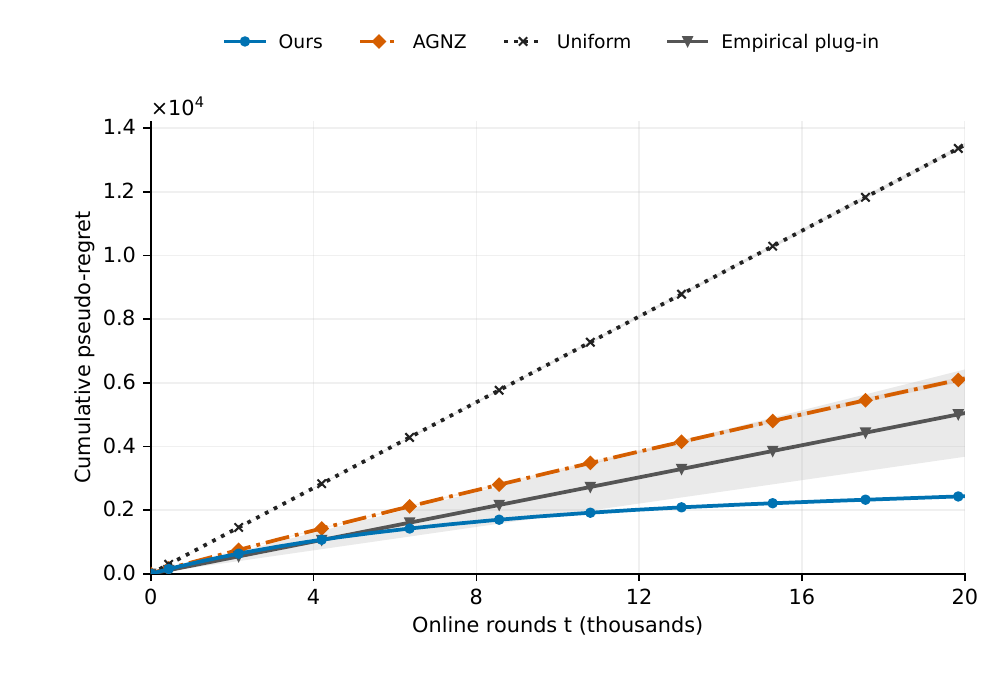}\caption{$n=50$.}\end{subfigure}\hfill
\begin{subfigure}[t]{.485\textwidth}\centering\includegraphics[width=\linewidth]{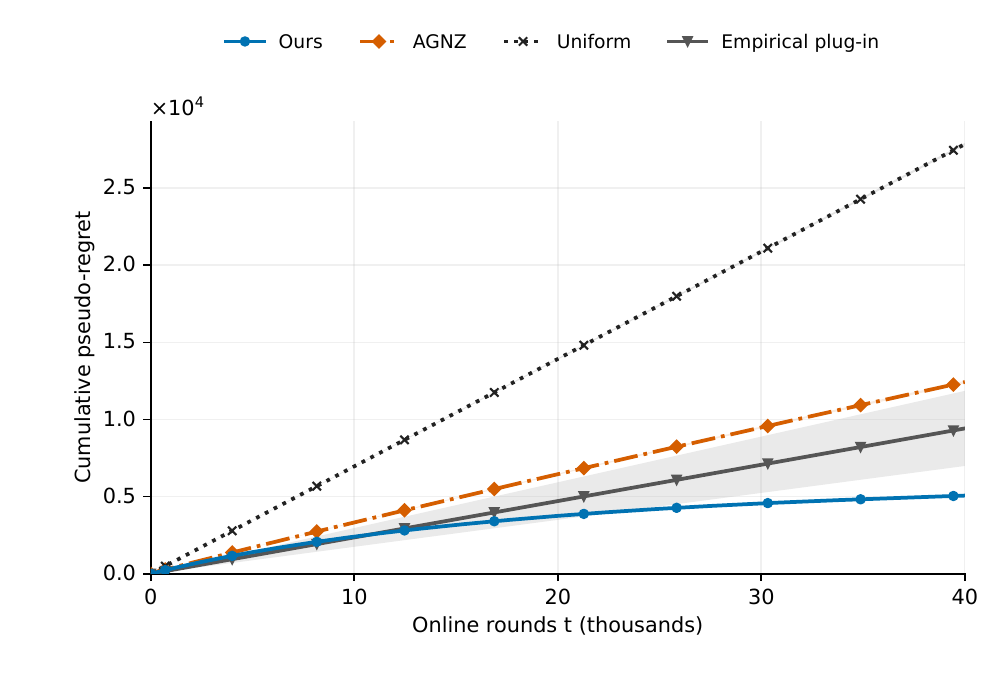}\caption{$n=100$.}\end{subfigure}

\medskip
\begin{subfigure}[t]{.485\textwidth}\centering\includegraphics[width=\linewidth]{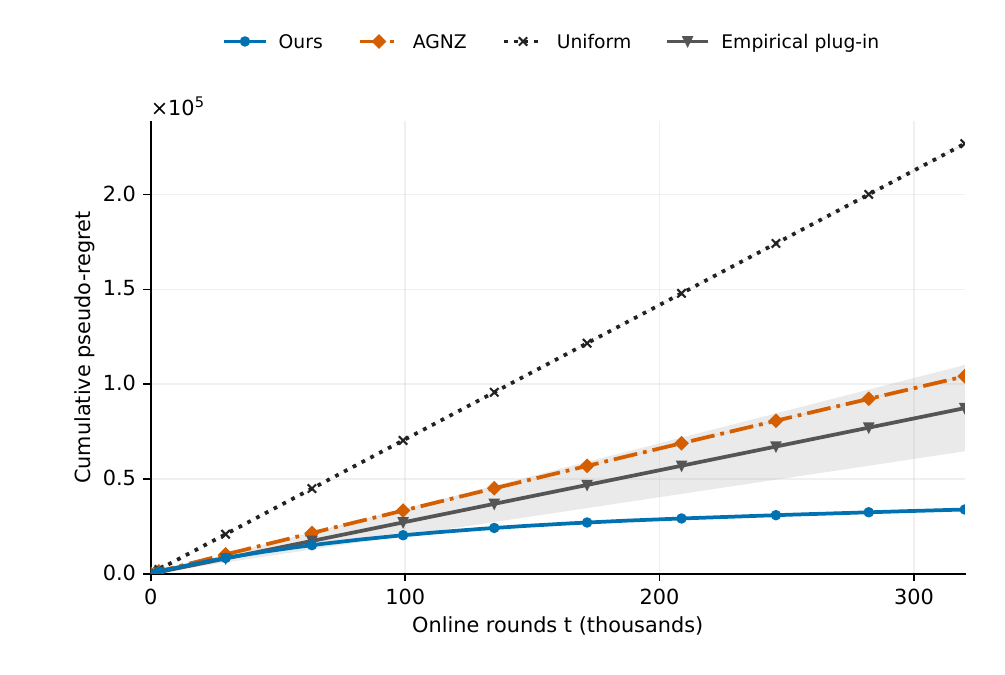}\caption{$n=500$.}\end{subfigure}\hfill
\begin{subfigure}[t]{.485\textwidth}\centering\includegraphics[width=\linewidth]{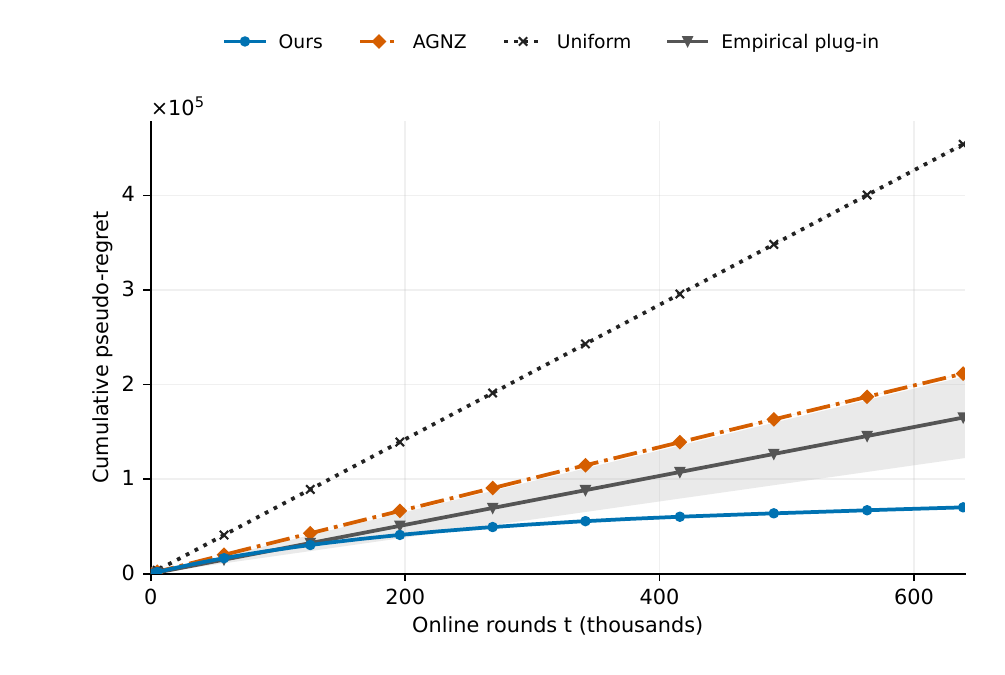}\caption{$n=1000$.}\end{subfigure}
\caption{Computational results for discrete prophet instances with $k=2$.}
\label{fig:app-prophet-discrete-two}
\end{figure*}

\begin{figure*}[htbp]
\centering
\begin{subfigure}[t]{.485\textwidth}\centering\includegraphics[width=\linewidth]{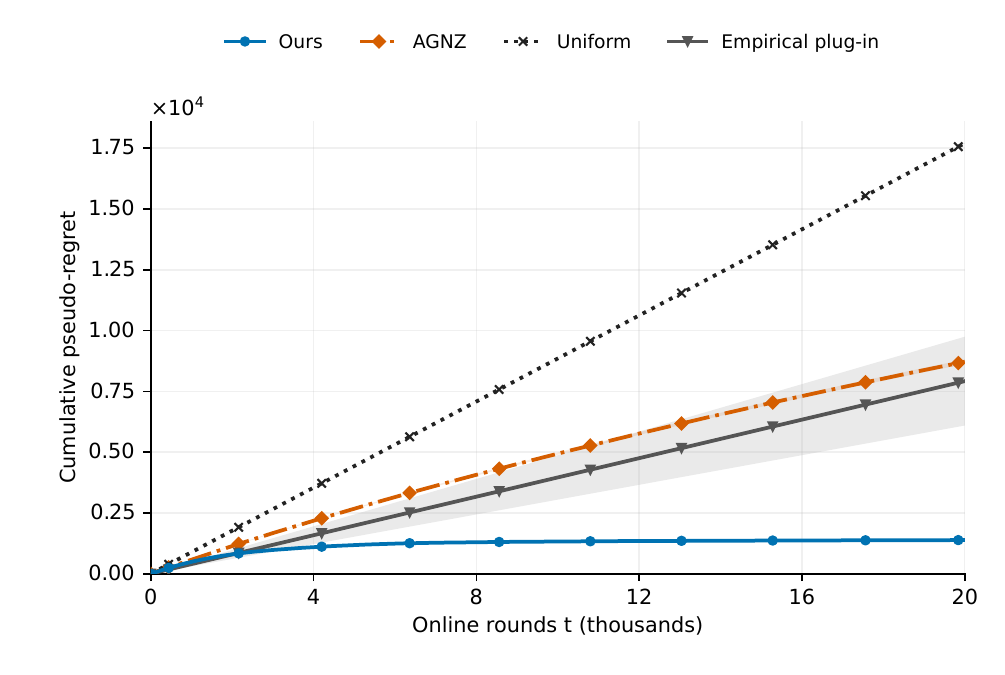}\caption{$n=50$.}\end{subfigure}\hfill
\begin{subfigure}[t]{.485\textwidth}\centering\includegraphics[width=\linewidth]{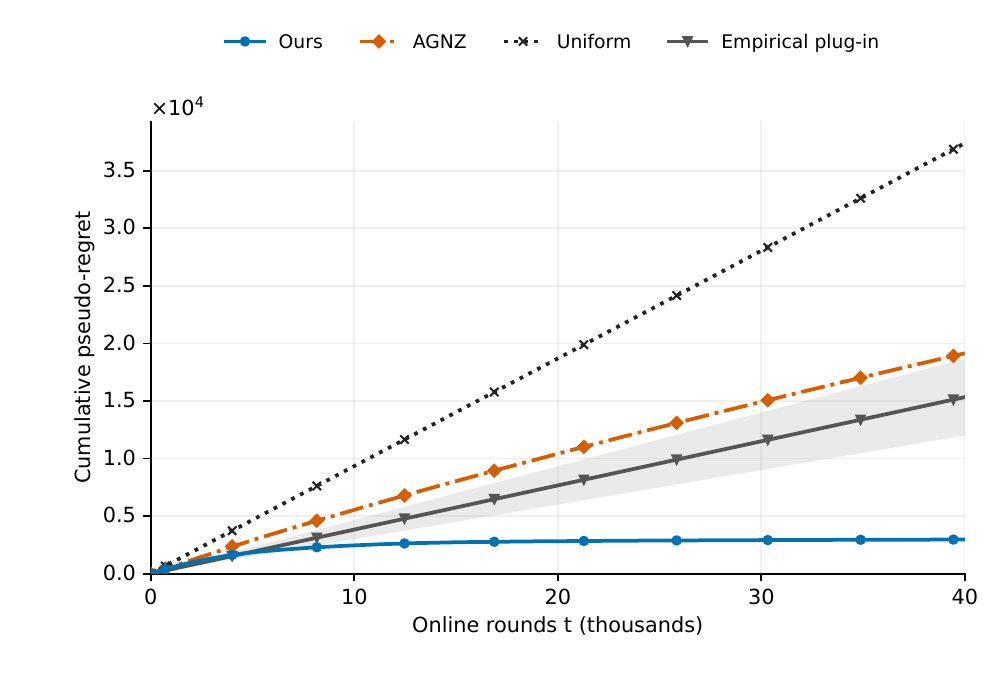}\caption{$n=100$.}\end{subfigure}

\medskip
\begin{subfigure}[t]{.485\textwidth}\centering\includegraphics[width=\linewidth]{plots/multi_prophet_k5_n500.pdf}\caption{$n=500$.}\end{subfigure}
\hfill
\begin{subfigure}[t]{.485\textwidth}\centering\includegraphics[width=\linewidth]{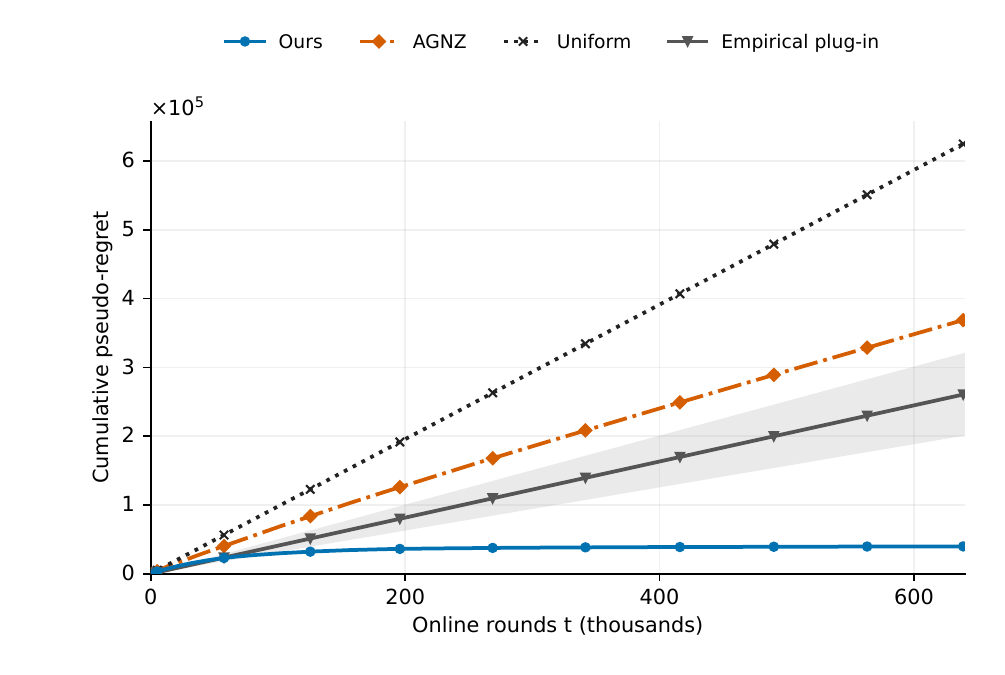}\caption{$n=1000$.}\end{subfigure}
\caption{Computational results for discrete prophet instances with $k=5$.}
\label{fig:app-prophet-discrete-five}
\end{figure*}

\begin{figure*}[htbp]
\centering
\begin{subfigure}[t]{.485\textwidth}\centering\includegraphics[width=\linewidth]{plots/without_lazy_prophet_k1_n50.pdf}\caption{$n=50$.}\end{subfigure}\hfill
\begin{subfigure}[t]{.485\textwidth}\centering\includegraphics[width=\linewidth]{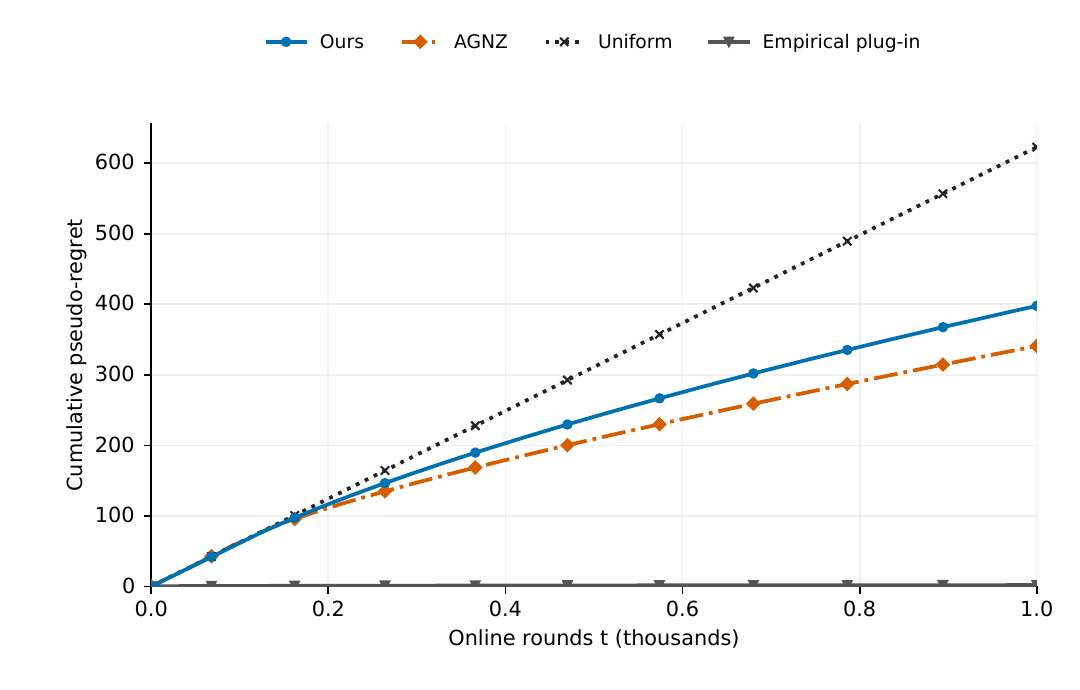}\caption{$n=100$.}\end{subfigure}

\medskip
\begin{subfigure}[t]{.485\textwidth}\centering\includegraphics[width=\linewidth]{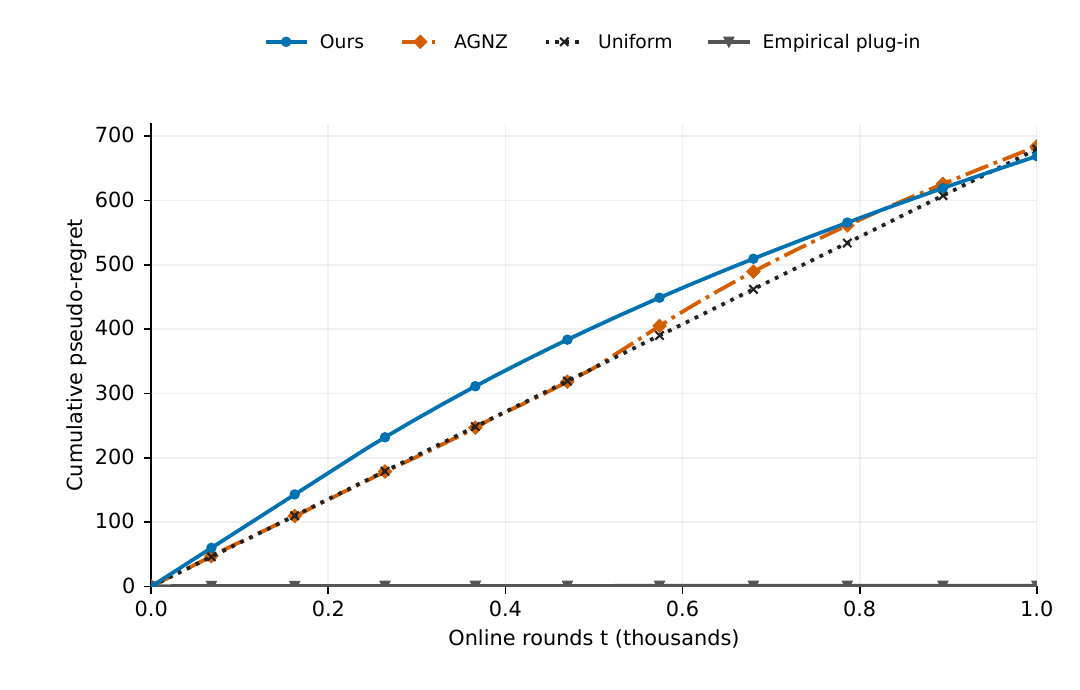}\caption{$n=500$.}\end{subfigure}
\hfill
\begin{subfigure}[t]{.485\textwidth}\centering\includegraphics[width=\linewidth]{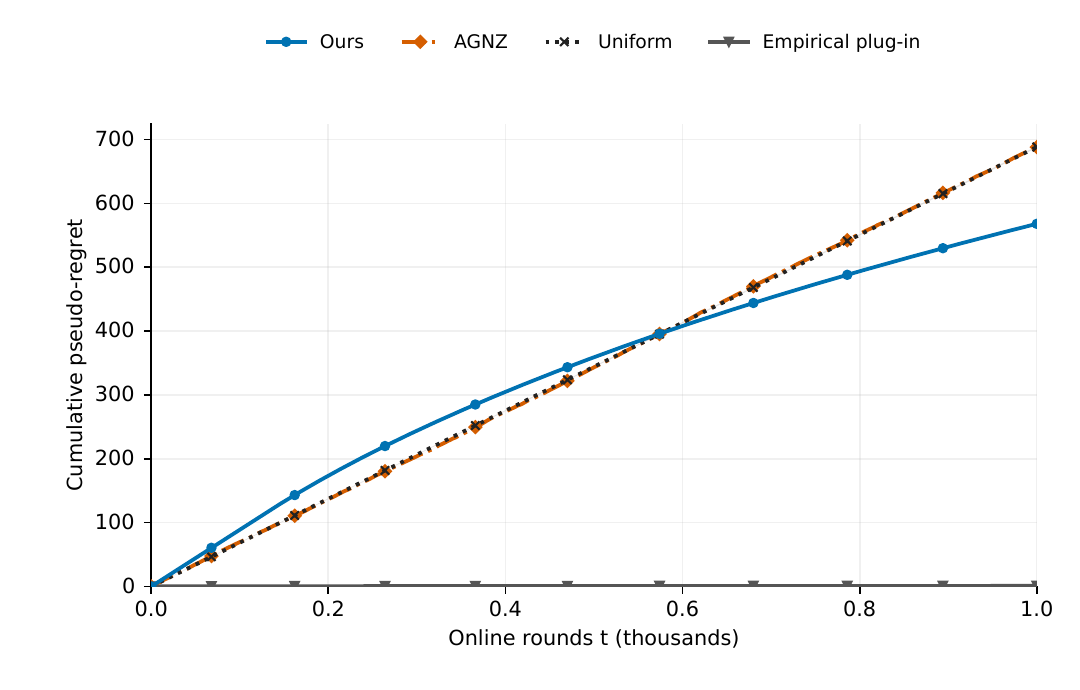}\caption{$n=1000$.}\end{subfigure}
\caption{Computational results for Uniform-endpoint prophet instances with $k=1$.}
\label{fig:app-prophet-uniform-one}
\end{figure*}

\begin{figure*}[htbp]
\centering
\begin{subfigure}[t]{.485\textwidth}\centering\includegraphics[width=\linewidth]{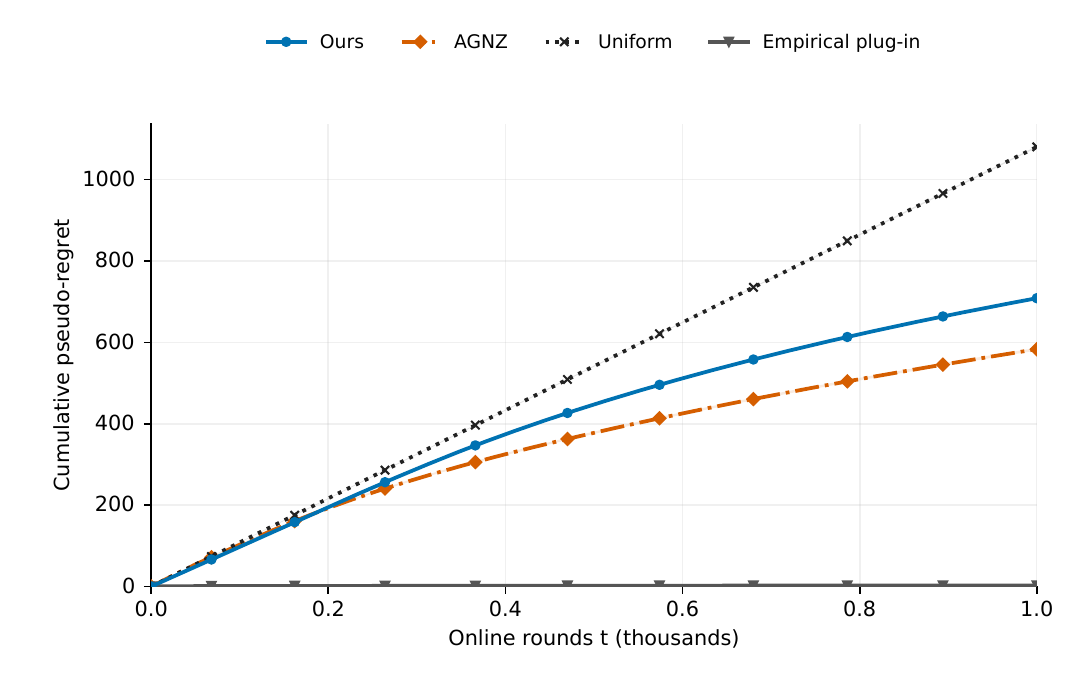}\caption{$n=50$.}\end{subfigure}\hfill
\begin{subfigure}[t]{.485\textwidth}\centering\includegraphics[width=\linewidth]{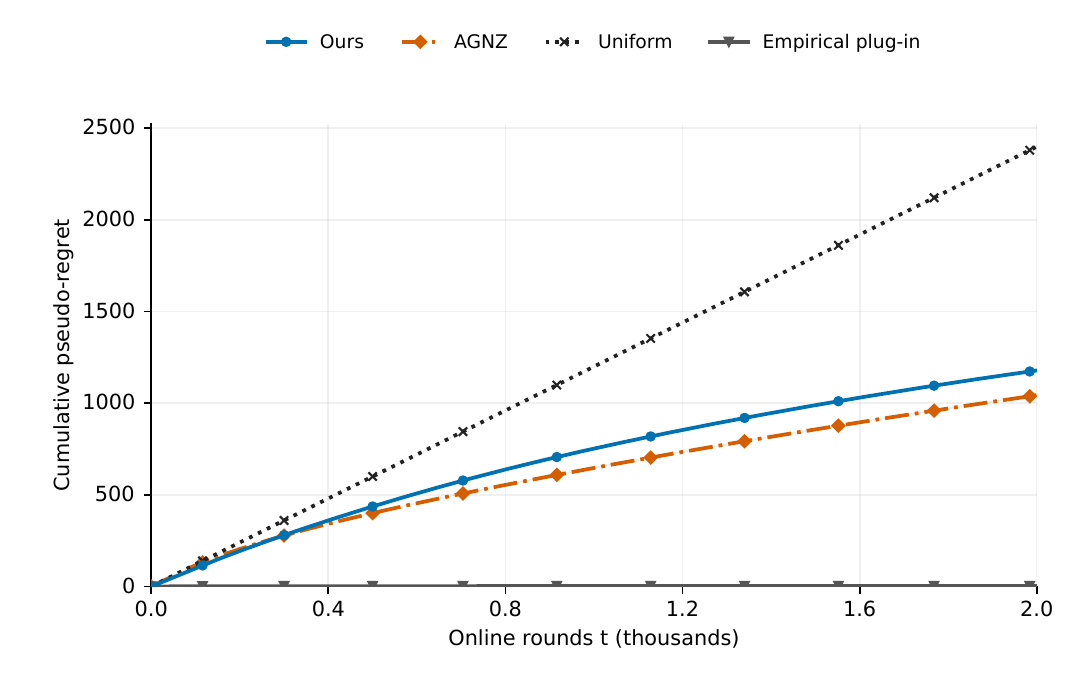}\caption{$n=100$.}\end{subfigure}

\medskip
\begin{subfigure}[t]{.485\textwidth}\centering\includegraphics[width=\linewidth]{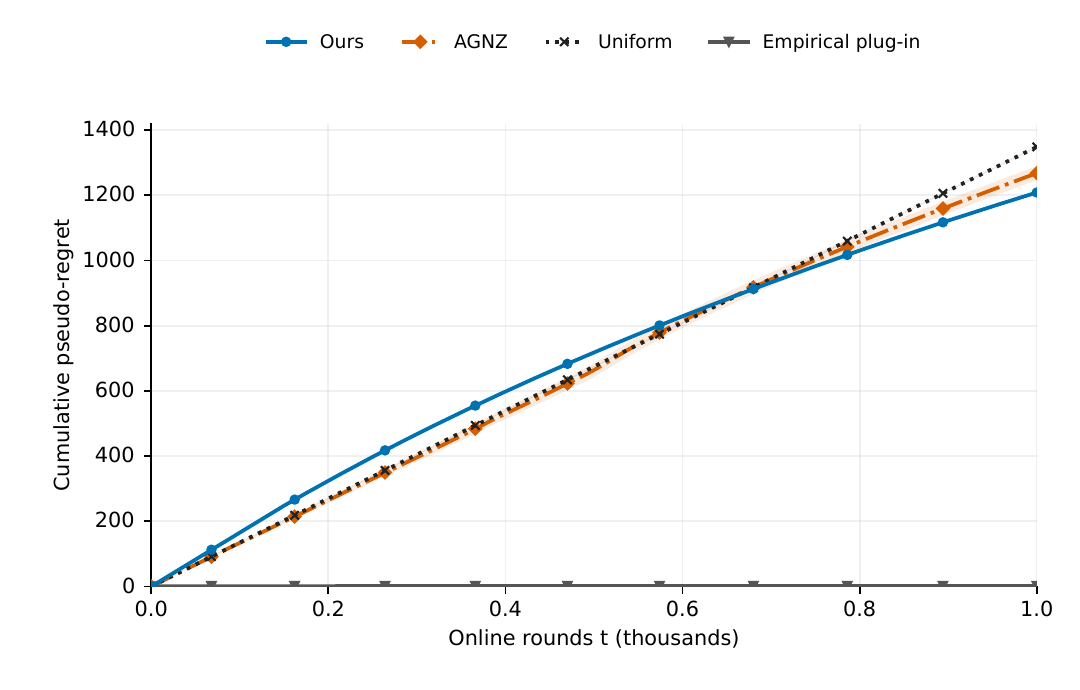}\caption{$n=500$.}\end{subfigure}\hfill
\begin{subfigure}[t]{.485\textwidth}\centering\includegraphics[width=\linewidth]{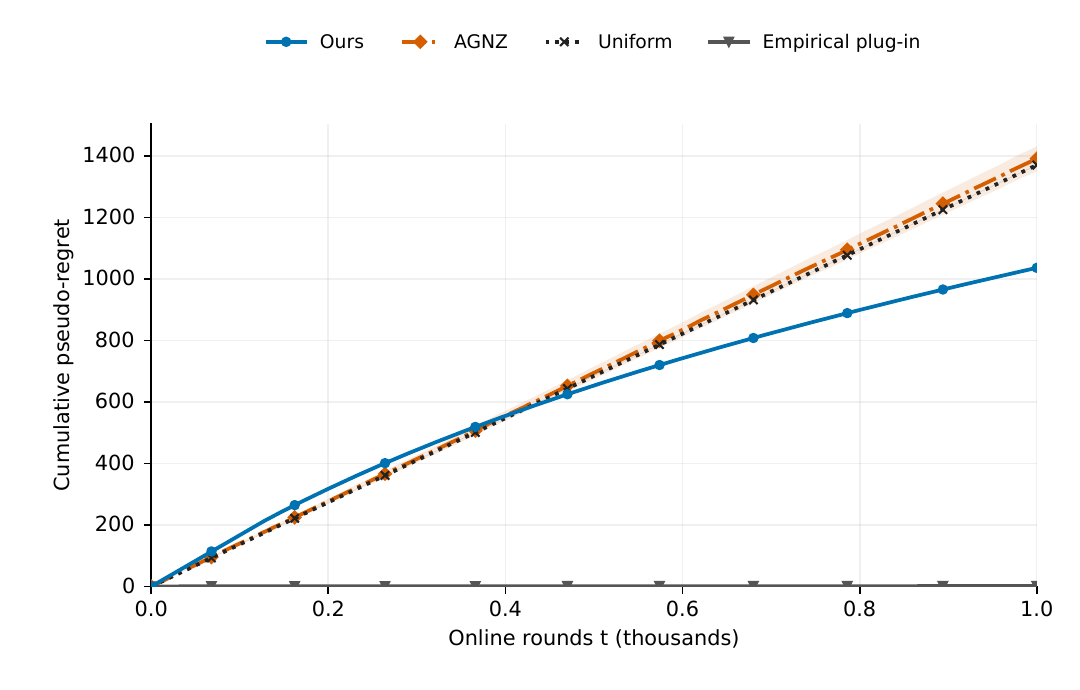}\caption{$n=1000$.}\end{subfigure}
\caption{Computational results for Uniform-endpoint prophet instances with $k=2$.}
\label{fig:cont-prophet-k2}
\end{figure*}

\begin{figure*}[htbp]
\centering
\begin{subfigure}[t]{.485\textwidth}\centering\includegraphics[width=\linewidth]{plots/without_lazy_prophet_k5_n50.pdf}\caption{$n=50$.}\end{subfigure}\hfill
\begin{subfigure}[t]{.485\textwidth}\centering\includegraphics[width=\linewidth]{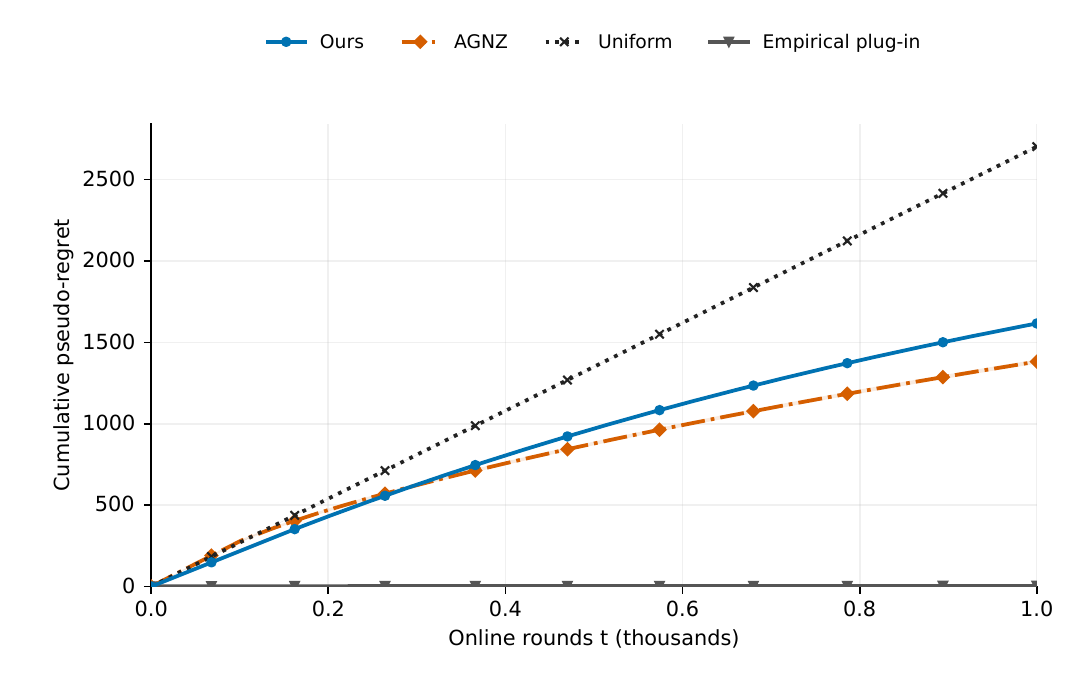}\caption{$n=100$.}\end{subfigure}

\medskip
\begin{subfigure}[t]{.485\textwidth}\centering\includegraphics[width=\linewidth]{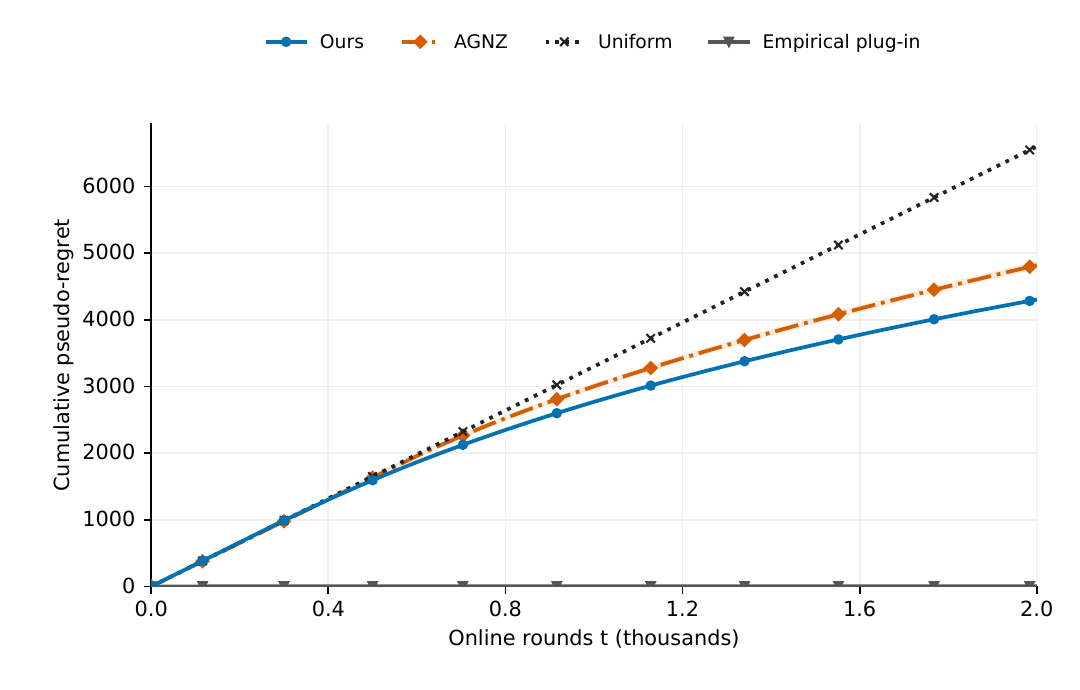}\caption{$n=500$.}\end{subfigure}
\hfill
\begin{subfigure}[t]{.485\textwidth}\centering\includegraphics[width=\linewidth]{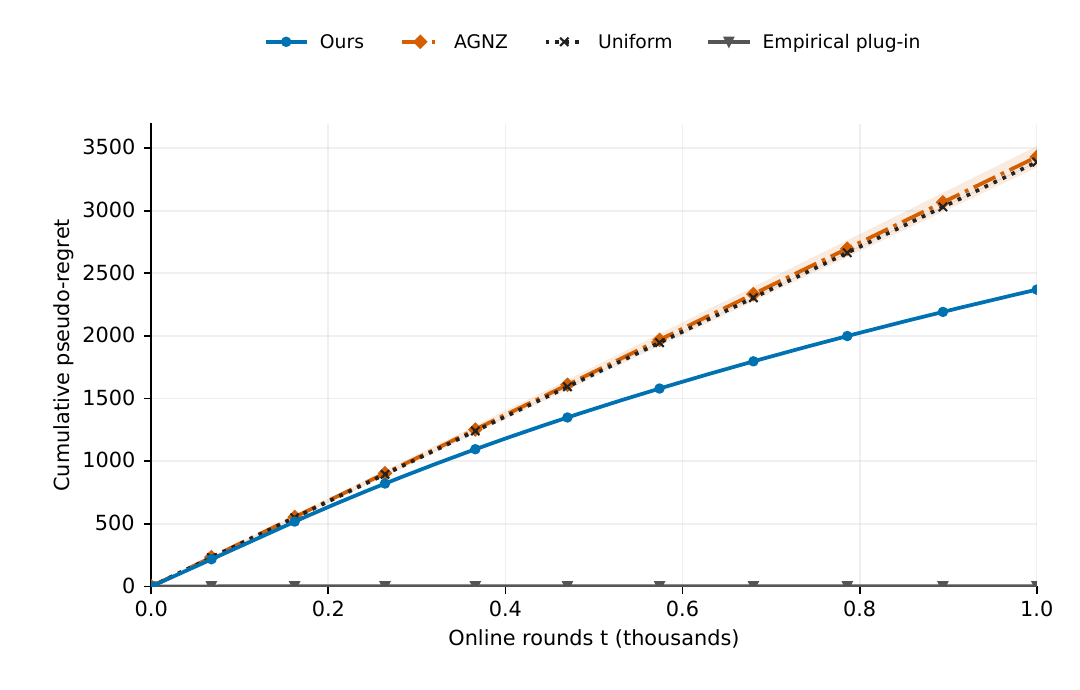}\caption{$n=1000$.}\end{subfigure}
\caption{Computational results for Uniform-endpoint prophet instances with $k=5$.}
\label{fig:app-prophet-uniform-five}
\end{figure*}

\begin{figure*}[htbp]
\centering
\includegraphics[width=.95\textwidth]{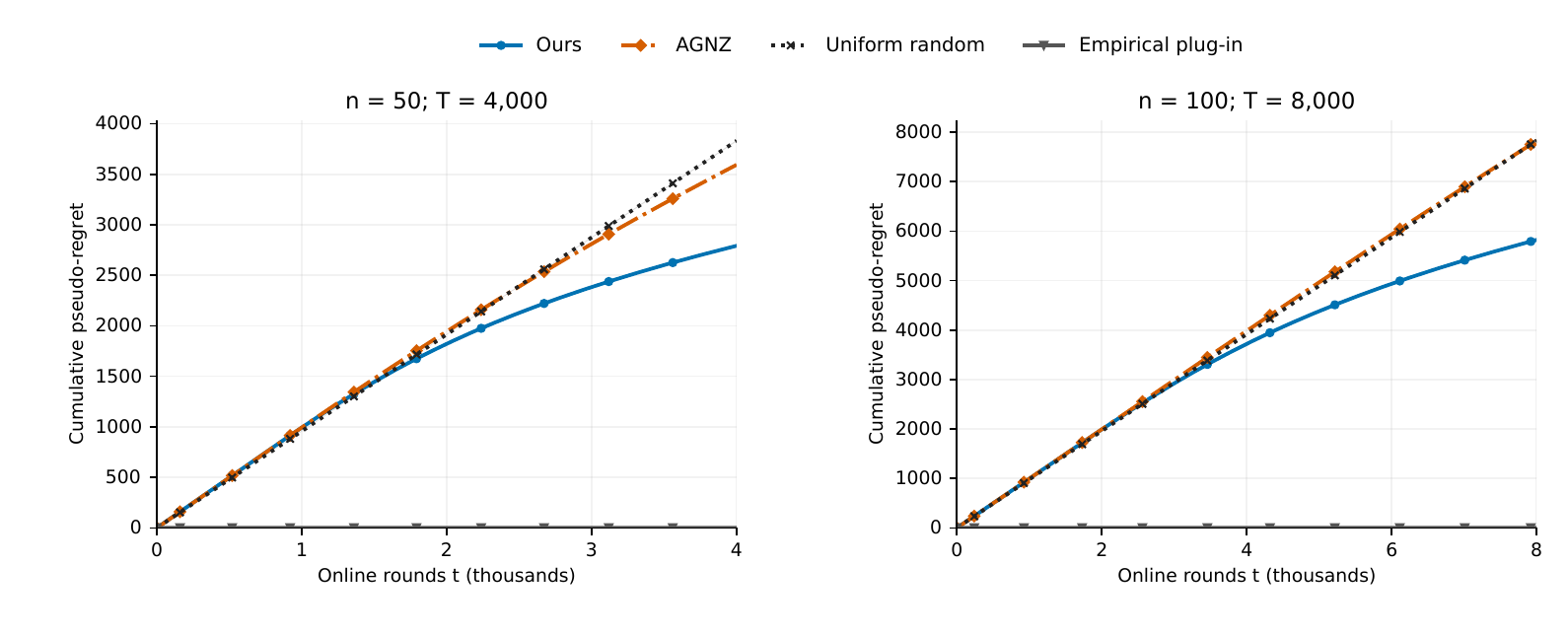}
\caption{Computational results for Uniform/Beta prophet instances with $k=2$.}
\label{fig:app-prophet-beta-two}
\end{figure*}

\begin{figure*}[htbp]
\centering
\includegraphics[width=.95\textwidth]{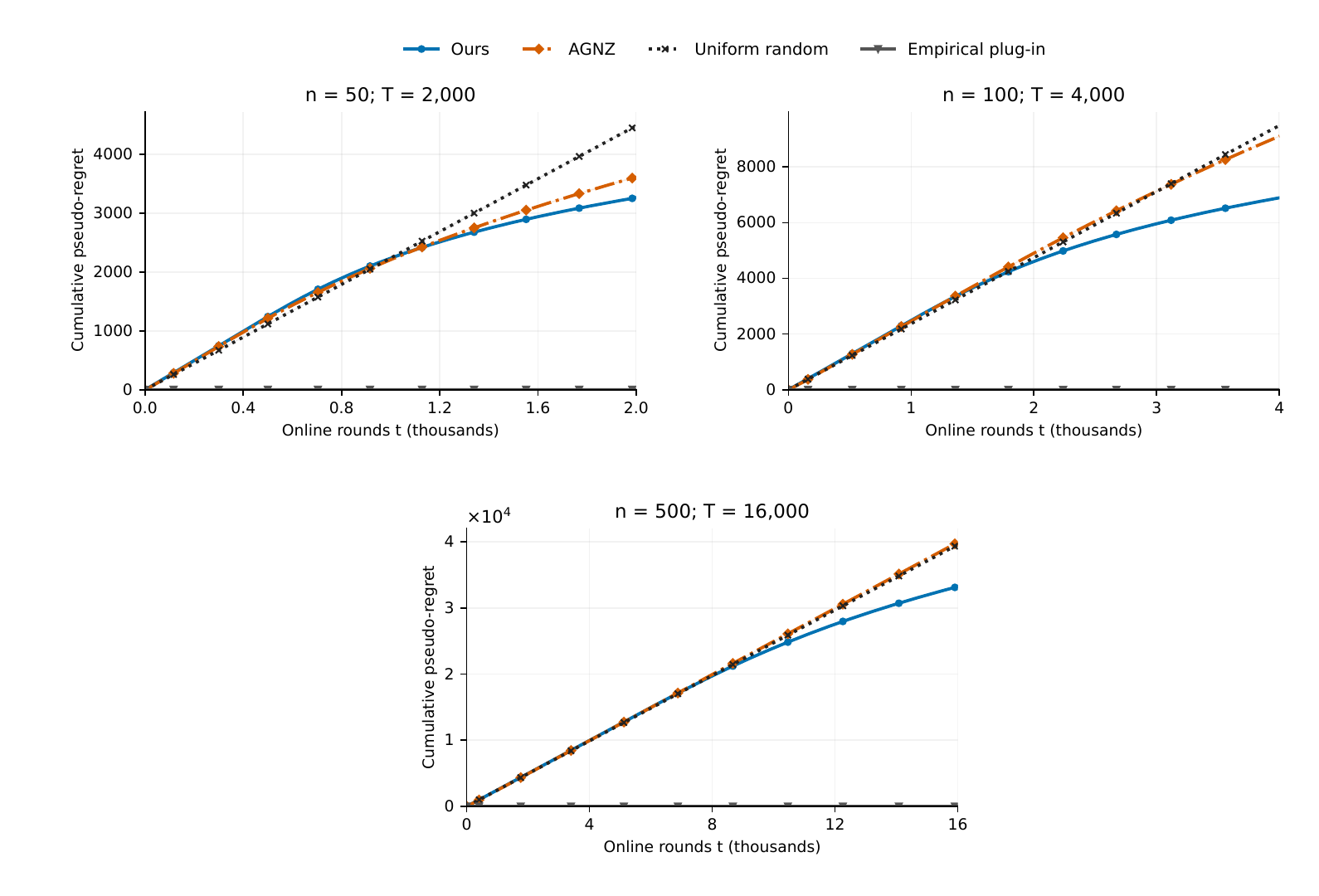}
\caption{Computational results for Uniform/Beta prophet instances with $k=5$.}
\label{fig:app-prophet-beta-five}
\end{figure*}

\clearpage
\subsection{Pandora's Box}
\label{app:pandora}

We report the remaining Pandora experiments described in \Cref{sec:computations}. In the Uniform/Beta instances, $k$ denotes the number of Uniform boxes.

\begin{figure*}[htbp]
\centering
\begin{subfigure}[t]{.485\textwidth}
\centering
\includegraphics[width=\linewidth]{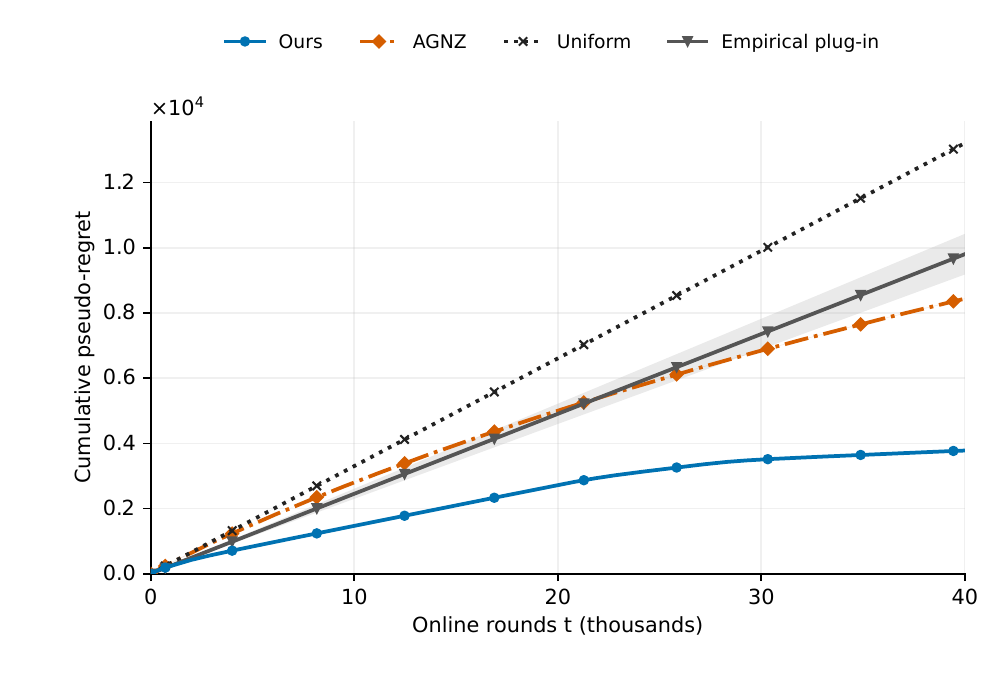}
\caption{$n=50$.}
\end{subfigure}\hfill
\begin{subfigure}[t]{.485\textwidth}
\centering
\includegraphics[width=\linewidth]{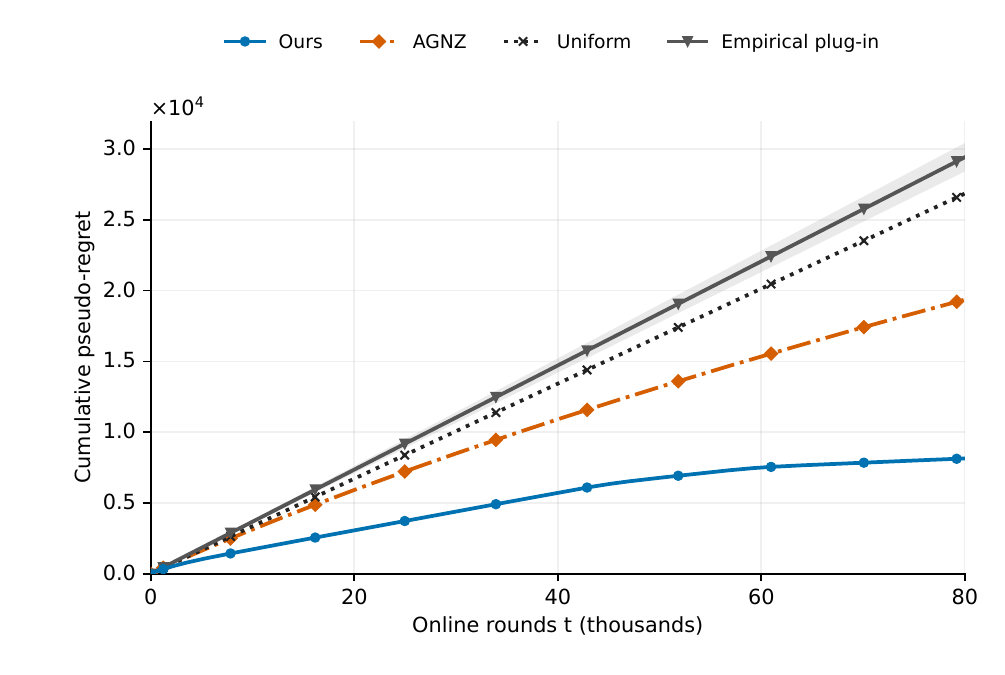}
\caption{$n=100$.}
\end{subfigure}

\medskip
\begin{subfigure}[t]{.485\textwidth}
\centering
\includegraphics[width=\linewidth]{plots/pandora_n500.pdf}
\caption{$n=500$.}
\end{subfigure}\hfill
\begin{subfigure}[t]{.485\textwidth}
\centering
\includegraphics[width=\linewidth]{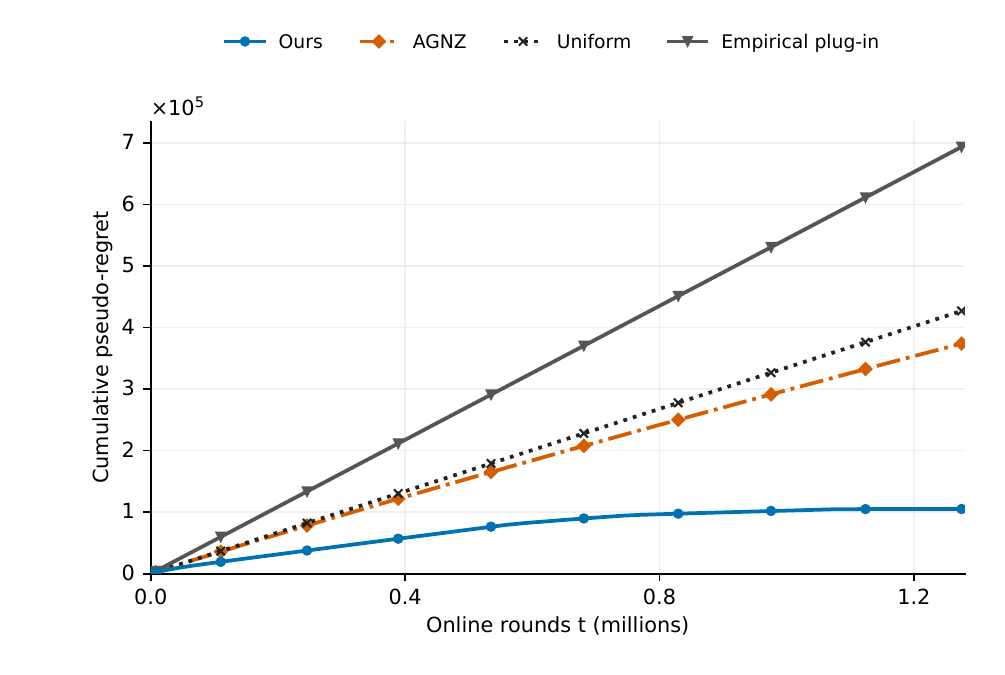}
\caption{$n=1000$.}
\end{subfigure}
\caption{Computational results for discrete Pandora instances.}
\label{fig:app-pandora-discrete}
\end{figure*}

\begin{figure*}[htbp]
\centering
\begin{subfigure}[t]{.485\textwidth}\centering\includegraphics[width=\linewidth]{plots/without_lazy_pandora_k1_n50.pdf}\caption{$n=50$.}\end{subfigure}\hfill
\begin{subfigure}[t]{.485\textwidth}\centering\includegraphics[width=\linewidth]{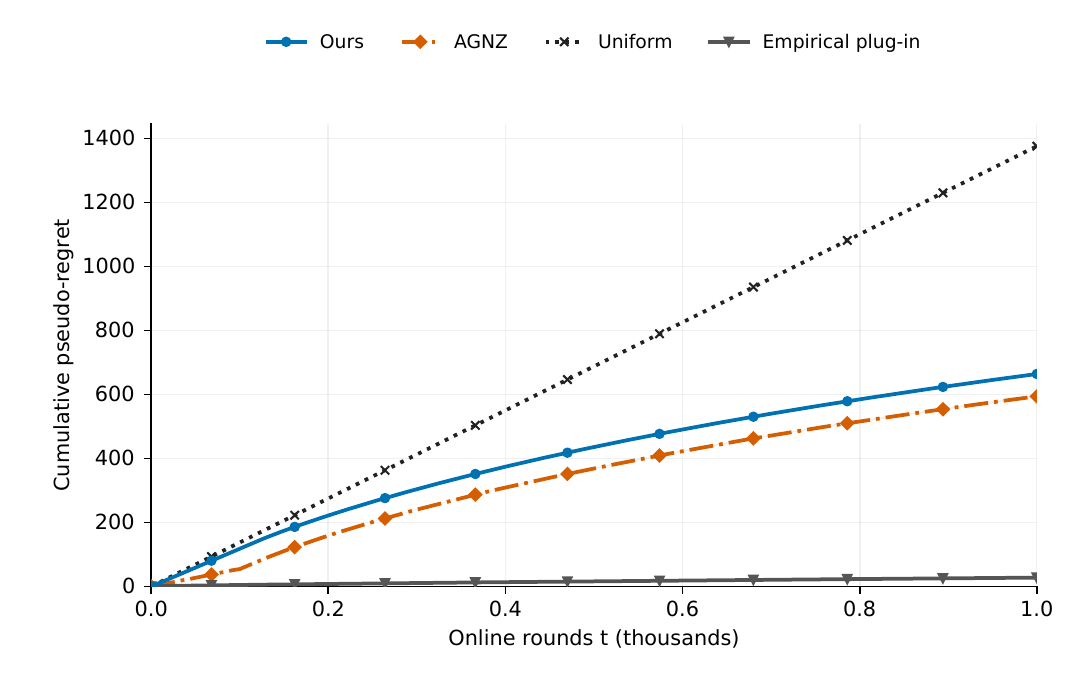}\caption{$n=100$.}\end{subfigure}

\medskip
\begin{subfigure}[t]{.485\textwidth}\centering\includegraphics[width=\linewidth]{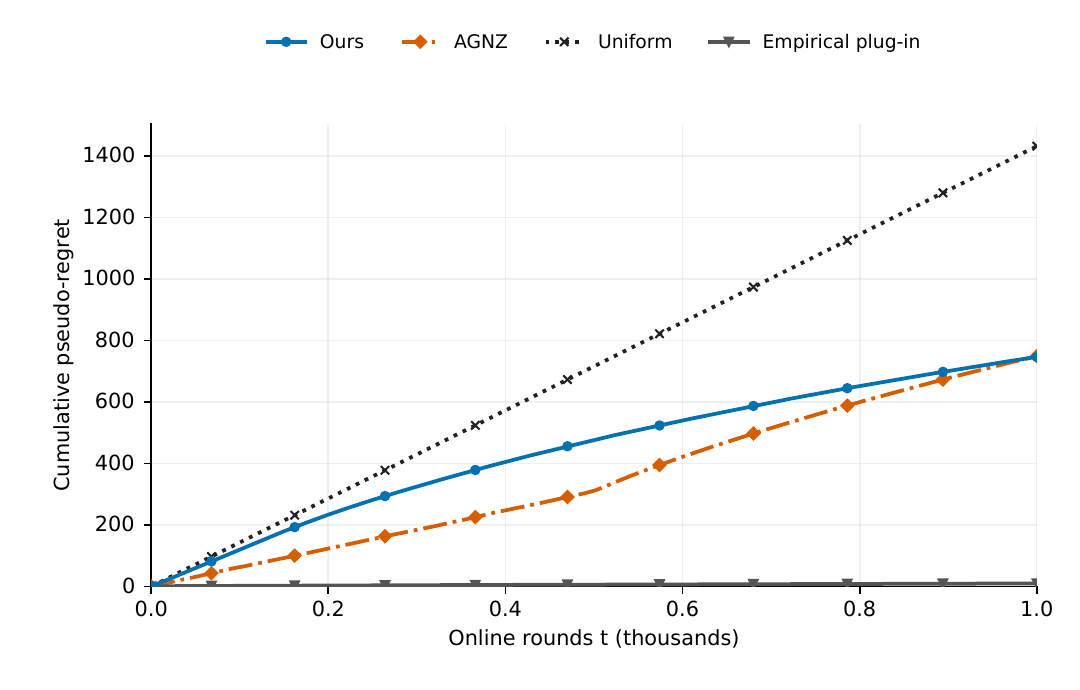}\caption{$n=500$.}\end{subfigure}\hfill
\begin{subfigure}[t]{.485\textwidth}\centering\includegraphics[width=\linewidth]{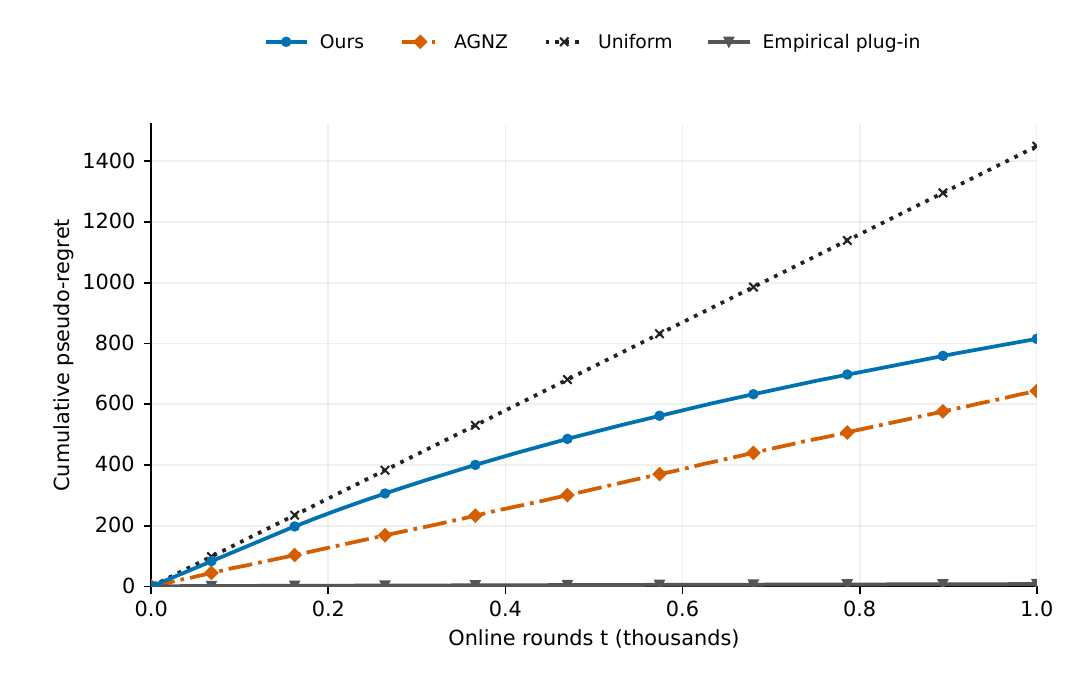}\caption{$n=1000$.}\end{subfigure}
\caption{Computational results for Uniform-endpoint Pandora instances.}
\label{fig:app-pandora-uniform}
\end{figure*}

\begin{figure*}[htbp]
\centering
\includegraphics[width=.95\textwidth]{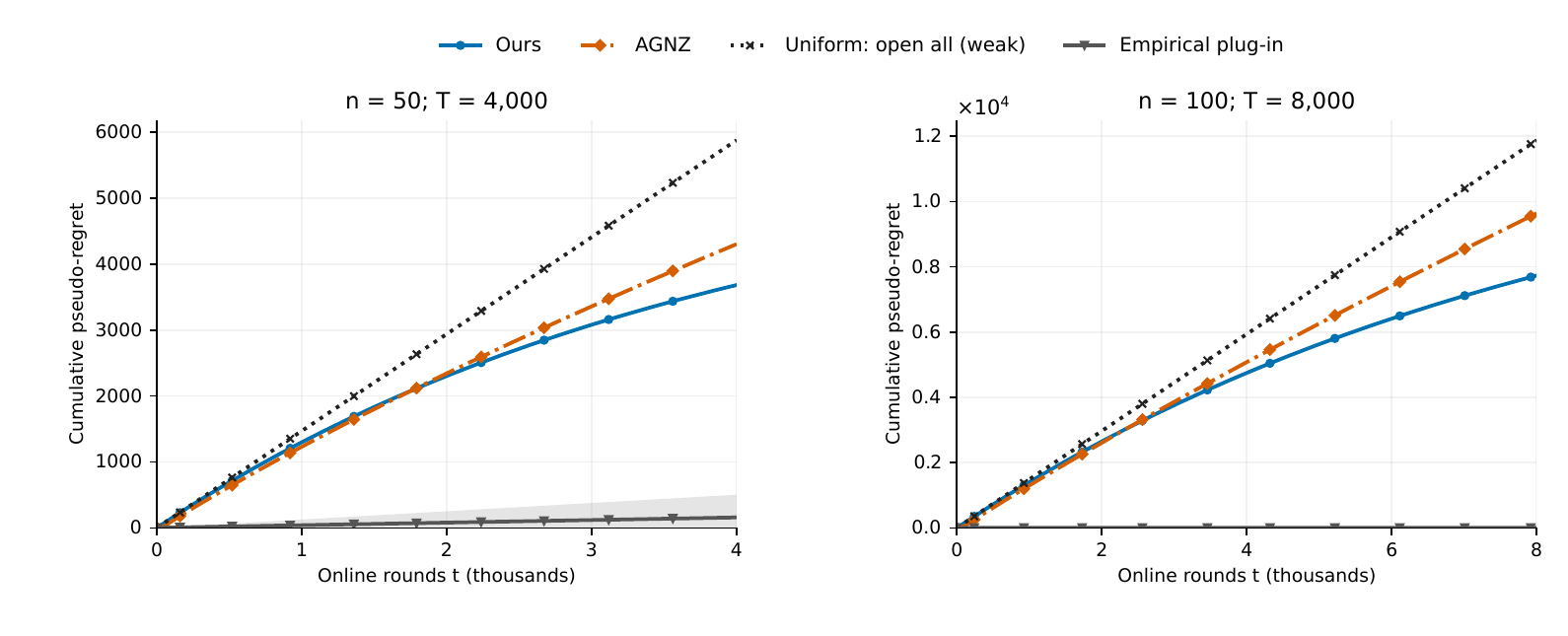}
\caption{Computational results for Uniform/Beta Pandora instances with $k=1$.}
\label{fig:app-pandora-beta-one}
\end{figure*}

\begin{figure*}[htbp]
\centering
\includegraphics[width=.95\textwidth]{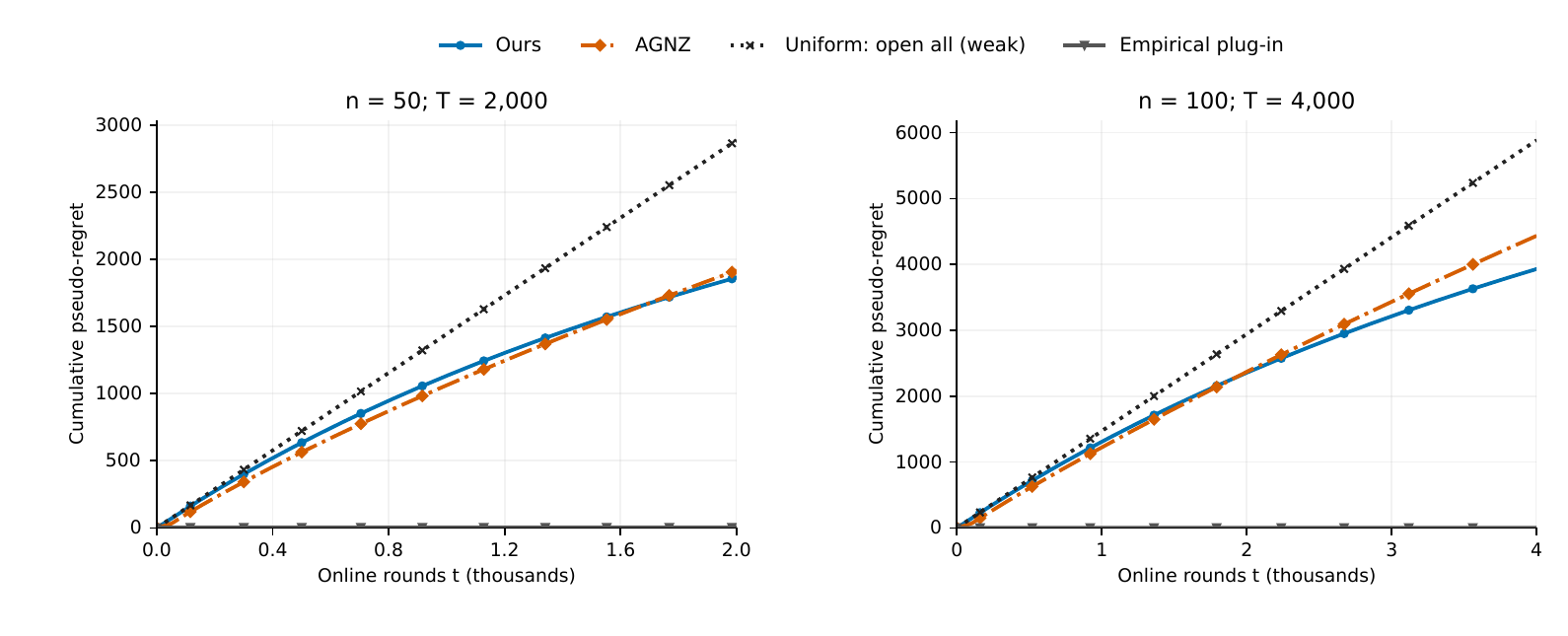}
\caption{Computational results for Uniform/Beta Pandora instances with $k=2$.}
\label{fig:app-pandora-beta-two}
\end{figure*}